\documentclass[lettersize,journal]{IEEEtran}
\usepackage{amsmath,amsfonts}
\usepackage{algorithm}
\usepackage{array}
\usepackage[font=normalsize,labelfont=sf,textfont=sf]{subfig}
\usepackage{textcomp}
\usepackage{stfloats}
\usepackage{url}
\usepackage{verbatim}
\usepackage{graphicx}
\usepackage{cite}
\usepackage{url}
\usepackage{color}
\usepackage{multirow}
\usepackage{booktabs}
\usepackage{bbm}
\usepackage{soul}
\usepackage{graphicx}
\usepackage{booktabs}
\usepackage{tabularx}
\usepackage{bm}
\usepackage{adjustbox}
\usepackage{lipsum}
\usepackage{pifont}
\usepackage{algpseudocode}
\usepackage{amsmath,amsfonts}
\usepackage{tikz,xcolor}
\usepackage{hyperref}
\hypersetup{colorlinks=true,
	allcolors=black,
	pdfstartview=Fit,
	breaklinks=true}

\definecolor{lime}{HTML}{A6CE39}
\DeclareRobustCommand{\orcidicon}{
\begin{tikzpicture}
\draw[lime, fill=lime] (0,0)
circle[radius=0.16]
node[white]{{\fontfamily{qag}\selectfont \tiny \.{I}D}};
\end{tikzpicture}
\hspace{-2mm}
}
\foreach \x in {A, ..., Z}{%
\expandafter\xdef\csname orcid\x\endcsname{\noexpand\href{https://orcid.org/\csname orcidauthor\x\endcsname}{\noexpand\orcidicon}}
}

\begin{document}
\title{Fourier Basis Mapping: A Time-Frequency Learning Framework for Time Series Forecasting}


\author{Runze Yang\hspace{-1.5mm}\orcidA{},
Longbing Cao\hspace{-1.5mm}\orcidD{},
Xin You, \hspace{-1.5mm}\orcidE{},
Kun Fang, \hspace{-1.5mm}\orcidF{},
Jianxun Li\hspace{-1.5mm}\orcidB{},
Jie Yang\hspace{-1.5mm}\orcidC{}


\thanks{ R. Yang is with the School of Computing, Macquarie University, Sydney, Australia, and the Department of Automation, Shanghai Jiao Tong University, Shanghai, China, (e-mails: runze.yang@hdr.mq.edu.au, runze.y@sjtu.edu.cn). L. Cao is with the School of Computing, Macquarie University, Sydney, Australia (e-mail: longbing.cao@mq.edu.au). X. You, J. Yang and J. Li are with the Department of Automation, Shanghai Jiao Tong University, Shanghai, China (e-mails: \{sjtu\_youxin, jieyang, lijx\}@sjtu.edu.cn). Kun Fang is with the department of Electrical and Electronic Engineering, Hong Kong Polytechnic University, Hong Kong, China. (e-mails: kun.fang@polyu.edu.hk). This research is partially supported by NSFC (No. 62376153), and ARC DP240102050 and LE240100131.
Corresponding authors: L. Cao and J. Yang. The project is online at: https://github.com/runze1223/FBM-S }
}
\markboth{IEEE Transactions on xxxxxxxxx}%
{Shell \MakeLowercase{\textit{et al.}}: A Sample Article Using IEEEtran.cls for IEEE Journals}

\maketitle

\begin{abstract}
The integration of Fourier transform with deep learning opens new avenues for time series forecasting. We reconsider the Fourier transform from a basis functions perspective. Specifically, the real and imaginary parts of the frequency components can be regarded as the coefficients of cosine and sine basis functions at tiered frequency levels, respectively. We find that existing Fourier-based methods face inconsistent starting cycles and inconsistent series length issues, failing to interpret frequency components precisely and overlooking temporal information. Accordingly, the proposed novel Fourier Basis Mapping (FBM) method addresses these issues by integrating time-frequency features through Fourier basis expansion and mapping in the time-frequency space. Our approach extracts explicit frequency features while preserving temporal characteristics. FBM supports plug-and-play integration with various types of neural networks by only adjusting the first initial projection layer for better performance. First, we propose FBM-L, FBM-NL, and FBM-NP to enhance linear, MLP-based, and Transformer-based models, respectively, demonstrating the effectiveness of time-frequency features. Next, we propose a synergetic model architecture, termed FBM-S, to decompose the seasonal, trend, and interaction effects into three separate blocks, each designed to model time-frequency features in a specialized manner. Finally, we introduce several techniques tailored for time-frequency features, including interaction masking, centralization, patching, rolling window projection, and multi-scale down-sampling. The results are validated on diverse real-world datasets for both long-term and short-term forecasting tasks with SOTA performance. 

\end{abstract}


\section{Introduction}
\IEEEPARstart{T}{ime} series forecasting (TSF) plays a vital role across a wide range of industries, including energy, weather prediction, financial markets, and transportation systems. For example, time series forecasting is crucial for forecasting weather to support disaster readiness, predicting financial market movements to inform investment strategies and policymaking, and estimating traffic flow to aid in urban planning and optimize transportation systems. However, it faces considerable challenges, such as modeling both short-term and long-term temporal dependencies, along with frequency-oriented global dynamics. Recently, deep neural networks (DNNs) have thrived to tackle TSF challenges for the presence of hierarchical effects, varied outliers, and nonlinear dynamics. They use various DNN architectures including recurrent neural networks (RNNs) \cite{hochreiter1997long,chung2014empirical,rangapuram2018deep,chang2017dilated,liu2021impact,gangopadhyay2021spatiotemporal,jia2023witran,salinas2020deepar,zhou2022film},  convolution neural networks (CNNs) \cite{luo2024moderntcn,bai2018empirical,liu2022scinet,wang2022micn,franceschi2019unsupervised,sen2019think, wu2022timesnet}, multi-layer perceptron (MLP)-based networks \cite{TimeMixer,tang2025unlocking,chen2023tsmixer,oreshkin2019n, challu2023nhits,yi2023frequency}, Transformer-based networks \cite{vaswani2017attention,nie2022time,ni2023basisformer,zhou2021informer,wu2021autoformer,liu2021pyraformer,zhang2022crossformer,cao2023inparformer,fu2024encoder,zhang2023temporal,cao2023tempo,Pathformer,LiuHZWWML24,qiu2024duet, liu2024timer, fan2022sepformer, zhou2022fedformer} and graph neural networks (GNNs)\cite{huang2023crossgnn,yi2023fouriergnn, cini2023graph, sriramulu2023adaptive, jin2022multivariate} and Mamba-based networks \cite{alkilane2024mixmamba, wang2025mamba, zeng2024cmamba}. Interestingly, the recent NLinear study \cite{zeng2023transformers} demonstrates that a  simple normalized linear model can surprisingly surpass the performance of most DNN-based approaches. This raises the question: \textit{Would a complex DNN architecture necessarily lead to better TSF performance?} According to CrossGNN \cite{huang2023crossgnn}, DNN-based methods are susceptible to noisy inputs, often assigning high attention scores or weights to irrelevant or unexpected signals.

\begin{figure*}[t]  
\centering
\includegraphics[width=\linewidth]{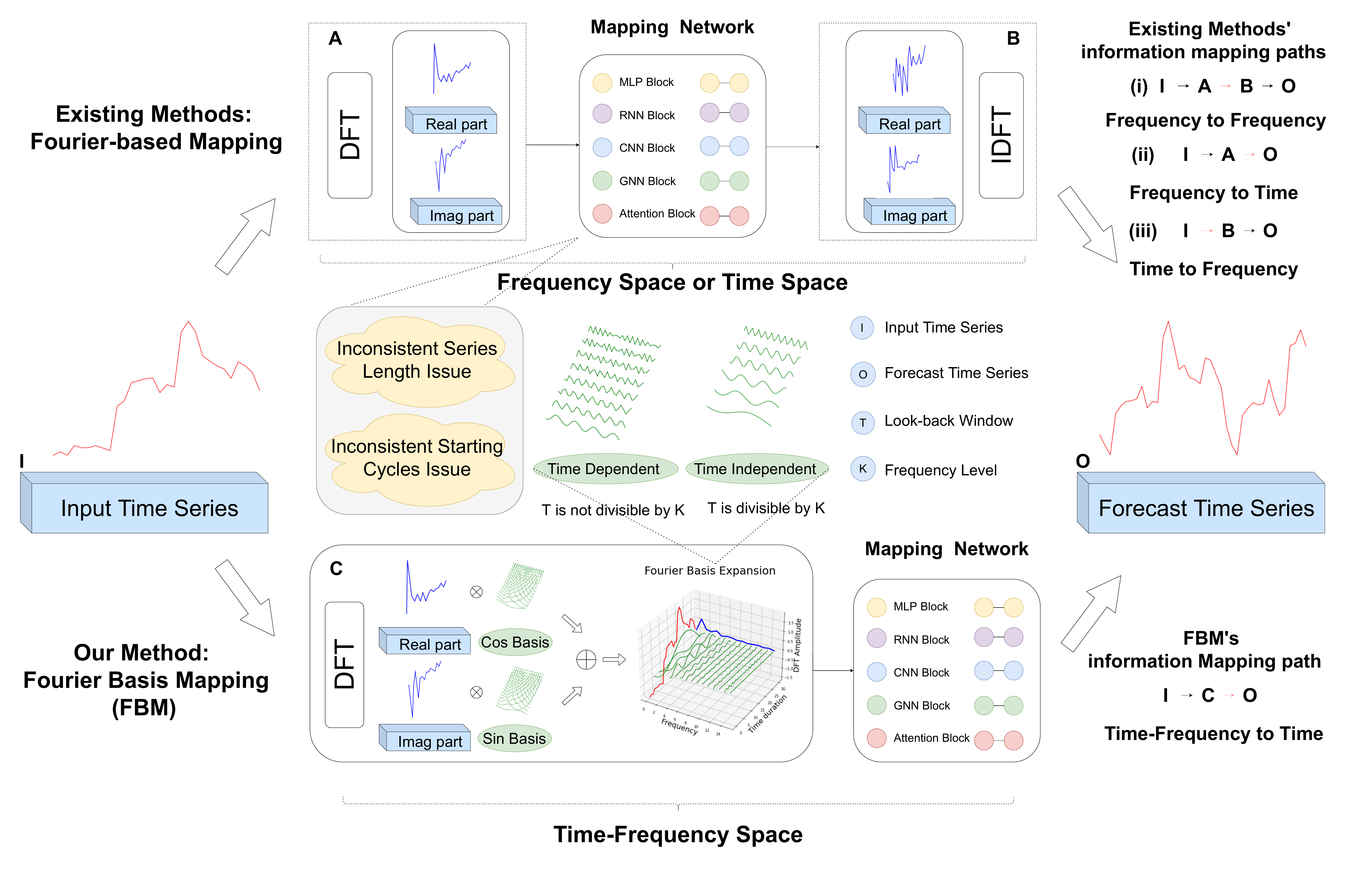}
\vspace{-1.5em}
\caption{Comparison of Existing Fourier-based Methods with Our Approach Fourier Basis Mapping (FBM). Existing methods primarily operate in the frequency or time space, with their mapping focusing on frequency or time features. FBM simultaneously operates in time-frequency space, with the mapping focusing on time-frequency features.}
\vspace{-15pt}
\label{summary}
\end{figure*}

Thus, Fourier-based time series modeling emerges as a new paradigm to remove noise signals by decomposing diverse effects hierarchically at different frequency levels. If we rethink the Fourier transform from a basis functions perspective, the real and imaginary parts can be interpreted as the coefficients of cosine and sine basis functions at tiered frequencies, respectively. However, existing Fourier-based methods do not involve basis functions, thus failing to interpret frequency coefficients precisely and do not consider the time-frequency relationships sufficiently. They face two main issues: inconsistent starting cycles and inconsistent series length issues, as detailed in Section \ref{section2.1}. For instance, FEDformer \cite{zhou2022fedformer}, FreTS \cite{yi2023frequency}, FiLM \cite{zhou2022film}, FITS \cite{xu2023fits}, FGNet \cite{fu2024encoder}, and FL-Net \cite{huang2024fl} use the real and imaginary parts of frequency components as input and conduct the mapping in the frequency space. From our above perspective, we find that the amplitude and arctangent of the real and imaginary components carry more explicit meanings than the components themselves, as adding a cosine wave and a sine wave with the same frequency leads to a shifted cosine wave with the same frequency but new amplitude and phase. Furthermore, the meanings of frequency components are bounded by the series length, and overlooking this causes ambiguity in interpreting precise frequencies in these models. For example, a $k$ Hz sine or cosine wave can have different meanings depending on the series length. More importantly, Fig. \ref{summary} shows that a Fourier basis function is time-dependent when the input length is not divisible by the frequency level, making it even more challenging for the model to accurately interpret those frequency components without the basis functions. Existing methods receive these coefficients unaware they correspond to specific sine and cosine basis functions. Consequently, constructing the mapping in the frequency space is not enough and neglects temporal information, which has been ignored by existing Fourier-based methods. Although some methods like TimesNet \cite{wu2022timesnet} consider both time and frequency features, the original time domain information has been compromised, as summation over the frequency dimension does not recover the original time series. We provide a more detailed discussion of their limitations in Section \ref{rela}.

Accordingly, we propose Fourier Basis Mapping (FBM) to address the aforementioned issues by involving Fourier basis functions, allowing the model to capture explicit frequency information from a global perspective while retaining temporal characteristics for fine-grained representations. In Fig. \ref{summary}, we visualize the difference between our method and existing Fourier-based methods. In the first stage, we embed the discrete Fourier transform with basis functions, referred to as \textit{Fourier basis expansion}, to extract time-frequency features. In the second stage, we map the time-frequency features into output time series. As our time-frequency features retain time domain information, thus it can be applied to any time-based mapping methods, by only adjusting the first initial projection layer but considering both time and frequency modality. First, we validate the effectiveness of  time-frequency features to enhance Linear-, MLP-, and Transformer-based networks, resulting in three FBM variants: FBM-L, FBM-NL, and FBM-NP. This shows that Fourier basis expansion can enhance any type of DNNs, serving as a plug-and-play module by only adjusting the initial layer. Second, we propose a synergetic FBM model, referred to as FBM-S, investigating how to model time-frequency features more effectively. The model decomposes seasonal, trend, and interaction effects into three separate blocks, each of them models the time-frequency features in a specialized manner. Our method highlights that separating endogenous and exogenous effects into separate blocks is crucial. Existing studies either use independent channels or dependent channels but fail to combine the strengths of both effectively. \textit{This is because they overlook the fact that the influence between multivariate time series usually occurs over short-term periods.} Therefore, input and output masks are applied to the interaction blocks to enable smoother integration of their respective strengths. Finally, we introduce several techniques tailored for time-frequency features within each block, including masking and centralization in the interaction block; patching, centralization and multi-scale down-sampling in the trend block; and rolling window projections in the seasonal block.

The proposed FBM variants are compared against six categories of TSF baseline methods: (1) Linear method, (2) Transformer-based methods, (3) MLP-based methods, (4) RNNs-based methods, (5) CNNs-based methods, and (6) Fourier-based methods. Evaluations are conducted on diverse datasets for both long-term and short-term forecasting tasks with SOTA performance.


Our main contributions include the following:
\begin{itemize}
    \item We identify two issues in existing Fourier-based methods: inconsistent starting cycles issue and inconsistent series length issue. 
    \item We introduce the FBM framework to extract explicit time-frequency features, addressing the aforementioned issues by mapping within the time-frequency space.
    \item We demonstrate that FBM supports plug-and-play integration with various types of neural networks to enhance performance, as validated by three proposed FBM variants: FBM-L, FBM-NL, and FBM-NP.
    \item We propose FBM-S, a synergetic FBM model that efficiently captures time-frequency relationships by decomposing effects into trend, seasonal, and interaction components within three specialized blocks, incorporating several useful techniques for time-frequency features.
    

\end{itemize}

This study is an extension of its conference version \cite{yang2024rethinking}. In the  conference version, we first point out the inconsistent starting cycles and series length issues, and demonstrated the effectiveness of the proposed time-frequency through plug-and-play in the existing methods, primarily focusing on the long-term TSF task. In this extension version, we further investigate how to use time-frequency features more effectively, and create a new architecture--a synergetic FBM model tailored for time-frequency mapping, focusing on both the long-term and short-term TSF tasks. 
This version includes substantial new developments, including: (i) We revise the workflow of the introduction and related work to reflect the new motivations mentioned above. (ii) We propose a new synergistic FBM model, termed FBM-S, which introduces trend, seasonal, and interaction decomposition along with several techniques tailored for time-frequency features, as detailed in Section \ref{FBMS}. (iii) In Section \ref{Experiment}, we add experimental settings for short-term TSF with more diverse prediction lengths on the PEMS dataset and include a new dataset M4. (iv) We completely rewrite Section \ref{Main_reults}, with a summary provided at the beginning. In Section \ref{Main_reults}, except for the parts on the long-term TSF results, new content includes short-term TSF results, efficiency analysis, ablation studies, visualization, and case studies for FBM-S to support the broader scope of this extended version.



\section{Related Work}
\label{rela}
Recent research has highlighted the potential of the Fourier transform in addressing challenges in TSF. Accordingly, we categorize the relevant work into two groups: (1) frequency-based methods and (2) time-based methods. We discuss the unique strengths and weaknesses of each approach and explain how our method is designed to overcome their limitations and complement their strengths.

\subsection{Frequency-based Methods}

Fourier transform has been integrated into a wide range of network architectures for TSF, including RNNs \cite{zhou2022film}, CNNs \cite{wu2022timesnet}, MLP-based networks \cite{yi2023frequency,oreshkin2019n,challu2023nhits}, Transformer-based networks \cite{zhou2022fedformer,fan2022sepformer}, and graph neural networks (GNNs) \cite{huang2023crossgnn,yi2023fouriergnn}. However, by rethinking the Fourier transform from a basis functions perspective, we identify the inconsistent starting cycles and series length issues. In  Fig. \ref{summary}, Path (i) refers to methods like FEDformer \cite{zhou2022fedformer}, FreTS \cite{yi2023frequency}, FiLM \cite{zhou2022film}, FITS \cite{xu2023fits}, FGNet \cite{fu2024encoder}, and FL-Net \cite{huang2024fl}, which use real and imaginary parts as inputs and conduct the mapping in a frequency space but the networks cannot easily interpret those coefficients because crucial information is stored in the amplitude, phase and length of each cycle, which are inferred by basis functions. Path (ii) refers to methods leveraging frequency information while temporal information is compromised. For example, CrossGNN \cite{huang2023crossgnn} uses the discrete Fourier transform (DFT) to select the top k amplitudes for noise filtering, retaining only the first cycle and neglecting the phase shift, along with the fact that the basis functions can become time dependent when the frequency is not divisible by the length of the series. Additionally, a higher amplitude does not necessarily indicate a useful frequency, and a lower amplitude is not necessarily useless. TimesNet \cite{wu2022timesnet} introduces the multi-window Fourier transform. However, their approach lacks mathematical rigor, as their extracted features are complex and inefficient. Both time- and frequency-domain information is compromised due to the use of windows, resulting in the loss of fine-grained characteristics. Path (iii) refers to methods such as N-BEATS \cite{oreshkin2019n} and N-Hits \cite{challu2023nhits}, which can similarly be viewed as forecasting through the inverse discrete Fourier transform (IDFT) by computing the output frequency spectrum of the time series, but their networks struggle to capture time-dependent effects and do not leverage frequency information effectively. In contrast, our FBM distinguishes itself from existing approaches by leveraging the Fourier basis expansion to provide a mixture of time-frequency features, thus avoiding the aforementioned issues.

\subsection{Time-based Methods}

The popular time-based architectures for TSF involve RNNs \cite{hochreiter1997long,chung2014empirical,rangapuram2018deep,chang2017dilated,liu2021impact,gangopadhyay2021spatiotemporal,jia2023witran,salinas2020deepar}, CNNs \cite{luo2024moderntcn,bai2018empirical,liu2022scinet,wang2022micn,franceschi2019unsupervised,sen2019think}, MLP-based networks \cite{TimeMixer,tang2025unlocking,chen2023tsmixer}, Transformer-based methods \cite{vaswani2017attention,nie2022time,ni2023basisformer,zhou2021informer,wu2021autoformer,liu2021pyraformer,zhang2022crossformer,cao2023inparformer,fu2024encoder,zhang2023temporal,cao2023tempo,Pathformer,LiuHZWWML24, liu2024timer,qiu2024duet}, and graph neural networks (GNNs) \cite{cini2023graph, sriramulu2023adaptive, jin2022multivariate}. Among Transformer-based methods, PatchTST \cite{nie2022time} and iTransformer \cite{LiuHZWWML24} have emerged as two of the most influential architectures for independent and dependent modeling, respectively. PatchTST introduces patching by treating a segment of local time points as a semantic vector and models each channel independently. In contrast, iTransformer reverses the traditional Transformer structure by embedding each time series as a variate token, using the attention mechanism to capture inter-variates relationships. Meanwhile, the approach of embedding multivariate time series into a single token for each time point, as adopted by models such as LogTrans \cite{li2019enhancing}, Pyraformer \cite{liu2021pyraformer}, Informer \cite{zhou2021informer}, and Autoformer \cite{wu2021autoformer}, has become less popular in recent years. We find that independent channel modeling is more suitable for capturing endogenous temporal relationships within each time series, while interaction modeling is better at capturing exogenous relationships between multivariate time series. These two aspects can be complemented and studied separated; however, there has been no effective investigation into combining them, as prior work has not recognized that interaction effects typically occur only over short time periods. To address this, we propose a masking mechanism. Finally, we discuss several techniques that are useful for time-based mapping networks. VH-NBEATS \cite{yang2025variational} incorporates pre-trained basis functions to capture hourly, daily, and weekly effects, which is particularly effective for long-term TSF and low-granularity data. Autoformer \cite{wu2021autoformer} advocates trend and seasonal decomposition with the moving average kernel, consistent with other methods such as DLinear \cite{zeng2023transformers}. However, the effectiveness of  their decomposition is based on the choice of kernel size, which is improved by our methods. In \cite{kim2021reversible}, standardization and centralization are introduced in the time domain to enhance robustness. TimeMixer \cite{TimeMixer} introduces a multi-resolution downsampling strategy to study different effects. However, the effectiveness of these techniques for mapping time-frequency features has not yet been studied. In this work, we develop similar techniques that are tailored for time-frequency representations.


In conclusion, time-based methods lack a global perspective of time series but facilitate the capture of fine-grained temporal relationships. In contrast, frequency-domain-based mapping offers a global view and supports the decomposition of various effects, but it loses important temporal details. Therefore, we introduce a time-frequency learning framework that leverages Fourier basis expansion to extract time-frequency features, effectively combining the strengths of both domains.


\begin{figure*}[t]
  \centering
  \includegraphics[width=0.9\linewidth]{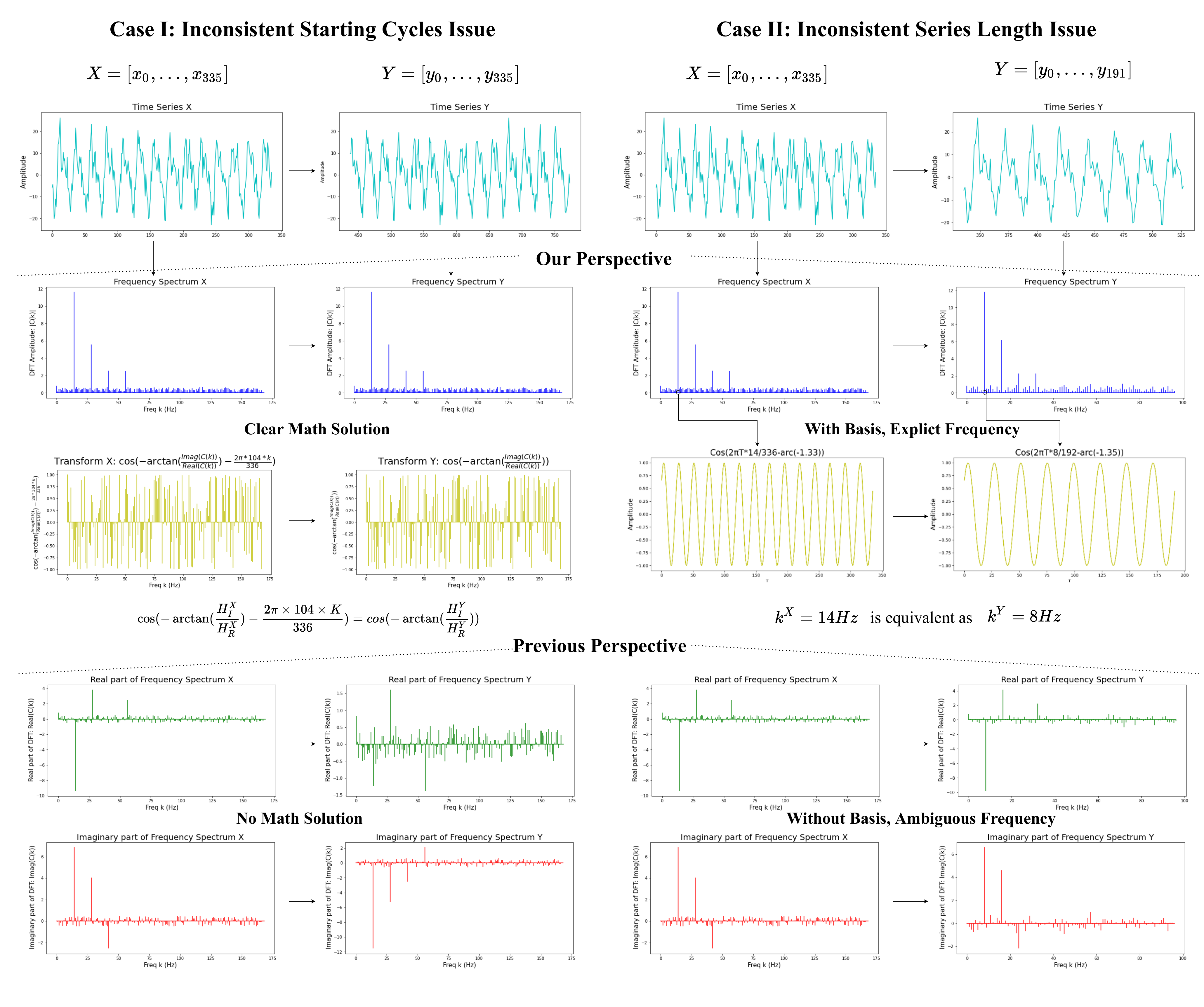}
  \caption{Two Issues of Existing Fourier-based TSF Models: Inconsistent Starting Cycles Issue and Inconsistent Series Length Issue. Two cases illustrate them. In Case I, $\mathbf{X}$ and $\mathbf{Y}$ have a starting cycle gap of 104 over 336. In Case II, $\mathbf{X}$ and $\mathbf{Y}$ have sequence lengths of 336 and 192, respectively.}
  \label{freq}
\end{figure*}

\section{Rethinking Fourier transform w.r.t. basis functions}
\label{section2.1}

In this section, we present a new perspective of the Fourier transform for TSF. First, we discuss the mathematical reasoning behind the DFT and IDFT in terms of basis functions. From this new basis perspective, we observe that the real and imaginary parts of the frequency components correspond to the coefficients of cosine and sine basis functions across different frequency levels. As such, we identify inconsistent starting cycles and inconsistent series length issues in existing studies.

Let $\mathbf{X}$ and $\mathbf{Y}$ represent the input and output time series, respectively, and $T$ and $L$ refer to the look-back window and forecast horizon, both assumed to be even numbers. $\mathbf{H}^{X}$ and $\mathbf{H}^{Y}$ denote the frequency spectrum of the input and output, respectively. Then, DFT and IDFT of the input time series $\mathbf{X}$ can be expressed as follows:
\begin{equation}
\begin{aligned}
\mathbf{H}(k)&=DFT(\mathbf{X})=\sum_{n=0}^{T-1} \mathbf{X}[n] \exp \left(-i \frac{2 \pi k n}{T} \right),\\
\mathbf{X}[n]&=IDFT(\mathbf{H})=\frac{1}{T}\sum_{k=0}^{T-1} \mathbf{H}[k] \exp \left(i \frac{2 \pi k n}{T} \right), \\
\quad & n=0,1, \ldots, T-1, \quad k=0,1, \ldots, T-1
\end{aligned}
\end{equation}

From the perspective of basis functions, IDFT can be expressed by $\frac{T}{2}+1$ orthogonal cosine basis functions and $\frac{T}{2}-1$ orthogonal sine basis functions. This is because the frequency components of a real-valued signal are Hermitian symmetric. The proof can be found in Appendix \ref{proof1}. Subsequently, we can rewrite the IDFT w.r.t. basis functions, and the connection between $\mathbf{X}$ and $\mathbf{H}^{X}$ can be expressed as follows:
\begin{equation}
\begin{aligned}
\mathbf{X}[n]&=\frac{1}{T}\sum_{k=0}^{\frac{T}{2}}\left(\mathbf{a_k} \cos \left(\frac{2 \pi k n}{T}\right)- \mathbf{b_k} \sin \left(\frac{2 \pi k n}{T}\right)\right), \\ & n= 0,1,\ldots T-1, \\ 
 \mathbf{a_k}&= \begin{cases} \mathbf{H_R}[k], &  \\ 2 \cdot \mathbf{H_R}[k], & \end{cases}   \mathbf{b_k}= \begin{cases}  \mathbf{H_I}[k], & k=0,\frac{T}{2} \\ 2 \cdot \mathbf{H_I}[k], & k=1, \ldots, \frac{T}{2}-1.\end{cases}   \\ 
\label{eq4}
\end{aligned}
\end{equation}

$\mathbf{H_R}[k]$ and $\mathbf{H_I}[k]$ represent the real and imaginary parts of $\mathbf{H}[k]$ respectively, where $k$ refers to the frequency level, and $\mathbf{H}[k]=\mathbf{H_R}[k]+i \mathbf{H_I}[k]$. Eq. (\ref{eq4}) provides an essential insight that the real and imaginary parts of the frequency spectrum can be interpreted as the coefficients of the cosine and sine basis functions, respectively. Thus, computing the frequency spectrum of the output time series is equivalent to computing the coefficients of cosine and sine basis functions, a process that aligns with the design of the N-BEATS \cite{oreshkin2019n}. 

Consequently, we highlight two issues in existing Fourier-based studies: inconsistent starting cycles and inconsistent series length issues. The inconsistent starting cycles issue arises because the real and imaginary parts only carry explicit meanings when  their corresponding basis functions are combined and fused. This is because adding a sine and cosine wave of the same frequency results in a phase-shifted cosine wave at that frequency with a fused amplitude, as below:
\begin{equation}
\begin{aligned}
\mathbf{Z}(t)&= A\cos(wt)+B\sin(wt)= R\cos(wt-\phi), \\ 
R &= \sqrt{A^2+B^2},\quad \phi = arctan(B,A).
\label{eq5}
\end{aligned}
\end{equation}
Therefore, the key information is embedded in the amplitude and arctangent of the real and imaginary values rather than the values themselves. The inconsistent series length issue arises because the definition of frequency in hertz (Hz) is bounded on series length. We present two cases, Cases I and II, to illustrate them in Fig. \ref{freq}, using manually generated time series by sine and cosine basis functions, under the assumption that time series $\mathbf{X}$ is used to forecast time series $\mathbf{Y}$.

In Case I, $\mathbf{X}$ and $\mathbf{Y}$ have the same frequency and series length but different starting cycles. When the real and imaginary components of the input frequency spectrum are processed independently to compute the output frequency spectrum, no mathematical solution exists to establish such a mapping. However, their relationships can be easily identified through the amplitude and arctangent of the real and imaginary values, which are embedded within basis functions. As shown in Fig. \ref{freq}, a mathematical solution exists which is achievable through Eq. (\ref{eq5}). 

In Case II, $\mathbf{X}$ and $\mathbf{Y}$ have the same frequency and starting cycles but different series lengths. Although the frequency spectrum landscapes appear similar, their components (in Hz) carry different meanings. For example, an 8 Hz cosine function with a series length of 192 is equivalent to a 14 Hz cosine function with a series length of 336. Unfortunately, current models do not account for this frequency relationship precisely. As a result, the model faces challenges in precisely interpreting frequency components when series lengths vary. The presence of time-dependent basis functions makes it even harder. Instead, these issues can be resolved by incorporating basis functions, which will be discussed in Section \ref{FBM}.

\section{FBM: Fourier Basis Mapping}
\label{FBM}

\subsection{Time-Frequency Features}

\begin{figure}[ht]
\centering
\includegraphics[width=\linewidth]{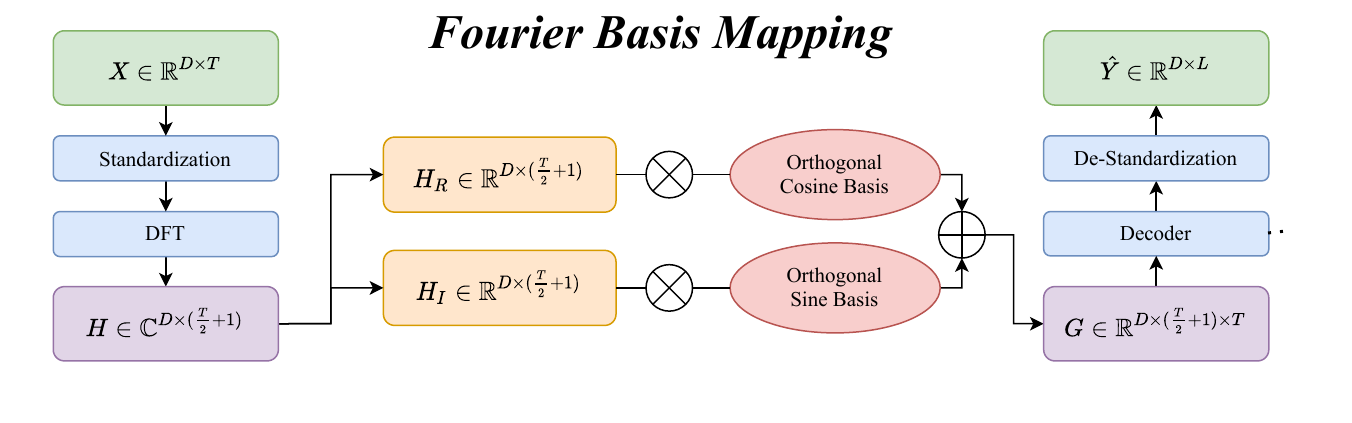}
\caption{Architecture of the Fourier Basis Mapping (FBM). }
\label{figure1}
\end{figure}

FBM addresses the two aforementioned issues. The general architecture is shown in Fig. \ref{figure1}. The primary strength of FBM lies in constructing time-frequency features that capture explicit frequency information while preserving temporal characteristics. Subsequently, the downstream mapping considers the time-frequency space rather than solely the time or frequency space for forecasting. Since the basis functions incorporate information from the time domain, the mapping issue mentioned previously in the frequency domain no longer exists. To obtain the time-frequency features, we multiply the real part of $\mathbf{H}$ (denoted as $\mathbf{H_R}$) with the orthogonal cosine basis $\mathbf{C}$ and the imaginary part of $\mathbf{H}$ (denoted as $\mathbf{H_I}$) with the orthogonal sine basis $\mathbf{S}$, then add them together. This process decomposes the time series into different frequency levels, while also accounting for phase shifts and amplitude merging at each level. Let $\mathbf{N}=[0,1,\ldots,T-1]$, then $\mathbf{C}$ and $\mathbf{S}$ can be expressed as follows:
\begin{equation}
\begin{aligned}
\mathbf{C}&=\frac{1}{T}[\mathbf{1},2\cos (\frac{2 \pi \mathbf{N}}{T}), \ldots , 2\cos (\frac{(T-1) \pi \mathbf{N}}{T}), \cos ( \pi \mathbf{N})],\\
\mathbf{S}&= -\frac{1}{T}[0, 2\sin (\frac{2 \pi \mathbf{N}}{T}), \ldots, 2\sin (\frac{(T-1) \pi \mathbf{N}}{T}), \sin (\pi \mathbf{N})]. 
 \end{aligned}
\end{equation}

\begin{figure*}[t]
  \centering
  \includegraphics[width=0.9\linewidth]{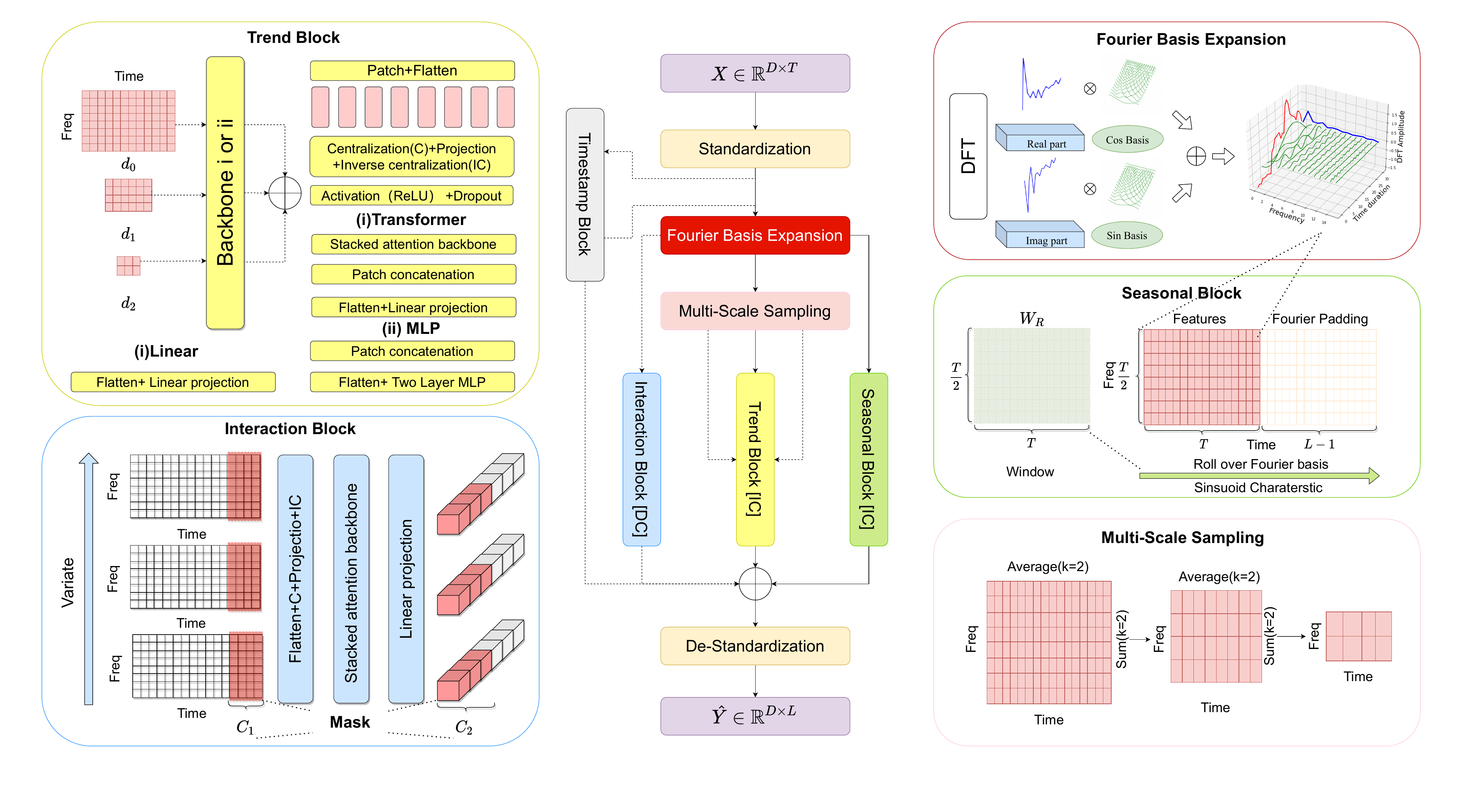}
  \caption{FBM-S: A Three-Block Architecture for Trend, Seasonality, and Interaction. In the seasonal block, a rolling window is applied to extract seasonal patterns. In the trend block, we allow the choice of linear, MLP-based, or Transformer-based architectures. In addition, we introduce techniques such as patching based on time segments of the time-frequency features, centralization and inverse centralization before and after the initial projection, and multi-scale down-sampling using average and sum kernels for time and frequency domain, respectively. In the interaction block, a masking mechanism is developed to consider the relationships between the masked input length $C_1$ and the masked output length $C_2$, as interaction effects typically occur over short periods. Centralization technique is  applied in the interaction block to improve the robustness of the extracted features. ID refers to independent channel modeling, while DC refers to dependent channel modeling. }
  \label{FBM_S}
  \vspace{-10pt}
\end{figure*}

\subsection{The Plug-and-Play Effects for FBM-L, FBM-NL, FBM-NP}

Since our time-frequency features preserve time-domain information, they can be applied to any method simply by adjusting the initial projection layer to map the time-frequency features into the hidden state. This projection considers both time and frequency modalities, which leads to improved performance. We first design three decoders to demonstrate the effectiveness of time-frequency features by plugging them into existing mapping methods for better performance: linear (L), non-linear three-layer perception (NL), and non-linear (NP). Consequently, three FBM variants are generated: FBM-L with linear network, FBM-NL with MLP network, and FBM-NP with Transformer-based network with patching. The first variant consists of a single vanilla linear layer and serves as a plug-and-play version of NLinear. While NLinear maps input time series features to output time series using a single layer, our approach maps time-frequency features to output time series using also a single layer. The second variant includes three fully connected layers with ReLU activation functions, representing a deeper version of NLinear. NLinear has shown that increasing the depth of a network by stacking MLP layers can lead to degraded performance. However, we prove that increasing the network depth tends to improve results when using time-frequency features. Thus, even a very simple one-layer or three-layer neural network can achieve strong performance by mapping in the time-frequency space. The last variant serves as a plug-and-play version of PatchTST, with the only difference lying in the initial projection. Specifically, PatchTST performs patching based on time segments and projects the patched  features into the hidden space. Similarly, we perform patching based on time segments of the time-frequency features, then flatten the patches and apply a projection. We also show that FBM-NP can outperform PatchTST with fewer patches and improved efficiency by effectively utilizing time-frequency features.


\subsection{FBM-S: A Synergetic Architecture}
\label{FBMS}
We propose a more efficient and effective approach for modeling time-frequency features: a synergetic FBM model, namely FBM-S, which captures trend, seasonality, and interaction effects through three distinct blocks. We consider different architectures for time-frequency features to study those effects, respectively. In Fig. \ref{FBM_S}, we provide the overall architecture. Since we perform standardization at the beginning, the first frequency level is removed, as mean is always zero. The timestamp block is directly copied from \cite{yang2025variational} and is used only for hourly long-term TSF datasets. 

\textbf{Seasonal Block}: In the seasonal block, the output features are expected to reflect seasonal characteristics. Therefore, using a rolling window filter is an optimal way to capture these patterns. This is because all the basis functions exhibit sinusoidal characteristics. Therefore, after generating the time-frequency features, we apply the Fourier padding to extend the Fourier basis with an additional length of $L - 1$. We then apply a rolling window with weights $\mathbf{W} \in \mathbb{R}^{T \times \frac{T}{2}}$ over the padded features to extract the seasonal components. $\hat{\mathbf{Y}}_S$ refers to the predicted output of the seasonal block, and the mathematical formula can be shown as follows: 
\begin{equation}
\begin{aligned}
\hat{\mathbf{Y}}_S[v] & =\frac{2}{T} \sum_{n=0}^{T-1} \sum_{k=1}^{\frac{\pi}{2}}\left(\mathbf{a}_{\mathbf{k}}\left(\mathbf{W}_{\mathbf{n}, \mathbf{k}} \cdot \cos \left(\frac{2 \pi k(n+v)}{T}\right)\right)\right. \\
& +\frac{2}{T} \sum_{n=0}^{T-1} \sum_{k=1}^{\frac{T}{2}}\left(\mathbf{b}_{\mathbf{k}}\left(\mathbf{W}_{\mathbf{n}, \mathbf{k}} \cdot \sin \left(-\frac{2 \pi k(n+v)}{T}\right)\right),\right. \\
 & \quad \quad \quad \quad \quad\quad\quad\quad\quad\quad\quad v =0,1, \ldots L-1, \\
&\hspace{-2em}\mathbf{a}_{\mathbf{k}} =\left\{\begin{array}{ll}
\frac{1}{2} \cdot\mathbf{H}_{\mathbf{R}}[k], \\
\mathbf{H}_{\mathbf{R}}[k],
\end{array}  \mathbf{b}_{\mathbf{k}}= \begin{cases} \frac{1}{2} \cdot\mathbf{H}_{\mathbf{I}}[k], \quad  k=0, \frac{T}{2}, \\
 \mathbf{H}_{\mathbf{I}}[k], \quad k=1, \ldots, \frac{T}{2}-1.\end{cases} \right.
\end{aligned}
\end{equation}
Here, we apply a small trick: we first let the weights multiply the padded Fourier basis functions, and then multiply the result with the real and imaginary values. This significantly improves both memory efficiency and backpropagation speed instead of using the default convolution in PyTorch.

\textbf{Interaction Block with Masking}: We consider channel interactions within the interaction block, which is particularly important in short-term TSF since interactions usually occur over short periods. For example, a traffic overload usually affects only nearby regions for a brief period; in the long run, the time series is primarily governed by its own trend and seasonal effects. Inspired by this observation, we utilize only the most recent time-frequency features for interaction inference, as indicated by the red mask in Fig. \ref{FBM_S}, corresponding to an input length of $C_1$. In addition, we also find that its influence does not last for a long period. Therefore, we use an additional output mask $C_2$ for long-term TSF. We also introduce centralization for segments of time-frequency features to improve the robustness of the extracted hidden representations, as it helps the model better understand whether a time series is in a normal, peak, or off-peak stage, which is crucial for interaction inference. The experimental results show that this interaction backbone significantly improves short-term TSF performance and slightly improves long-term TSF performance. 


\textbf{Trend Block}: We aim to capture non-linear trending effects, but we also retain the simple linear architecture as an option. To this end, we adopt either an MLP-based architecture with patching or a Transformer-based architecture with patching. This block integrates the strengths of FBM-L, FBM-NP, and FBM-NL. The PatchTST model has demonstrated that patching is an effective technique for time-domain modeling. Building upon this, we further show that patching is also beneficial for time-frequency features. In FBM-NP, we find that patching time-frequency features based on time segment is particularly beneficial for Transformer models, and thus we also apply this strategy to the MLP-based network. In FBM-NL, we used whole flattened time-frequency features as initial input, which increased the burden of initial projection. However, when patching is applied, the complexity of the initial projection layer can be reduced by a factor of $P^2$, where $P$ is the number of patches. It reduces the modeling complexity while also achieving better performance.

Thus, after patching, we perform the projection of the patched time-frequency features. Since we want to capture the trend effects here, we apply a non-linear activation function after the initial projection. This is always effective as adding the activation function in the initial projection increases the robustness of the extracted features even though the downstream backbones are nonlinear. Finally, we summarize the only difference between the MLP-based network and the Transformer-based network is that the MLP-based version removes the stack of attention layers and replaces it with a single projection layer with activation, as the final projection is always the same format. The results in Section \ref{Experiment} show that the MLP-based method consistently achieves better efficiency and performance. With time-frequency features, we can use a simpler downstream mapping network. Additionally, we introduce centralization and multi-scale down-sampling in the trend block to further enhance performance.

\textbf{Centralization}:
The implementation of centralization and inverse centralization techniques proves to be highly effective throughout the modeling process. This strategy enhances performance not only at the initial and final stages but also during intermediate phases, particularly before and after projecting patched time-frequency features within both the trend and interaction blocks. Although entire time-frequency features exhibit a zero mean due to the initial standardization, the mean of the flattened features within each patch deviates from zero. Consequently, applying centralization prior to projection and decentralization afterward normalizes their distributions, thereby improving model performance. In notation $x_{dpn}^{(i)}$, $d \in [1, \ldots, D]$ denotes variate index, $p \in [1, \ldots, P]$ indicates the patch index, and $n \in [1, \ldots, N]$ denotes the elements within the patched time-frequency features, and $i$ refers to the $i$-th layer. Specifically, $N= \frac{T^2}{2P}$ in the trend block, whereas in the seasonal block, $N = C_{1} \times \frac{T}{2}$ as only the last patch with mask $C_{1}$ will be involved in the modeling. The corresponding mathematical formulas are shown below: 
\begin{equation}
\begin{aligned}
&\mathbb{E}_t\left[x_{dp}^{(i)}\right]=\frac{1}{N} \sum_{n=1}^{N} x_{dpn}^{(i)}, \\
&\operatorname{Var}\left[x_{dp}^{(i)}\right]=\frac{1}{N} \sum_{n=1}^{N}\left(x_{dpn}^{(i)}-\mathbb{E}_t\left[x_{dp}^{(i)}\right]\right)^2, \\
&x_{dpn}^{(i)}=\gamma_d\left(\frac{x_{dpn}^{(i)}-\mathbb{E}_t\left[x_{dp}^{(i)}\right]}{\sqrt{\operatorname{Var}\left[x_{dp}^{(i)}\right]+\epsilon}}\right)+\beta_d, \\
&x_{dpn}^{(j)}=\sqrt{\operatorname{Var}\left[x_{dp}^{(i)}\right]+\epsilon} \cdot\left(\frac{x_{dpn}^{(j)}-\beta_d}{\gamma_d}\right)+\mathbb{E}_t\left[x_{dp}^{(i)}\right].
\end{aligned}
\end{equation}
where $i$ and $j$ is the initial and final layer within the trend and interaction blocks and $\gamma, \beta \in \mathbb{R}^{D}$ are learnable affine parameter vectors. Standardization refers to the case when the affine transformation is not applied.

\textbf{Multi-scale Down-sampling}: We also introduce a multi-scale mapping mechanism, inspired by the TimeMixer \cite{TimeMixer} and U-Net \cite{ronneberger2015u} architectures. Specifically, we downsample the time-frequency features to generate lower-resolution representations by averaging along the time dimension and summing across the frequency dimension, resulting in downsampled features $d_1$ and $d_2$ with kernel sizes of $2$ and $4$, respectively. This approach is particularly effective for handling high-granularity data, such as PEMS with five-minute intervals. The multi-scale features are processed through separate layers and aggregated subsequently.

\section{Preliminary Experimental Setup}
\label{Experiment}

\subsection{Data}
We conduct our experiments on twelve real-world datasets: ETT\footnote{https://github.com/zhouhaoyi/ETDataset} (ETTh1, ETTh2, ETTm1, ETTm2), Electricity (ECL) \footnote{\url{https://archive.ics.uci.edu/ml/datasets/ElectricityLoadDiagrams20112014}}, Traffic (TRA)\footnote{http://pems.dot.ca.gov}, Weather (WTH)\footnote{https://www.bgc-jena.mpg.de/wetter/}, Exchange rate (Exchange)\footnote{https://github.com/laiguokun/multivariate-time-series-data} and PEMS\footnote{https://www.kaggle.com/datasets/elmahy/pems-dataset} (PEMS04, PEMS08M, PEMS03, PEMS07) and M4\footnote{https://paperswithcode.com/dataset/m4}(yearly, quarterly, monthly, weekly, daily and hourly). The granularity of ETTh1, ETTh2, Electricity, and Traffic is at an hourly time scale, while a fifteen-minute time scale for ETTm1 and ETTm2,  a ten-minute time scale for Weather,  a daily time scale for Exchange, and a five-minute time scale for PEMS.
In dataset M4, participants are tasked with forecasting a fixed number of time steps: 6 for yearly series, 8 for quarterly, 18 for monthly, 13 for weekly, 14 for daily, and 48 for hourly series, respectively. It was created by selecting a random sample of 100,000 time series from the ForeDeCk database. The ETT datasets include seven oil and load characteristics of electricity transformers, which span from July 2016 to July 2018. Traffic comprises hourly road occupancy rates measured by 862 sensors in the San Francisco Bay area from 2015 to 2016. Electricity records the hourly electricity consumption (in kWh) of 321 clients from 2012 to 2014. Weather includes 21 weather indicators for 2020 in Germany, such as air temperature, humidity, and so on. Exchange tracks the daily exchange rates of eight countries from 1990 to 2016. PEMS contains the public traffic network data in California with four public subsets. 

\subsection{Baselines}
\label{baseline}
We compare FBM with eight baseline methods: NLinear, PatchTST, iTransformer, TimeMixer, N-BEATS, CrossGNN, FITS, and FreTS on eight datasets for long-term TSF (LTSF). Furthermore, we compare FBM with six baseline methods: the NLinear, PatchTST, TimeMixer, iTransformer, FiLM, and TimesNet on four PEMS datasets for short-term TSF (STSF). These methods are chosen because they represent six categories of modeling methods: (1) Linear method: NLinear; (2) Transformer-based methods: iTransformer and PatchTST; (3) MLP-based methods: N-BEATS, FreTS, and TimeMixer; (4) RNNs-based method: FiLM; (5) CNNs-based method: TimesNet; and (4) Fourier-based methods: N-BEATS, FITS, FreTS, CrossGNN, FiLM and TimesNet.

\subsection{Experiment Settings}

The mean squared error (MSE) and the mean absolute error (MAE) are used as evaluation metrics for both LTSF and STSF. The look-back window is set to $336$, and the forecast horizons are set to ${96, 192, 336, 720}$ for LTSF and ${12, 24, 48, 96}$ for STSF. We split the ETT dataset into 12/4/4 months and all the other datasets into training, validation, and test sets by the ratio of $6.5/1.5/2$. We use the same batch size for our proposed methods and baseline methods to ensure fair comparisons, where 128 for ETTh1, ETTm1, ETTh2, ETTh2, WTH, and Exchange; 64 for PEMS03, PEMS04, PEMS07, and PEMS08; and 16 for Electricity and Traffic. On the other hand, the experimental setup for the M4 dataset follows that of TimeMixer \cite{TimeMixer}. The input length of M4 is twice the forecast horizon. The hyperparameters of FBM-S are shown in Table \ref{hyperparameter}, and their meanings, along with the model architecture, are elaborated in Section \ref{complexity}. Since time series are decomposed into frequency levels, a smaller learning rate (LR) is required to learn the time-frequency features, and the 'OneCycle' LR optimization strategy is omitted for long-term TSF. For more implementation details, please refer to our project page. We use the same random seed and provide all implementation details and scripts on the project page to ensure the best reproducibility of the reported experiments for every dataset.

\begin{table}[ht]
\centering
\caption{Hyperparameter Details for Different Datasets. }
\begin{adjustbox}{width=\columnwidth,center}
\begin{tabular}{cccccccccccccccc} 
&  Trend & Seasonal & Interaction  & Patch &  Multi-Scale  & $P$ & $h_1$ &  $h_2$ &  $h_3$ &  $K$  & $C_1$  &$C_2$ & lr &Timestamp    \\ 
\midrule 
PEMS03 & MLP & $\checkmark$ &  $\checkmark$ & $\checkmark$ & $d_1$ & 14 & 256 & 1440 & 512  & 3 & 24 & L & 0.0005&  \ding{55}\\
PEMS04 & MLP & $\checkmark$ &  $\checkmark$ & $\checkmark$ & $d_1$ & 14 & 256 & 1440 & 512  & 3 & 24 & L & 0.0005 & \ding{55}\\
PEMS07 & MLP & $\checkmark$ &  $\checkmark$ & $\checkmark$ & $d_1$ & 14 & 256 & 1440 & 512  & 3 & 24 & L & 0.0005&\ding{55}\\
PEMS08 & MLP & $\checkmark$ &  $\checkmark$ & $\checkmark$ & $d_1$ & 14 & 256 & 1440 & 512  & 3 & 24 & L & 0.0005& \ding{55}\\
ECL & MLP & $\checkmark$ &  $\checkmark$ & $\checkmark$ & $d_0$ & 14 & 256 & 1440 & 512 & 3 & 24 & 24 & 0.0005 & $\checkmark$ \\
Traffic & Transformer & $\checkmark$ &  $\checkmark$ & $\checkmark$ & $d_0$ & 14 & 128 & 128 & 512 & 4 & T & L & 0.0001 & $\checkmark$ \\
WTH & MLP & $\checkmark$ &  $\checkmark$ & $\checkmark$ & $d_0+d_1$ & 14 & 256 & 1440 & 256 & 3 & 96 & 12& 0.00005 & \ding{55} \\
ETTm1 & MLP & $\checkmark$ &  $\checkmark$ & $\checkmark$ & $d_0+d_1+d_2$ & 14 & 128 & 1440 & 128  & 3 & 48 & 48 &  0.00004 &\ding{55}\\
ETTm2 & MLP & $\checkmark$ &  $\checkmark$ & $\checkmark$ & $d_0$ & 14 & 128 & 1440& 128  & 3  & 48 & 48 & 0.00004  & \ding{55}\\
ETTh1 & Linear & $\checkmark$ &  \ding{55} & \ding{55} & $d_0$ & - & - & - & -  & - & - & - & 0.00002&  $\checkmark$ \\
ETTh2 & Linear & $\checkmark$ &  \ding{55} & \ding{55} & $d_0$ & - & - & - & -  & - & - & -  & 0.00001 & $\checkmark$ \\
Exchange & Linear & $\checkmark$ &  \ding{55} & \ding{55} & $d_0$ & - & - & - & -  & - & - & - & 0.00002&\ding{55} \\
M4 & MLP & $\checkmark$ &  \ding{55} & \ding{55} & $d_0$ & - & 1440 & 1440 & -  & - & - & - & 0.0001& \ding{55}\\
\midrule 
\end{tabular}
\end{adjustbox}
\label{hyperparameter}
\end{table}

\begin{table*}[t]
\centering
\caption{Performance of FBM-S, FBM-L, FBM-NL, and FBM-NP,  Compared to Eight Baseline Methods on Eight Datasets for Long-term TSF. }
\begin{adjustbox}{width=2\columnwidth,center}
\begin{tabular}{cccccccccc|cccccccccccccccccc} 
\midrule
\multicolumn{1}{c}{Method}&  & \multicolumn{2}{c}{FBM-S}  & \multicolumn{2}{c}{FBM-L}    & \multicolumn{2}{c}{FBM-NL} & \multicolumn{2}{c|}{FBM-NP}            & \multicolumn{2}{c}{NLinear \cite{zeng2023transformers}}                   & \multicolumn{2}{c}{PatchTST \cite{nie2022time}}   & \multicolumn{2}{c}{iTransformer \cite{LiuHZWWML24}} & \multicolumn{2}{c}{TimeMixer \cite{TimeMixer}}& \multicolumn{2}{c}{N-BEATS \cite{oreshkin2019n}} & \multicolumn{2}{c}{CrossGNN \cite{huang2023crossgnn}}& \multicolumn{2}{c}{FITS \cite{xu2023fits}}& \multicolumn{2}{c}{FreTS \cite{yi2023frequency}} \\ 
\midrule
\multicolumn{1}{c}{Error}&  & MSE & MAE & MSE & MAE  & MSE                    & MAE                    & MSE   & MAE                 & MSE   & MAE                   & MSE   & MAE  & MSE   & MAE       & MSE   & MAE  & MSE   & MAE    & MSE   & MAE    & MSE   & MAE& MSE   & MAE      \\ 
\midrule
\multirow{4}{*}{ETTh1}  & 96 & \textcolor{red}{0.363}& \textcolor{red}{0.389} & \underline{0.366}                & \underline{0.390} &0.368&0.395& 0.367& 0.395      & 0.391 & 0.416 & 0.374& 0.399 &0.399 & 0.417 & 0.385& 0.408  & 0.387& 0.410 & 0.376 &  0.400    &  0.368 & 0.392 & 0.404 &  0.423    

\\[5pt] 

                         & 192 & \textcolor{red}{0.399} & \textcolor{red}{0.409} &\underline{0.403} & \underline{0.411}  &0.408&0.418&0.407&0.416         & 0.421 &0.426& 0.417 & 0.422     &   0.436 & 0.440 & 0.429 &0.432       & 0.428 & 0.434  & 0.419   & 0.427& 0.404 & 0.412 & 0.461& 0.460         \\ [5pt]

                         & 336 &\textcolor{red}{0.402} &\textcolor{red}{ 0.413}& \underline{0.418} & \underline{0.420} & 0.425& 0.430& 0.433& 0.438   & 0.435                  & 0.435               & 0.431 & 0.436    &    0.446& 0.451& 0.456& 0.450   & 0.448 & 0.447                &  0.439 & 0.442 & 0.419 & 0.435 &0.488 & 0.480              \\[5pt] 

                         & 720&\textcolor{red}{0.403} &\textcolor{red}{0.433} & \underline{0.414} & \underline{0.438} &0.456& 0.466&0.439& 0.459 &   0.443                  & 0.457                   & 0.445 & 0.463    & 0.502 & 0.503&0.457 &0.462       & 0.466 & 0.471            & 0.447 & 0.465        & 0.431 & 0.458 & 0.566& 0.553     \\ [5pt]
\midrule
\multirow{4}{*}{ETTh2}   & 96 &\textcolor{red}{ 0.271}& \textcolor{red}{0.331} & \underline{0.271} & \underline{0.331}&0.287&0.343   &0.280&0.340    &  0.283             &  0.342                & 0.276 & 0.338          &0.303&0.362&0.276&0.339     & 0.303 & 0.363            & 0.283  &0.344 & 0.276 & 0.338 & 0.327& 0.388              \\ [5pt]

                         & 192 &\textcolor{red}{0.332} & \textcolor{red}{0.373} &\underline{0.332}                  &\underline{ 0.373} &0.351&0.386 &0.342&0.382               & 0.350& 0.387 &  0.341 &  0.378         &0.372& 0.403 &0.340&  0.381     & 0.364 & 0.402     &  0.342&      0.387        & 0.336 & 0.377 & 0.428& 0.450           \\[5pt] 

                         & 336 &\textcolor{red}{ 0.320}& \textcolor{red}{0.376}& \underline{0.321} & \underline{0.376}&     0.352 &0.394&0.354&0.401  & 0.344                  & 0.395      & 0.332 & 0.385    &0.401&0.424 & 0.362&0.404          & 0.360 & 0.407                  & 0.361 &     0.408         & 0.324 & 0.379 & 0.499& 0.497               \\ [5pt]

                         & 720& \textcolor{red}{0.361} & \textcolor{red}{0.408} &\underline{ 0.369} & \underline{0.412} &0.397&0.432  &0.386&0.424  & 0.395                  &  0.436                 & 0.379 & 0.420    & 0.420&0.446  &0.398&0.433           & 0.428 & 0.465               &0.423 &     0.460         & 0.373 & 0.416 & 0.727  & 0.637        \\ [5pt]
\midrule
\multirow{4}{*}{ETTm1} & 96&\textcolor{red}{0.278}& \textcolor{red}{0.332} &  0.301& 0.343 & \underline{ 0.286} & \underline{ 0.339}      & 0.293 & 0.346          &  0.307     & 0.349 &  0.295 & 0.344  & 0.309 &  0.361 &  0.303 & 0.350        & 0.324&     0.367     & 0.300 &   0.343  & 0.305 & 0.347  &0.326  & 0.373               \\[5pt] 

                         & 192 & \textcolor{red}{0.317} & \textcolor{red}{0.358}& 0.337& \underline{0.364} & \underline{0.324} & 0.365   & 0.334 &  0.368 &  0.347     &  0.374 &  0.333  & 0.370       & 0.345  & 0.383 & 0.356 & 0.385      & 0.363 &   0.388       & 0.335 &  0.369          & 0.338 & 0.366  & 0.359  & 0.392             \\[5pt] 

                         & 336& \textcolor{red}{0.356} & \textcolor{red}{0.382} & 0.371 & \underline{0.384} &\underline{0.359} &  0.385  & 0.371 & 0.389  &   0.377     & 0.390 &   0.363 &  0.394      & 0.380 & 0.401 & 0.366 & 0.392      & 0.400 & 0.408         & 0.375 &   0.390    &  0.372 &  0.386 & 0.389 & 0.408       \\[5pt] 

                             & 720&  \textcolor{red}{0.412} & \underline{0.418} & 0.425& \textcolor{red}{0.415}& 0.422 & 0.424 &  0.426 &  0.420 &     0.436    &  0.425  & \underline{ 0.421 } & 0.420       & 0.448 & 0.442& 0.435 & 0.434      & 0.468  & 0.448         & 0.429 & 0.420        & 0.427 & 0.416 & 0.445 & 0.441         \\[5pt]      
\midrule
\multirow{4}{*}{ETTm2} & 96& \textcolor{red}{0.164} & \textcolor{red}{0.252}  & \underline{0.164 } &  \underline{0.252}    &   0.165&  0.254  & 0.167 & 0.258           &      0.169 & 0.259 & 0.173 &   0.261     &  0.180&  0.272& 0.174 & 0.258      & 0.168 &   0.259       &  0.164 &  0.252          & 0.167 & 0.256  &0.202 &0.288               \\[5pt] 

                         & 192& \underline{0.219} & \underline{0.290}  & \textcolor{red}{0.219} & \textcolor{red}{0.290}  & 0.225 &  0.296 & 0.224 &   0.296           &  0.223      & 0.294 & 0.255 & 0.306     & 0.239 &0.311  & 0.238 & 0.300        &  0.225& 0.301         & 0.220  & 0.294            &   0.222 &    0.293& 0.250 & 0.322                    \\[5pt] 

                         & 336& \underline{0.273} &\underline{0.326} & \textcolor{red}{0.271}  &\textcolor{red}{0.325 }    &  0.276 & 0.331  & 0.277 & 0.331          &    0.277   & 0.331 & 0.285 & 0.336        & 0.389 & 0.341& 0.272  & 0.327       & 0.282 &  0.336        & 0.276  & 0.330           & 0.277 &  0.329  & 0.328 & 0.368                   \\[5pt] 

                         & 720& \underline{0.365}& \underline{0.382} & \textcolor{red}{0.364}  &  \textcolor{red}{0.381 }          &  0.365 & 0.386 & 0.367 & 0.386   &   0.371    & 0.387 & 0.365 &  0.386       & 0.374  & 0.392 & 0.368 & 0.389     & 0.376 &   0.394       &  0.372 & 0.390            & 0.366 &  0.382  & 0.431 & 0.436                 \\[5pt]

\midrule
\multirow{4}{*}{Electricity}   & 96& \textcolor{red}{0.127}& \textcolor{red}{0.220}  & 0.142 & 0.237& \underline{0.132} & \underline{0.227} & 0.133& 0.227   &  0.143   & 0.239 & 0.133 &    0.227    &0.137   &0.232  &  0.134  &0.230     & 0.144 &   0.240        &  0.147      & 0.246 & 0.145 &0.242  & 0.145 & 0.245              \\ [5pt]

                         & 192 & \textcolor{red}{0.144}& \textcolor{red}{0.237} & 0.155 & 0.248 &\underline{ 0.149} &0.243 & 0.149 &\underline{0.242}  &    0.157     & 0.250 & 0.151 & 0.244      &  0.156 & 0.249  & 0.153 &0.245     &  0.158 &  0.252       &  0.161 & 0.258         & 0.158  & 0.253  & 0.158 &  0.255    \\ [5pt]

                         &  336& \textcolor{red}{0.161} & \textcolor{red}{0.254}&   0.172            &  0.265      &  \underline{0.167} & \underline{0.261} & 0.167 &0.261      &    0.174    & 0.267 & 0.167 & 0.261     &  0.171 & 0.266 & 0.172 &0.267         & 0.175 & 0.269         & 0.178 & 0.274        &  0.174& 0.269 & 0.178 & 0.275       \\ [5pt]

                         & 720 & \textcolor{red}{0.195} & \textcolor{red}{0.285} &      0.212       &  0.297        &  0.207 & 0.295 & 0.208 & 0.295 &  0.214     & 0.299 & 0.210 &  0.297    & \underline{0.195}  & \underline{0.288}  & 0.212 & 0.298      & 0.217 & 0.304       &  0.214 & 0.299      & 0.213 & 0.301 & 0.220 & 0.315 \\[5pt] 

\midrule
\multirow{4}{*}{Traffic} & 96 &\textcolor{red}{0.357} & \textcolor{red}{0.246} & 0.421 & 0.281     &  0.384& 0.264 & \underline{0.373} &  \underline{  0.253}     &  0.425  & 0.288& 0.381 & 0.257    & 0.376 & 0.263 & 0.381 & 0.261         & 0.429 & 0.295       & 0.428  & 0.291        & 0.421&  0.282 & 0.434 & 0.313                \\[5pt] 

                         & 192 & \textcolor{red}{0.382} & \textcolor{red}{0.257} & 0.434 & 0.286 & 0.399 & 0.269&  \underline{0.396}& \underline{0.266}   & 0.438      & 0.291 & 0.402 &  0.270     & 0.396 & 0.274 & 0.408  & 0.273      & 0.441 & 0.299        & 0.441 & 0.295    & 0.435 & 0.288  & 0.471  & 0.311         \\[5pt] 

                         & 336& \textcolor{red}{0.393}& \textcolor{red}{0.263} & 0.447& 0.292 & 0.419 & 0.282 &0.411& \underline{0.276}  &  0.452      & 0.300 & 0.422 &  0.283  &  \underline{0.407}& 0.283& 0.434 & 0.297    & 0.455&  0.307      & 0.455 & 0.302    & 0.448&  0.293   & 0.493 & 0.321            \\[5pt] 

                         & 720& \textcolor{red}{0.430} & \textcolor{red}{0.285} & 0.477 & 0.309 & 0.448 & 0.297  & \underline{0.442} & \underline{0.291}  & 0.482      & 0.317 & 0.454 &  0.296  & 0.449 & 0.305 & 0.469 & 0.319        & 0.486 &  0.326       & 0.486  & 0.318           & 0.478 & 0.310 &  0.535& 0.339             \\[5pt] 

\midrule
\multirow{4}{*}{Weather} & 96 & \textcolor{red}{0.147} & \textcolor{red}{0.196} &0.159 & 0.207  & 0.152 &  0.199       &0.156 & 0.204        &  0.176     & 0.226& 0.156& 0.206   & 0.162 & 0.211 & 0.158 & 0.204     & 0.186 &  0.238       & 0.163&   0.227     &  \underline{0.149} & \underline{0.198}   & 0.159 & 0.218             \\[5pt] 

                         & 192& \textcolor{red}{0.189} & \textcolor{red}{0.238} & 0.203&0.247& \underline{0.194} &\underline{0.242}  & 0.198 & 0.245  & 0.220     & 0.262 & 0.200 & 0.246    & 0.204 & 0.249 & 0.197 & 0.246        & 0.227 & 0.275         & 0.205 &0.261   & 0.196  & 0.244   & 0.207   & 0.270           \\[5pt] 

                         & 336 & \textcolor{red}{0.238}& \textcolor{red}{0.276}&  0.252 & 0.285 & 0.244 &  0.282  & 0.248 & 0.285   & 0.265       & 0.296 & 0.252 &   0.285  & 0.248 & 0.285 &  \underline{0.242} &  \underline{0.281}   & 0.274  &  0.307       & 0.250 &  0.295    & 0.245 &   0.283  & 0.252  & 0.299             \\[5pt] 

                         & 720& \textcolor{red}{0.311} & \textcolor{red}{0.328}&  0.319& 0.335& \underline{0.317} & \underline{0.334}  & 0.319 & 0.337 &   0.332      &  0.345  & 0.321  &  0.336  & 0.322 & 0.335 &  0.319 &  0.335     & 0.342 & 0.361          & 0.320 & 0.347         & 0.321 & 0.338 & 0.319 & 0.342              \\[5pt] 

\midrule
\multirow{4}{*}{Exchange} & 96 &  \textcolor{red}{0.093}& \underline{0.211} &\underline{0.093 }&  0.211     & 0.104 & 0.226 & 0.096& 0.196  & 0.098   &  0.219&0.104 & 0.227  & 0.128 & 0.254& 0.119 & 0.247  & 0.147 &  0.274       &  0.093&    \textcolor{red}{ 0.211}    &  0.109 & 0.235   & 0.209 & 0.350           \\[5pt] 

                         & 192 & \underline{0.194} & \underline{0.308}& 0.195 & 0.309& 0.210 & 0.326& 0.196 & 0.312  & 0.203     & 0.316  & 0.210 &  0.325    & 0.241 & 0.353& 0.238  &  0.354    & 0.312 &  0.406    & \textcolor{red}{0.188} &  \textcolor{red}{ 0.305} & 0.229  & 0.350   & 0.346  & 0.437         \\[5pt] 

                         & 336 &\textcolor{red}{0.346} & \textcolor{red}{0.419}& \underline{0.347} & \underline{0.421} &  0.398 & 0.460 & 0.353 & 0.425  & 0.356       & 0.426 &  0.366 &   0.435   & 0.393  & 0.459& 0.417 & 0.472  & 0.522 &   0.532      & 0.363 &   0.430   & 0.400& 0.463 & 0.634   & 0.583             \\[5pt] 

                         & 720& \underline{0.963}& \underline{0.732}& 0.965 & 0.732 & 1.040  & 0.762 &  0.970&  0.734& 0.965     & 0.733  & 1.026  & 0.757   & 1.00  & 0.763& 1.074 &   0.790      & 1.412  &   0.907        &  \textcolor{red}{ 0.931}&     \textcolor{red}{ 0.722}      & 1.095& 0.781 & 2.418   & 1.233            \\[5pt] 
     \midrule        
              \multicolumn{1}{c}{Average} & & \textcolor{red}{0.309} & \textcolor{red}{0.332} & 0.323   & \underline{0.339}& 0.325 & 0.344 & \underline{0.321}  & 0.340 &  0.333&  0.349& 0.326     & 0.344  & 0.341  & 0.353   & 0.335  & 0.352& 0.366 &   0.371      & 0.330  &   0.350        &  0.331&      0.347      & 0.431& 0.406         \\  [5pt]  
\hline
\end{tabular}
\end{adjustbox}
\label{table1}
\end{table*}

\section{Experiment Results}
\label{Main_reults}

We evaluate the forecasting performance of four FBM variants against diverse baseline models for both Long-term TSF (LTSF) and short-term TSF (STSF). These baselines cover a wide range of architectures, including time- and frequency-based mapping methods. Table \ref{table1} presents results for LTSF, while Table \ref{table_short} and Table \ref{table3} display results for STSF. In our experiments, we first demonstrate the effectiveness of time-frequency features by proposing three FBM variants: FBM-L, FBM-NL, and FBM-NP, which can serve as plug-and-play modules within existing methods. We then demonstrate that our proposed FBM-S model can achieve SOTA performance across nearly all datasets for both LTSF and STSF tasks through our three specialized blocks. In particular, we perform an efficiency analysis and elaborate on the structure in seasonal, trend, and interaction blocks in Section \ref{complexity}. The effectiveness of each technique for time-frequency features is thoroughly validated through ablation studies in Section \ref{Ablation}. We provide visualizations to illustrate the role of rolling windows in Section \ref{rolling}, and two case studies to demonstrate the synergetic effects between three specialized blocks in Section \ref{case_study}. Finally, we provide interpretable experiment results based on the characteristics of the data through the distributions of the input frequency spectrum in Section \ref{Distribution}.



\subsection{Long-term TSF Results}
\label{section4}

\textbf{FBM-L vs Linear Network}: The FBM-L model demonstrates superior performance over the NLinear model across all datasets and prediction horizons. Specifically, it reduces the average MSE from $0.333$ to $0.323$ and the MAE from $0.349$ to $0.339$. This enhancement underscores the effectiveness of FBM-L, which employs Fourier basis expansion to decompose temporal data based on frequency components. By operating in the time-frequency space, FBM can better distinguish noises from meaningful effects, leading to improved forecast accuracy. The gains are particularly notable in datasets such as ETTh1 and ETTh2, which contain richer frequency components and higher levels of noises. Furthermore, our experiments reveal that even a simple vanilla linear network FBM-L has the potential to outperform SOTA DNN-based architectures.

\textbf{FBM-NL vs MLP-based Methods and FBM-NP VS Transformer-based Methods}:  The results in \cite{zeng2023transformers} show that increasing the depth of the network does not improve performance for time-based method. In contrast, FBM-NL demonstrates that increasing the depth of the network is beneficial for most datasets when time-frequency features are incorporated. This highlights the potential of the time-frequency learning framework by considering both time and frequency modality. Additionally, FBM-NL performs better than TimeMixer and FBM-NP performs better than PatchTST at most of the time, where the former two are MLP-based networks, and the latter two are Transformer-based networks with patching. The experimental results further demonstrate the effectiveness of time-frequency features, suggesting that extracting time-frequency features efficiently is more advantageous than relying solely on deeper architectures. It is worth noting that TimeMixer also decomposes the original time series into trend and seasonal effects, but its performance largely depends on the choice of the moving average kernel size or Top-k frequencies. In contrast, our Fourier basis expansion hierarchically decomposes various effects across distinct frequency levels, enabling the downstream mapping network to consider their relative importance. Although FBM-NP uses fewer patches than PatchTST, it even achieves better performance. Our FBM variants also consistently outperform other Transformer-based and MLP-based architectures, including iTransformer and N-BEATS. 

\textbf{FBM Variants vs. Fourier-based Methods}: We also observe that FBM variants make a significant improvement over all Fourier-based methods, including N-BEATS, FreTS, CrossGNN, and FITS. This discrepancy may largely be attributed to the inconsistent starting cycles and series length issues discussed in Section \ref{section2.1}. Notably, FITS and CrossGNN also emphasize the importance of the amplitude of real and imaginary values. However, they overlook the fact that Fourier basis functions are time-dependent when the input length is not divisible by a certain frequency level. Consequently, mapping in the frequency domain disregards time-domain characteristics and fails to capture fine-grained relationships. For instance, CrossGNN retains only the values from the first cycle but ignores the variations in subsequent cycles for time-dependent basis functions, as well as the different phase shifts across frequency levels. In contrast, FBM variants offer a more effective representation of time-frequency feature and mapping within the time-frequency space. 

\begin{table*}[t]
\centering
\caption{Performance of FBM-S, FBM-L, FBM-NL, and FBM-NP, Compared to Six Baseline Methods on the PEMS Datasets for Short-term TSF.}
\begin{adjustbox}{width=2\columnwidth,center}
\begin{tabular}{cccccccccccccccccccccc} 
\midrule
\multicolumn{1}{c}{Method}& &\multicolumn{2}{c}{FBM-S} & \multicolumn{2}{c}{FBM-L}                 & \multicolumn{2}{c}{FBM-NL}                   & \multicolumn{2}{c}{FBM-NP} & \multicolumn{2}{c}{NLinear \cite{zeng2023transformers}}& \multicolumn{2}{c}{PatchTST \cite{nie2022time}} & \multicolumn{2}{c}{TimeMixer \cite{TimeMixer}}& \multicolumn{2}{c}{iTransformer \cite{LiuHZWWML24}}& \multicolumn{2}{c}{FiLM \cite{zhou2022film}}& \multicolumn{2}{c}{TimesNet \cite{wu2022timesnet}} \\ 
\midrule
\multicolumn{1}{c}{Error}& & MSE & MAE  & MSE & MAE  & MSE                    & MAE                    & MSE   & MAE                 & MSE   & MAE                   & MSE   & MAE  & MSE   & MAE       & MSE   & MAE  & MAPE   & MAE   & MSE   & MAE      \\ 

\midrule
                        
\multirow{4}{*}{PEMS03}  & 12& \textcolor{red}{ 0.057}  & \textcolor{red}{0.156}  & 0.077  &  0.183                & \underline{0.060}   & \underline{ 0.161}   &  0.061 &  0.162  & 0.078 & 0.184 &  0.065 & 0.174  & 0.062 & 0.164 & 0.061 & 0.163  &0.108  & 0.223 &  0.089& 0.199 \\[5pt] 

                         & 24&\textcolor{red}{ 0.069}& \textcolor{red}{0.170} &0.111 &0.210 &\underline{ 0.071}  & \underline{0.171}& 0.073 & 0.174 & 0.114 & 0.214& 0.075 & 0.179& 0.074& 0.182 &0.080 & 0.189 & 0.179 & 0.274& 0.096 & 0.204\\ [5pt]
                        
                         & 48& \textcolor{red}{0.085} & \textcolor{red}{0.188} & 0.159 & 0.243 &0.094 &\underline{ 0.194} &0.101 & 0.201 &0.162 & 0.248& 0.101& 0.208 & \underline{0.092}& 0.201&0.103 & 0.213 & 0.235 &0.321 & 0.115 &0.220\\ [5pt]
                        
                         & 96&\textcolor{red}{ 0.103} & \textcolor{red}{0.206} & 0.195 & 0.267 & 0.118& \underline{0.215} & 0.126 & 0.222 & 0.198 & 0.273& 0.139& 0.240& \underline{0.105}& 0.216 &0.123 & 0.229 & 0.201 & 0.299& 0.118 & 0.229\\ [5pt]

\midrule
\multirow{4}{*}{PEMS04} & 12&\textcolor{red}{ 0.062}&\textcolor{red}{0.157}&  0.088  &  0.195                & \underline{0.071}   &  0.172 & 0.071 &\underline{0.170} & 0.089  & 0.196 & 0.075&  0.179&  0.071 & 0.173 & 0.073& 0.176 & 0.118 & 0.237 & 0.091 & 0.196    \\[5pt] 

                       & 24 & \textcolor{red}{0.072} & \textcolor{red}{0.168}& 0.120 & 0.223 & 0.080   &   0.179& 0.084 &0.182               & 0.123 & 0.228    & 0.085  & 0.190  &\underline{0.073}  & \underline{0.174}  & 0.086 & 0.189 & 0.177  &0.278  &  0.090 & 0.190 \\ [5pt]

                         & 48 & \textcolor{red}{0.088} & \textcolor{red}{0.182} & 0.167 & 0.257   &   0.099             & 0.194           &   0.107    &  0.204 &  0.171  & 0.265  &  0.108 & 0.211 & \underline{0.089} & \underline{0.189}  &0.107  & 0.209 & 0.241 & 0.325 & 0.094& 0.193 \\ [5pt]
                         
                         & 96 & \textcolor{red}{0.104} & \textcolor{red}{0.196} & 0.207 & 0.282  &    0.125            &   0.216        &  0.135             & 0.224 &  0.210  &  0.287  & 0.130  & 0.229 & \underline{0.107}  & \underline{0.211} & 0.127 & 0.227  & 0.207 & 0.295  & 0.112 & 0.215\\ [5pt]

\midrule
                   \multirow{4}{*}{PEMS07}  & 12&\textcolor{red}{0.049}& \textcolor{red}{0.137} & 0.073 & 0.180    &   0.053        &\underline{0.148}   & 0.054  &   0.150          &  0.073 &  0.180 &  0.057 & 0.164   & 0.053 & 0.151  & \underline{ 0.053} & \underline{ 0.148} &0.101  &0.221  &0.079 &0.182  \\ [5pt]
                   
                         & 24& \textcolor{red}{0.056}& \textcolor{red}{0.145} & 0.107 & 0.212 & 0.063 & 0.160 & 0.062 & 0.157 & 0.109 & 0.216 & 0.065& 0.172& \underline{0.061} & \underline{0.158} & 0.069 & 0.171 & 0.199& 0.299& 0.080& 0.177\\ [5pt]
                        
                         & 48&\textcolor{red}{0.064} &\textcolor{red}{ 0.155} & 0.157 & 0.251& 0.074 & \underline{0.168} &0.079 & 0.175 & 0.160 & 0.255 & 0.079 & 0.189 & \underline{0.072} & 0.172 & 0.077 & 0.177 & 0.238 & 0.331 & 0.084 & 0.183\\ [5pt]
                        
                         & 96& \textcolor{red}{0.072} & \textcolor{red}{0.162} & 0.197 & 0.279 & 0.090& \underline{0.184} & 0.093 & 0.186 & 0.200 & 0.285 &  0.093& 0.199 & 0.091 & 0.199 & \underline{0.087} & 0.190 & 0.190 & 0.281 &0.089 & 0.188\\ [5pt]

\midrule
\multirow{4}{*}{PEMS08}

                         & 12&\textcolor{red}{0.055} &\textcolor{red}{0.147} & 0.081 & 0.189                &  \underline{ 0.060}               &\underline{ 0.159}            &  0.061 & 0.159             & 0.081  &  0.190  & 0.062  & 0.165   & 0.060 & 0.160 & 0.062 & 0.165 & 0.108 & 0.226 & 0.094 & 0.204 \\ [5pt]

                         & 24& \textcolor{red}{0.064}  & \textcolor{red}{0.157} & 0.115 & 0.211        &     0.069           &  0.165          & 0.069              & 0.168 &  0.118  & 0.227  & 0.072   & 0.178 & 0.068 &  0.170 & \underline{0.066} & \underline{0.160} &0.182  & 0.285 & 0.097  & 0.198\\ [5pt]

                          & 48&\textcolor{red}{ 0.073} & \textcolor{red}{0.167} & 0.173 & 0.264                &     0.084           & \underline{0.179}           &  0.085  & 0.183 &  0.180  & 0.271  & 0.088   & 0.196 & \underline{0.080}  & 0.183  & 0.090 & 0.195  & 0.261 & 0.341 & 0.102  & 0.204\\ [5pt]
                         
                          & 96& \textcolor{red}{0.083 }& \textcolor{red}{0.177} & 0.228 & 0.298                &         0.102       &  0.195          & 0.103              & 0.196 & 0.234   & 0.303  & 0.105  &  0.207 & 0.098 &  0.201& \underline{0.091} & \underline{0.183}  &0.233  & 0.302  &0.121   &0.226\\ [5pt]
     \midrule 
                         Average & & \textcolor{red}{0.072} & \textcolor{red}{0.166} & 0.140 & 0.234  &0.082 &\underline{0.178} & 0.085  & 0.182 & 0.143 & 0.238 & 0.087 &  0.192 & \underline{0.078}& 0.181 & 0.084 & 0.186& 0.186 &  0.283& 0.096&0.200 \\ [5pt]
                        


\hline
\end{tabular}
\end{adjustbox}
\label{table_short}
\end{table*}

\begin{table*}[t]
\centering
\caption{Performance of FBM-S, Compared to Thirteen Baseline Methods on the M4 Dataset for Univariate TSF. The sampling frequencies and forecast horizons ranging from 6 to 48. A lower SMAPE, MASE or OWA indicates a better prediction.}
\begin{adjustbox}{width=\columnwidth*2,center}
\begin{tabular}{cccccccccccccccccc}
\midrule
\multicolumn{2}{c}{\multirow{2}{*}{\textbf{Models}}} & \multirow{2}{*}{\textbf{FBM-S}} & \multirow{2}{*}{\textbf{TimeMixer \cite{TimeMixer}}} &\multirow{2}{*}{\textbf{TimesNet \cite{wu2022timesnet}}} & \multirow{2}{*}{\textbf{N-HiTS \cite{challu2023nhits} }} & \multirow{2}{*}{\textbf{N-BEATS \cite{oreshkin2019n}}} &\multirow{2}{*}{\textbf{SCINet \cite{liu2022scinet}}} & \multirow{2}{*}{\textbf{PatchTST \cite{nie2022time}}} &\multirow{2}{*}{\textbf{MICN \cite{wang2022micn}}} &\multirow{2}{*}{\textbf{FiLM \cite{zhou2022film}}} &\multirow{2}{*}{\textbf{LightTS \cite{zhang2207less}}} &\multirow{2}{*}{\textbf{DLinear \cite{zeng2023transformers}}} &\multirow{2}{*}{\textbf{FEDformer \cite{zhou2022fedformer}}} &\multirow{2}{*}{\textbf{Stationary \cite{liu2022non}}} &\multirow{2}{*}{\textbf{Autoformer \cite{wu2021autoformer}}} \\
\\
\midrule
\multirow{3}{*}{Yearly} 
& SMAPE& \textcolor{red}{13.199}  & \underline{13.206} & 13.387 & 13.418 & 13.436 & 18.605 & 16.463 & 25.022 & 17.431 & 14.247 & 16.965 & 13.728 & 13.717 & 13.974  \\
& MASE&\underline{2.953} & \textcolor{red}{2.916} & 2.996 & 3.045 & 3.043 & 4.471 & 3.967 & 7.162 & 4.043 & 3.109 & 4.283 & 3.048 & 3.078 & 3.134  \\
& OWA &\textcolor{red}{0.775} & \underline{0.776} & 0.786 & 0.793 & 0.794 & 1.132 & 1.003 & 1.667 & 1.042 & 0.827 & 1.058 & 0.803 & 0.807 & 0.822  \\
\midrule
\multirow{3}{*}{Quarterly} 
& SMAPE&\textcolor{red}{9.955} & \underline{9.996} & 10.100 & 10.202 & 10.124 & 14.871 & 10.644 & 15.214 & 12.925 & 11.364 & 12.145 & 10.792 & 10.958 & 11.338 \\
& MASE &\textcolor{red}{1.163} & \underline{ 1.166} & 1.182& 1.194& 1.169& 2.054& 1.278& 1.963& 1.664& 1.328& 1.520& 1.283& 1.325& 1.365 \\
& OWA  & \underline{0.876}& \textcolor{red}{0.825} & 0.890 & 0.899 & 0.886 & 1.424 & 0.949 & 1.407 & 1.193 & 1.000 & 1.106 & 0.958 & 0.981 & 1.012  \\
\midrule
\multirow{3}{*}{Monthly} 
& SMAPE &\textcolor{red}{12.318} &\underline{ 12.605}& 12.670 &12.791 &12.677& 14.925& 13.399 &16.943& 15.407& 14.014& 13.514& 14.260& 13.917& 13.958 \\
& MASE &\textcolor{red}{0.903} & \underline{0.919}& 0.933 &0.969 &0.937 &1.131& 1.031& 1.442& 1.298 &1.053 &1.037& 1.102& 1.097 &1.103  \\
& OWA  & \textcolor{red}{0.852} &\underline{ 0.869}& 0.878& 0.899 &0.880& 1.027& 0.949& 1.265& 1.144& 0.981& 0.956& 1.012& 0.998& 1.002\\
\midrule
\multirow{3}{*}{Others} 
& SMAPE&\textcolor{red}{4.358} & \underline{4.564} &4.891& 5.061& 4.925 &16.655& 6.558& 41.985& 7.134& 15.880& 6.709& 4.954& 6.302& 5.485  \\
& MASE & \textcolor{red}{3.041} &\underline{3.115} &3.302& 3.216& 3.391& 15.034& 4.511& 62.734& 5.09& 11.434& 4.953& 3.264& 4.064& 3.865 \\
& OWA &\textcolor{red}{0.938} &\underline{ 0.982}&1.035& 1.040 & 1.053 & 4.123& 1.401 & 14.313 & 1.553 & 3.474& 1.487 & 1.036 & 1.304 & 1.187 \\
\midrule
\multirow{3}{*}{Average} 
& SMAPE&\textcolor{red}{11.555} & \underline{ 11.723} &11.829& 11.927& 11.851& 15.542& 13.152& 19.638& 14.863& 13.525& 13.639& 12.840& 12.780& 12.909 \\
& MASE &\textcolor{red}{1.544}&\underline{ 1.559 }& 1.585& 1.613& 1.559& 2.816& 1.945& 5.947& 2.207& 2.111& 2.095& 1.701& 1.756& 1.771\\
& OWA &\textcolor{red}{0.830} & \underline{ 0.840} & 0.851& 0.861& 0.855& 1.309& 0.998& 2.279& 1.125& 1.051& 1.051& 0.918& 0.930& 0.939\\

\midrule
\end{tabular}
\end{adjustbox}
\vspace{1mm}
\label{table3}
\footnotesize{Some results in this table are directly copied from TimeMixer (2024), as we use the same evaluation method and setting they used.}
\vspace{-8pt}
\end{table*}

\textbf{FBM-S vs. the Other Three FBM Variants}: We decompose the separated effects into trend, seasonal, and interaction components. First, we introduce a convolution filter for capturing seasonal effects, which significantly reduces the complexity while improving the robustness of the extracted seasonal features. Second, we improve both efficiency and performance by applying patching to MLP- and Transformer-based architectures in the trend block. We find that an MLP backbone works better than a transformer backbone in the trend block, with the Traffic dataset being the only exception. For all other datasets, a single-layer projection is sufficient to replace stacked of attention layers, yielding better performance. However, the linear backbone also works for a few high-noise datasets, such as ETTh1, ETTh2, and Exchange. Consequently, we use a Transformer backbone only for the Traffic dataset and an MLP backbone for the remaining datasets, respectively. Third, we observe that interaction effects hold a value but play a relatively minor role in long-term TSF. The interaction block is effective for Electricity, Traffic, ETTm1, ETTm2, and WTH datasets when the interaction masking is applied. Additionally, we observe that multi-scale down-sampling is more effective for high-granularity datasets such as ETTm1 at the 15-minute level rather than the hourly level datasets. Finally, FBM-S outperforms the other three FBM variants in nearly all cases, achieving SOTA performance in most scenarios. Detail discussion can be found in the ablation studies in Section \ref{Ablation}.


\subsection{Short-term TSF Results}
\label{analysis}
In Table \ref{table_short}, we compare our proposed FBM-variants with other baselines on four PEMS datasets for STSF. The PEMS datasets have high granularity, with data recorded at 5-minute intervals. As a result, FBM-S shows a notably greater performance improvement over the other three FBM variants across all four PEMS datasets by involving the interaction block. This can be attributed to two main reasons. First, the forecast horizon is shortened for $L=(12,24,48,96)$. Second, the data are recorded at 5-minute intervals. Thus, the actual forecast horizon in real world time is very short. The interaction effects become much more pronounced over short periods, as previously mentioned. The trend and seasonal effects are primarily driven by endogenous temporal patterns within each time series, whereas the interaction effects are influenced by exogenous dependencies across multivariate time series. It is worth mentioning that centralization technique and the interaction mask also play an important role in improving the significance of the interaction block. We use an input mask with $C_1=24$ and an output mask with $C_2=L$. The input mask helps remove outdated redundant temporal information for interaction effects. It also helps improve the efficiency of the initial projection. Centralization technique helps the model determine whether the recent time series segment corresponds to a high-peak, off-peak, or normal state period for each variate. Additionally, the proposed multi-scale down-sampling and patching with a MLP bankbone in the trend block are also effective for short-term TSF. Finally, we observe that FBM-S achieves SOTA performance all the time. Detail discussion can be found in the ablation studies in Section \ref{Ablation}.

In Table \ref{table3}, we compare our proposed FBM-S with other baselines on the M4 dataset across different sampling frequencies. Since M4 is a univariate dataset, we won't use the interaction block. We follow the official setting used in TimeMixer, where the input length is set to twice the forecast horizon. As mentioned, the forecast horizons are 6 for yearly, 8 for quarterly, 18 for monthly, 13 for weekly, and 14 for daily and 48 for hourly, respectively. Given the relatively short input lengths, patching becomes unnecessary. We observe that when the forecast horizon increases, our method becomes increasingly competitive and achieves the best overall performance. This is because the number of meaningful frequency levels is proportional to the input series length; more frequency levels lead to better results. All experiments are conducted using the same random seed and the fixed hyperparameters listed in Table \ref{hyperparameter} to ensure the best reproducibility of the reported results. The results demonstrate that our model is simple and effective.

\begin{table*}[t]
\centering
\caption{Efficiency Analysis on the PEMS08 Dataset (Batch Size = 64, $T=336$, $L=96$)}
\vspace{5pt}
\begin{adjustbox}{width=\columnwidth*2,center}
\begin{tabular}{ccccccccccccccc} 
&  FBM-S &   FBM-L   & FBM-NL  &  FBM-NP  & NLinear \cite{zeng2023transformers} & PatchTST \cite{nie2022time}& TimeMixer \cite{TimeMixer}& iTransformer \cite{LiuHZWWML24}& FiLM \cite{zhou2022film}& TimesNet \cite{wu2022timesnet} \\ 
\midrule 
Training time (s) & 10.94  & 4.39&  23.34 & 13.50 &  2.38 & 33.53 & 7.95 & 4.94 & 134.95 & 432.95   \\ 
\midrule 
Memory (KMiB)& 13.06 &9.55 & 10.11 & 15.61 & 0.11 & 20.04 & 2.63 & 2.95 & 43.41 & 5.28 \\
\midrule 
GFLOPs & 172.15 & 59.31 & 913.71 & 141.31 & 0.35 & 187.87 & 36.40 & 53.90 & - & 41786.75 \\
\midrule 
\end{tabular}
\end{adjustbox}
\label{efficiency}
\end{table*}

\begin{table*}[ht]
\centering
\caption{FBM-S Model Configuration Details in the Trend, Seasonal, and Interaction Blocks. Results are measured on the PEMS08 dataset with a batch size of 64 with $T=336$ and $L=96$. }
\vspace{5pt}
\begin{adjustbox}{width=\columnwidth*2,center}
\begin{tabular}{cccccccccccc} 
&  Trend Block (MLP)      &  Trend Block (MLP+Patch)  & Trend Block (Transformer+Patch) & Interaction Block (Transformer) & Seasonal Backbone \\ 
\midrule 
Hyperparameter&  $h_1=h_2=1440$ & $P=14$, $h_1=256$, $h_2=1440$   &  $P=14$, $h_1=h_2=256$, $K=3$ & $h_3=512$, $K=3$, $C_1=24$ & -- \\
\midrule 
Initial Projection& $ \frac{1}{4}\times\frac{T^2}{2} \times h_1=20.32M$  & $ \frac{1}{4}\times\frac{T^2}{2P} \times h_1=0.26M $  &  $ \frac{1}{4}\times\frac{T^2}{2P}\times h_1=0.26M $  &  $ (24 \times \frac{T}{2})\times h_3 =2.06M$ & -- \\
\midrule 
Intermediate Layer & $ h_1 \times h_2=2.07M$  & $ (h1 \times P) \times h_2=5.16M$  & $h1 \times  h_1  \times 4 \times 3 + h1 \times  h_2  \times 2 \times 3  =1.18M $   &  $ h_3 \times  h_3 \times 6 \times 3= 4.71M $ & $0.05M$ \\  
\midrule 
Final Projection & $ h_2 \times L=0.14M$  &  $ h_2 \times L=0.14M$  &   $( h_1 \times P )\times L=0.17M$   & $h_3 \times L=0.05M$ & -- \\   
\midrule 
Total Parameter & $22.53 M$ & $5.56M$    & $1.61M$ & $6.82M$   & $0.05M$   \\    
\midrule 
FLOPs(G)(Batch Size=64) &  245.16 &  96.96  & 223.67 & 74.49  &  0.70 \\  
\midrule 
\end{tabular}
\end{adjustbox}
\label{Configuration}
\footnotesize{$h_1$ and $h_2$ refer to the hidden states within trend block, $h_3$ refer to the hidden states within the interaction block. $P$ refers to the number of patches, $K$ refers to the number of attention stacks, and $C_1$ and  $C_2$ refers to the input and output interaction mask.}
\vspace{-10pt}
\end{table*}

\subsection{Efficiency Analysis}
\label{complexity}

Table \ref{efficiency} shows the computational efficiency of our method with baseline methods, reporting average training time per epoch, memory usage, and FLOPs. Table \ref{Configuration} specifies the default configuration of each block, including layer architecture, number of parameters, and FLOPs. All evaluations are conducted on the PEMS08 dataset with $T = 336$, $L = 96$, and a batch size of 64. 

In Table \ref{efficiency}, FBM-S demonstrates good overall efficiency and achieves the best performance among the existing methods. FBM-S is much more efficient than the non-linear FBM variants, FBM-NL and FBM-NP, while also delivering significantly better performance. It is also more efficient than PatchTST, FiLM, and TimesNet, though it slightly increases complexity compared to TimeMixer and iTransformer.


In Table \ref{Configuration}, we show the detailed configuration of each layer within each block, where each block, except for the seasonal block, can be decomposed into an initial projection, the intermediate layer, and a final projection layer. The intermediate layer are either a single projection layer in the MLP backbone or the stacks of attention layers in the attention backbone. In the trend block, the patching technique substantially improves the overall efficiency of the MLP-based architecture. We also observe that a three-layer MLP with patching achieves lower FLOPs compared to Transformer-based methods. The speed is also faster since the trend block with MLP requires only three rounds of matrix multiplication, whereas the trend block with Transformer requires twenty-four rounds of matrix multiplication. The results demonstrate that both seasonal block (MLP + patch) and interaction block achieve high efficiency, as evidenced by low FLOPs with a batch size of 64. 

This further highlights the importance of time-frequency features. By slightly increasing the complexity of the initial layer, we can reduce that of the downstream layers. For example, FBM-NP outperforms PatchTST not only in predictive performance but also in computational efficiency as we use fewer patches of 14. The experiments are conducted on a NVIDIA H100 GPU and Intel Xeon Gold CPU.

\begin{table}[ht]
\caption{Ablation Study I: Enhancing the Trend (T) Block with Seasonal (S) and Interaction (I) Blocks in LTSF. Masking (M) and Centralization (C) are applied in the interaction block.}
\centering
\begin{adjustbox}{width=\columnwidth}
\begin{tabular}{cccccccccccc} 
\midrule
\multicolumn{2}{c}{Model}   & \multicolumn{2}{c}{T}  & \multicolumn{2}{c}{S+T}  & \multicolumn{2}{c}{S+T+I(C)}&  \multicolumn{2}{c}{S+T+I(C+M)}\\ 

\midrule
\multicolumn{2}{c}{Error}   & MSE                    & MAE                    & MSE                    & MAE   & MSE  &   MAE      & MSE  &   MAE                                           \\ 
\midrule
\multirow{4}{*}{ETTh1}   & 96     &     0.367    &  0.392 &  \textcolor{red}{ 0.363}          &   \textcolor{red}{ 0.389}          & 0.366 &  0.390 &\underline{ 0.365} & \underline{ 0.389}    \\ 

                         & 192     &     0.403      &    0.411  &  \textcolor{red}{0.399 }            &  \textcolor{red}{0.409}       &0.401& 0.411   & \underline{ 0.399}&  \underline{ 0.409}        \\ 

                         & 336  &   0.418        &  0.420   &   \textcolor{red}{ 0.402}        &   \textcolor{red}{0.413 }   & 0.408&0.417   & \underline{ 0.403}& \underline{ 0.413}                       \\ 

                         & 720  &    0.417    &     0.439&    \textcolor{red}{0.403}          &     \textcolor{red}{0.433}  & 0.409  & 0.436& \underline{ 0.403} & \underline{ 0.433}  \\   

\midrule
\multirow{4}{*}{ETTm1}   & 96  &    0.290   & 0.346      &  0.278      & 0.332          & 0.383 & 0.338  & \textcolor{red}{0.278} &  \textcolor{red}{0.332}   \\ 

                         & 192 &     0.330        &  0.371     &  0.319      &  0.359         & 0.322 & 0.363  & \textcolor{red}{0.317} & \textcolor{red}{0.358}        \\ 

                         & 336 &   0.362          &  0.390     &  0.359      &     0.384     & 0.362& 0.390  & \textcolor{red}{0.356} & \textcolor{red}{0.382}                      \\ 

                         & 720  &       0.415      &  0.421     &   0.413     &    0.419      &  0.415& 0.419  & \textcolor{red}{0.412} & \textcolor{red}{0.418}    \\   

\midrule

\end{tabular}
\end{adjustbox}
\label{abalation1}
\end{table}

\begin{table}[ht]
\caption{Ablation Study II: Removing the Interaction Block, Centralization (C) and Masking (M) Techniques within the Interaction Block in STSF.}
\centering
\begin{adjustbox}{width=\columnwidth}
\begin{tabular}{cccccccccccc} 
\midrule
\multicolumn{2}{c}{Model}   & \multicolumn{2}{c}{S+T+I(C+M)}  & \multicolumn{2}{c}{ S+T+I(M)}   & \multicolumn{2}{c}{S+T+I(C)} &  \multicolumn{2}{c}{S+T} \\ 

\midrule
\multicolumn{2}{c}{Error}   & MSE                    & MAE                    & MSE                    & MAE   & MSE  &   MAE     & MSE  &   MAE                                          \\ 
\midrule
\multirow{4}{*}{PEMS08}   & 12  &   \textcolor{red}{ 0.055}          &   \textcolor{red}{0.147}     &     \underline{ 0.056}  & \underline{0.148}      & 0.057 & 0.150      &     0.058   & 0.151       \\ 

                         & 24  &  \textcolor{red}{ 0.064}         & \textcolor{red}{ 0.157}    &         \underline{0.065} & \underline{0.159}  & 0.066  & 0.159 & 0.069        & 0.163                      \\ 

                         & 48  &   \textcolor{red}{ 0.073}     &   \textcolor{red}{0.167}  &  \underline{0.076}         & \underline{ 0.170} & 0.077 & 0.171 & 0.085 & 0.178                         \\ 

                         & 96 &  \textcolor{red}{0.083}         &  \textcolor{red}{0.177}    & \underline{0.088}         &  \underline{0.181} & 0.088 & 0.181  &  0.102        & 0.194   \\   

\midrule
\end{tabular}
\end{adjustbox}
\label{abalation2}
\end{table}

\begin{table}[ht]
\caption{Ablation Study III: Effects of the Input Interaction Mask Range ($C_1$) with $C_2=L$ in STSF.}
\centering
\begin{adjustbox}{width=\columnwidth}
\begin{tabular}{cccccccccccc} 
\midrule
\multicolumn{2}{c}{Parameter}   & \multicolumn{2}{c}{$C_1=24$}  & \multicolumn{2}{c}{$C_1=48$} & \multicolumn{2}{c}{$C_1=96$} & \multicolumn{2}{c}{$C_1=336$} \\ 

\midrule
\multicolumn{2}{c}{Error}   & MSE                    & MAE                    & MSE                    & MAE                    & MSE   & MAE        & MSE   & MAE                                \\ 
\midrule
\multirow{4}{*}{PEMS08}   & 12  & \textcolor{red}{ 0.055}       & \textcolor{red}{0.147}     &   \underline{0.056}   &  \underline{0.148}       &0.056 & 0.150  & 0.057 & 0.150        \\ 

                         & 24 &   \textcolor{red}{ 0.064}     & \textcolor{red}{0.157}     &    \underline{0.064}  &  \underline{0.157}        & 0.065& 0.159  & 0.066  & 0.159                \\ 

                         & 48  &  \textcolor{red}{ 0.073 }     &  \textcolor{red}{0.167}    &  \underline{ 0.074}   &  \underline{0.168}       & 0.074   & 0.168  & 0.077 & 0.171                             \\ 

                         & 96 &    0.083     & 0.177     & \textcolor{red}{0.082 }   & \textcolor{red}{0.175}        &  \underline{0.083} &  \underline{0.176} & 0.088 & 0.181                          \\ 
\midrule
\end{tabular}
\end{adjustbox}
\label{abalation3}
\end{table}

\begin{table}[ht]
\caption{Ablation study IV: The Effects of the Output Interaction Mask Range ($C_2$) with $C_1=24$ in LTSF}
\centering
\begin{adjustbox}{width=\columnwidth}
\begin{tabular}{cccccccccccc} 
\midrule
\multicolumn{2}{c}{Parameter}   & \multicolumn{2}{c}{$C_2=0$}  & \multicolumn{2}{c}{$C_2=24$} & \multicolumn{2}{c}{$C_2=48$} & \multicolumn{2}{c}{$C_2=336$} \\ 

\midrule
\multicolumn{2}{c}{Error}   & MSE                    & MAE                    & MSE                    & MAE                    & MSE   & MAE        & MSE   & MAE                                \\ 
\midrule
\multirow{4}{*}{Electricity}   & 96  &   \underline{0.128}      & \underline{0.220 }    & \textcolor{red}{0.127}    &\textcolor{red}{ 0.220}          & 0.129 & 0.222& 0.129& 0.222       \\ 

                         & 192 &  \underline{0.144}  & \underline{0.237}     &  \textcolor{red}{ 0.144 } & \textcolor{red}{0.237}         & 0.148 & 0.240 & 0.151 & 0.242                \\ 

                         & 336  &  \underline{0.162}       &  \underline{0.255}   &  \textcolor{red}{0.161}   & \textcolor{red}{0.254}       & 0.164   & 0.256  &0.168  & 0.261                             \\ 

                         & 720 &  \underline{ 0.196}      & \underline{0.286}     &   \textcolor{red}{0.195}  &\textcolor{red}{ 0.285}         &  0.195 & 0.285 & 0.201 &  0.289                         \\ 
\midrule
\end{tabular}
\end{adjustbox}
\label{abalation4}
\end{table}

\begin{table}[ht]
\caption{Ablation Study V: Effects of the Multi-Scale Down-Sampling on PEMS08.}
\centering
\begin{adjustbox}{width=\columnwidth}
\begin{tabular}{cccccccccccc} 
\midrule
\multicolumn{2}{c}{Model}   & \multicolumn{2}{c}{$d_o$}  & \multicolumn{2}{c}{$d_1$}  & \multicolumn{2}{c}{$d_o+d_1$} & \multicolumn{2}{c}{$d_o+d_1+d_2$}\\ 

\midrule
\multicolumn{2}{c}{Error}   & MSE                    & MAE                    & MSE                    & MAE   & MSE  &   MAE  & MSE  &   MAE                                             \\ 
\midrule
\multirow{4}{*}{PEMS08}   & 12&   \underline{0.056}   & 0.148  &   \textcolor{red}{0.055}          &  \textcolor{red}{0.147}               &  0.057   &  \underline{0.147} &     0.057  & 0.149    \\ 

                         & 24  &   0.064       &  0.158  &  \underline{ 0.064 }        &  \underline{0.157}    & 0.064 & 0.158           &  \textcolor{red}{0.064}& \textcolor{red}{0.155}           \\ 

                         & 48   &    0.075       & 0.170 &  \textcolor{red}{ 0.073}     &  \textcolor{red}{0.167}   & 0.074 & 0.168  & \underline{ 0.074} &  \underline{0.168}                       \\ 

                         & 96 &      \underline{ 0.083}    & 0.177  & \textcolor{red}{0.083}         &  \underline{0.177}      &      0.083    & 0.176 & 0.084 & \textcolor{red}{0.176}  \\   

\midrule
\end{tabular}
\end{adjustbox}
\label{abalation5}
\end{table}

\begin{table}[ht]
\caption{Ablation Study VI: Effects of Patching with Centralization (C) Technique in the Trend Block}
\centering
\begin{adjustbox}{width=\columnwidth}
\begin{tabular}{cccccccccccc} 
\midrule
\multicolumn{2}{c}{Trend Block}   & \multicolumn{2}{c}{Patch(C)}  & \multicolumn{2}{c}{Patch} & \multicolumn{2}{c}{w/o Patch}  \\ 

\midrule
\multicolumn{2}{c}{Error}   & MSE                    & MAE                    & MSE                    & MAE                    & MSE   & MAE                                       \\ 
\midrule
\multirow{4}{*}{Weather}   & 96  &  \textcolor{red}{  0.147} & \textcolor{red}{ 0.196}     &  0.152   &    0.199       &  \underline{0.152} & \underline{0.199}     \\ 

                         & 192 &  \textcolor{red}{  0.189}   & \textcolor{red}{ 0.238}     & 0.194    &   0.241      & \underline{0.194} & \underline{0.241}                \\ 

                         & 336  &  \textcolor{red}{  0.238 }&  \textcolor{red}{ 0.276}   &  0.245  & 0.283       &  \underline{ 0.244}& \underline{0.282 }                              \\ 

                         & 720 & \textcolor{red}{  0.311}  & \textcolor{red}{ 0.328}    &  0.317  &  0.335       & \underline{0.316} & \underline{0.332}                          \\ 
\midrule
\end{tabular}
\end{adjustbox}
\label{abalation6}
\end{table}

\begin{table}[ht]
\caption{Ablation Study VII :  Linear VS Non-Linear Initial Projection and MLP vs Transformer Networks.}
\centering
\begin{adjustbox}{width=\columnwidth}
\begin{tabular}{cccccccccccc} 
\midrule
\multicolumn{2}{c}{\multirow{2}{*}{Model}}   & \multicolumn{2}{c}{Linear Proj}  & \multicolumn{2}{c}{Non-Linear Proj} & \multicolumn{2}{c}{Linear Proj} & \multicolumn{2}{c}{Non-Linear Pro} \\ 

                     &       &      \multicolumn{2}{c}{ (MLP)}     &   \multicolumn{2}{c}{ (MLP)}          &  \multicolumn{2}{c}{ (Transformer)} & \multicolumn{2}{c}{ (Transformer)}                 \\ 

\midrule
\multicolumn{2}{c}{Error}   & MSE                    & MAE                    & MSE                    & MAE                    & MSE   & MAE        & MSE   & MAE                                \\ 
\midrule
\multirow{4}{*}{PEMS08}   & 12  &  \underline{ 0.057}  & \underline{0.149}    &  \textcolor{red}{0.055}    & \textcolor{red}{ 0.147}  & 0.058 & 0.152 &  0.057 & 0.150       \\ 

                         & 24 &  0.066  &  0.161   &   \textcolor{red}{0.064}  &  \textcolor{red}{0.157}         & 0.065 & 0.159 &  \underline{0.065}& \underline{0.158}                \\ 

                         & 48  &  0.075  & 0.169   &   \textcolor{red}{0.073}   &  \textcolor{red}{ 0.167}     & 0.075  & 0.168  &\underline{ 0.075} & \underline{0.168}                             \\ 

                         & 96 & 0.086   & 0.180  &    \textcolor{red}{0.083}  &  \textcolor{red}{0.177}    & \underline{0.084} & \underline{0.177} & 0.084 & 0.178                          \\ 
\midrule
\multirow{4}{*}{Electricity}   & 96  &   0.127  & 0.220   &  \underline{ 0.127} & \underline{0.220} & \textcolor{red}{0.126} & \textcolor{red}{0.220}& 0.127  & 0.220      \\ 

                         & 192 &  0.145 & 0.237    &  \textcolor{red}{0.144}  & \textcolor{red}{0.237}        & \underline{0.144} & \underline{0.237} &  0.145& 0.237               \\ 

                         & 336  & 0.163   & 0.256   &  \textcolor{red}{ 0.161}  &\textcolor{red}{ 0.254}      & 0.162  &  0.255 &  \underline{0.161}& \underline{0.254}                           \\ 

                         & 720 &  0.197  & 0.287  &  \textcolor{red}{0.195}   & \textcolor{red}{0.285}        & 0.197& 0.287& \underline{0.195} & \underline{0.285}                        \\ 
\midrule

\end{tabular}
\end{adjustbox}
\label{abalation7}
\end{table}

\subsection{Ablation Studies}
\label{Ablation}

We conduct seven ablation studies of the designs for time-frequency features. We first analyze the network's decomposition into trend, seasonal, and interaction blocks. 

In Table \ref{abalation1}, we first observe that combining the trend block with the seasonal block consistently outperforms that with the trend block alone. This convolutional filter encourages the output features to exhibit seasonal characteristics, which enhances the robustness of the model. Second, we observe that the interaction block is generally less effective in LTSF, as adding the interaction block with masking only slightly increases the performance on ETTm1. Third, applying a mask with $C_1 = C_2 = 48$ consistently outperforms the setting without such a mask. Incorporating interaction mechanisms tends to degrade performance on ETTh1 but improves it on ETTm1, as the data granularity shifts from hourly to 15-minute intervals. This further highlights that interaction effects become more important over shorter time periods. This explains why FBM-S shows a significant performance improvement over the other three FBM variants in STSF. We then conduct a more detailed analysis of the effects of the interaction block on the PEMS dataset.

In Table \ref{abalation2}, we further evaluate the impact of the interaction block on PEMS08, as well as two design choices: centralization and masking. We find that removing the interaction block leads to a significant drop in performance on the PEMS08 dataset. This can be attributed to the shorter forecast horizon $L$ and the high granularity of the 5-minute sampling frequency in the PEMS dataset. These results suggest that while both dependent channel modeling and independent channel modeling are effective, dependent channel modeling plays an increasingly important role in the short term. In Table \ref{abalation2}, the masking and centralization techniques are also proven to be effective. Centralizing the final patch of time-frequency features allows the model to better interpret whether the current segment of the time series corresponds to a high-peak or low-peak period. This enhances the downstream interaction mapping by providing a clearer intensity level for each variate. The masking allows the model to incorporate only the most relevant time-frequency features while removing redundant information. 

In Table \ref{abalation3}, we then investigate the impact of the input interaction mask range on the PEMS dataset. The results show that using a longer range of time-frequency features does not improve performance. Instead, it gradually leads to performance degradation. This highlights that the interaction effects primarily arise from the most recent time-frequency features. The best results are achieved with $C_1 = 24$ and $C_2 = 48$. Using the full time-frequency representation leads to the worst performance, as it includes excessive and unnecessary information for interaction effects. On the other hand, longer-range time-frequency representations are useful for capturing trend and seasonal effects. This underscores the necessity of separating the trend, seasonal and interaction components, with masking applied specifically in the interaction block.

In Table \ref{abalation4}, we evaluate the effect of varying the output interaction mask range on the Electricity dataset for LTSF. This analysis is motivated by our earlier reasoning that the interaction effects may not has a long-term influence. By comparing the first and second columns, we observe that the interaction block provides a small performance improvement only when an output mask is applied. With the input mask $C_1$ fixed at 24, we find that the best performance is achieved when the output mask $C_2$ is also set to 24. Increasing or decreasing the value will lead to worse results.

In Table \ref{abalation5}, we evaluate the impact of multi-scale down-sampling and find that it is particularly effective for high-granularity data. The PEMS datasets are collected at a five-minute interval. Here, $d_0$ refers to the original time-frequency input, $d_1$ denotes down-sampled input with kernel $2$, and $d_2$ denotes down-sampled input with kernel $4$. We evaluate the performance in different combinations and observe that using $d_1$ yields the best results. This suggests that moderate down-sampling can enhance the quality of time-frequency features, as excessively high resolution may lead to overfitting.

In Table \ref{abalation6}, We evaluate the effects of patching with centralization technique within trend block. We find that applying patching together with centralization improves performance compared to using the entire time-frequency features. Notably, as discussed in Section \ref{complexity}, patching also significantly enhances computational efficiency. Thus, combining patching with centralization proves to be a highly effective approach. This is because centralization helps the model better capture the overall trend within a short temporal window, leading to improved performance.

In Table \ref{abalation7}, we first compare MLP-based and Transformer-based networks for patched time-frequency features. The MLP-based network consistently outperforms the Transformer-based network across those datasets. In Section \ref{complexity}, we show that the MLP-based architecture is more efficient than the transformer architecture. These findings underscore the superior efficiency and performance of the MLP-based network in modeling time-frequency features for trend effects. Second, we further test the effect of adding an activation function after the initial projection layer of the patched time-frequency features. We observe that the model achieves better performance with the activation function. This improvement is attributed to the activation function helping the model capture better nonlinear trend effects. Although the downstream intermediate layer includes activation functions, incorporating an additional activation function immediately after the initial projection also brings benefits and improve robustness. This enhancement is more obvious in MLP-based network than Transformer-based network.

In Table \ref{setting}, we analyze different experimental settings. We compare performance using either a train/validation/test split of 0.7/0.1/0.2 or an input length of $T=96$, as adopted in many previous works. We find that using an input length of $336$ consistently yields better results, and using a larger training set improves performance, therefore we will consider use a larger training set in future work.

\begin{table}[h]
\caption{Experiment Settings for Different Input Lengths and Train/Vali/Test Splits}
\centering
\begin{adjustbox}{width=\columnwidth}
\begin{tabular}{cccccccccccccccccc} 
\midrule
 \multicolumn{2}{c}{Dataset}&\multicolumn{4}{c}{Electricity}  & \multicolumn{4}{c}{Traffic}\\ 

\midrule
\multicolumn{2}{c}{L} & 96 & 192 & 336 &720  & 96 & 192 & 336 &720        \\ 
\midrule
\multirow{2}{*}{Ours:(T=336, 0.65/0.15/0.2)}   & MSE   & 0.127 & 0.144 & 0.161 & 0.195 & 0.357 & 0.382 & 0.393 &  0.430       \\ 

                         & MAE   & 0.220 & 0.237 &  0.254 & 0.285 & 0.246 & 0.257 & 0.263 & 0.285                           \\ 
\midrule
\multirow{2}{*}{(T=96, 0.65/0.15/0.2)}   & MSE & 0.138 & 0.153 & 0.170 & 0.206 & 0.408 &0.420 & 0.432 & 0.462            \\ 
                         & MAE  & 0.233 & 0.247 & 0.265 &0.297  & 0.257 & 0.269 & 0.276 &  0.291                       \\ 
\midrule                         
\multirow{2}{*}{(T=336, 0.7/0.1/0.2)}   & MSE&  0.124& 0.141 & 0.157 & 0.191  &0.346& 0.367&0.382 &0.414           \\ 

                         & MAE& 0.217 & 0.234 & 0.251 & 0.282     & 0.244& 0.252& 0.260& 0.281                              \\                          
                         
\midrule
\end{tabular}
\end{adjustbox}
\label{setting}
\end{table}

\begin{figure*}[t]
\begin{minipage}[t]{0.20\linewidth}
\centering
\includegraphics[width=\textwidth,height=0.7\textwidth]{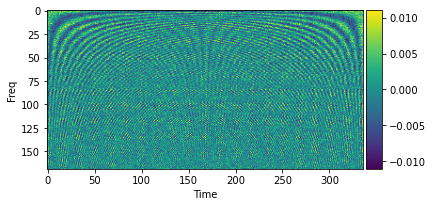}
\caption*{$W_{0}$}
\end{minipage}%
\begin{minipage}[t]{0.20\linewidth}
\centering
\includegraphics[width=\textwidth,height=0.7\textwidth]{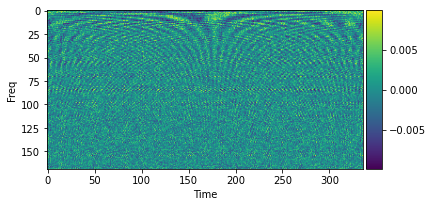}
\caption*{$W_{10}$}
\end{minipage}%
\begin{minipage}[t]{0.20\linewidth}
\centering
\includegraphics[width=\textwidth,height=0.7\textwidth]{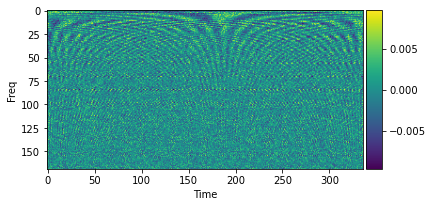}
\caption*{$W_{20}$}
\end{minipage}%
\begin{minipage}[t]{0.20\linewidth}
\centering
\includegraphics[width=\textwidth,height=0.7\textwidth]{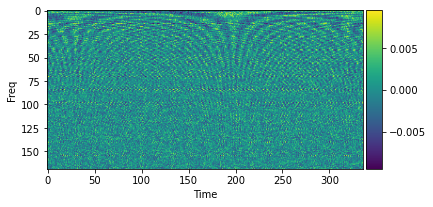}
\caption*{$W_{30}$}
\end{minipage}%
\begin{minipage}[t]{0.20\linewidth}
\centering
\includegraphics[width=\textwidth,height=0.7 \textwidth]{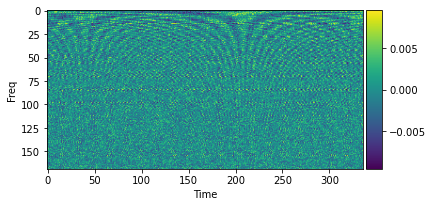}
\caption*{$W_{40}$}
\end{minipage}
\caption{Visualization of the Weights $W_0$, $W_{10}$, $W_{20}$, $W_{30}$, and $W_{40}$ of FBM-L on the Electricity Dataset. Each $\mathbf{W}_i$ represents the influence of time-frequency features on the $i$-th time step of the predicted output $\mathbf{\hat{Y}}$. The $x$-axis denotes time, and the $y$-axis denotes frequency. }
\label{weight2}
\end{figure*}


\begin{figure*}[t]
\centering
\begin{minipage}[t]{0.23\linewidth}
\centering
\includegraphics[width=\textwidth,height=0.7\textwidth]{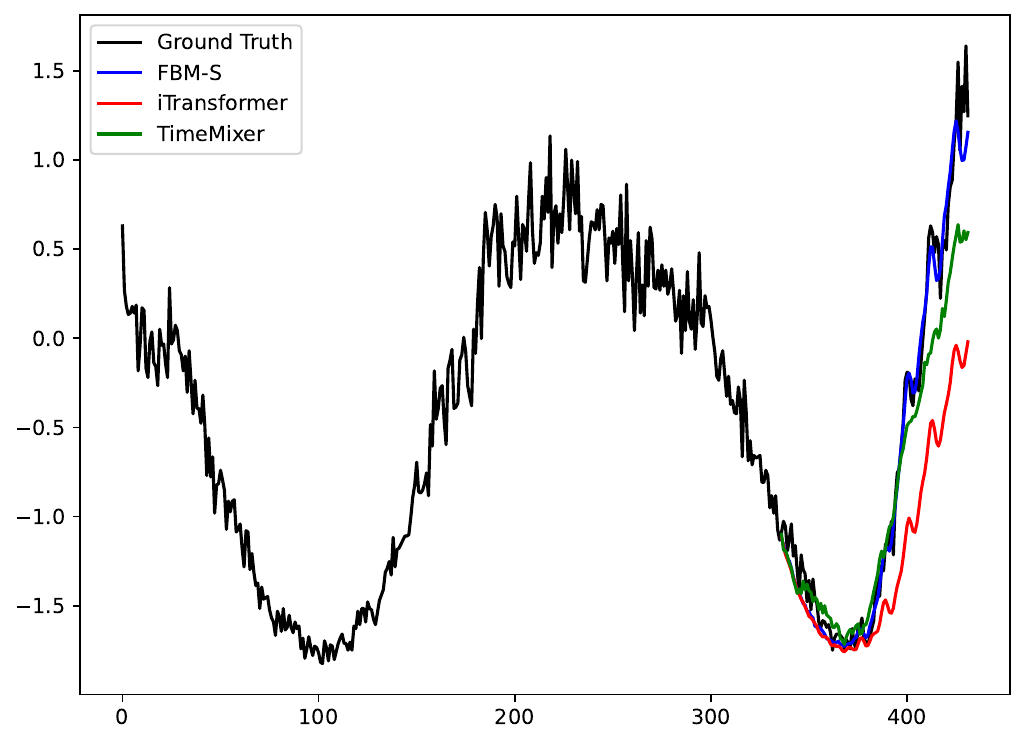}
\caption*{(a)}
\end{minipage}
\hfill
\begin{minipage}[t]{0.23\linewidth}
\centering
\includegraphics[width=\textwidth,height=0.7\textwidth]{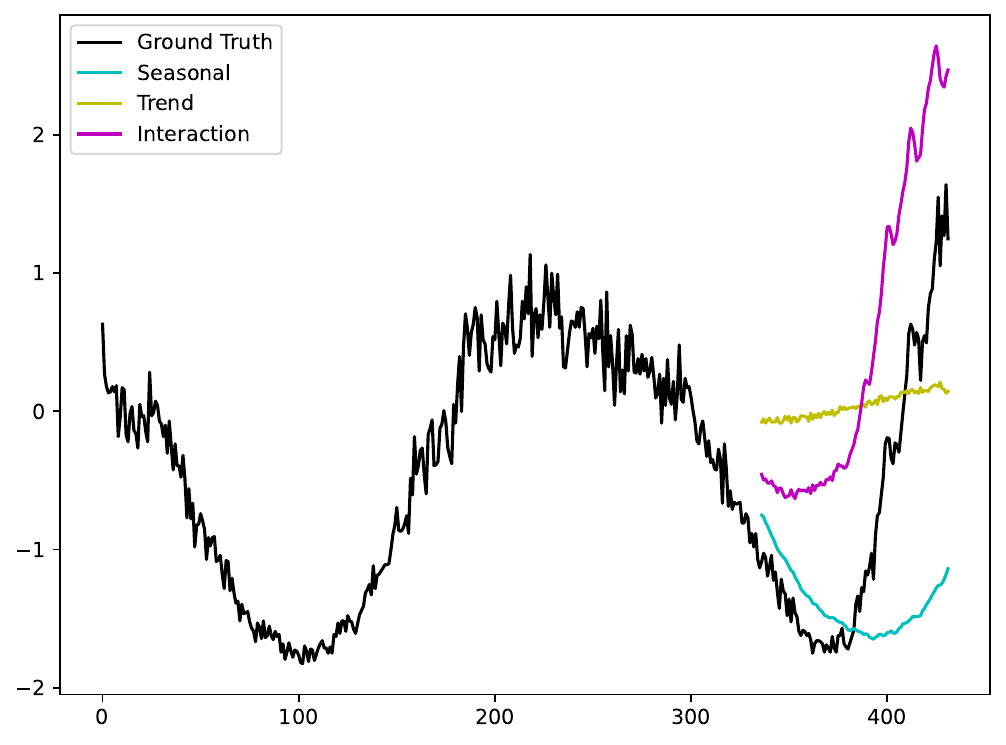}
\caption*{(b)}
\end{minipage}
\hfill
\begin{minipage}[t]{0.23\linewidth}
\centering
\includegraphics[width=\textwidth,height=0.7\textwidth]{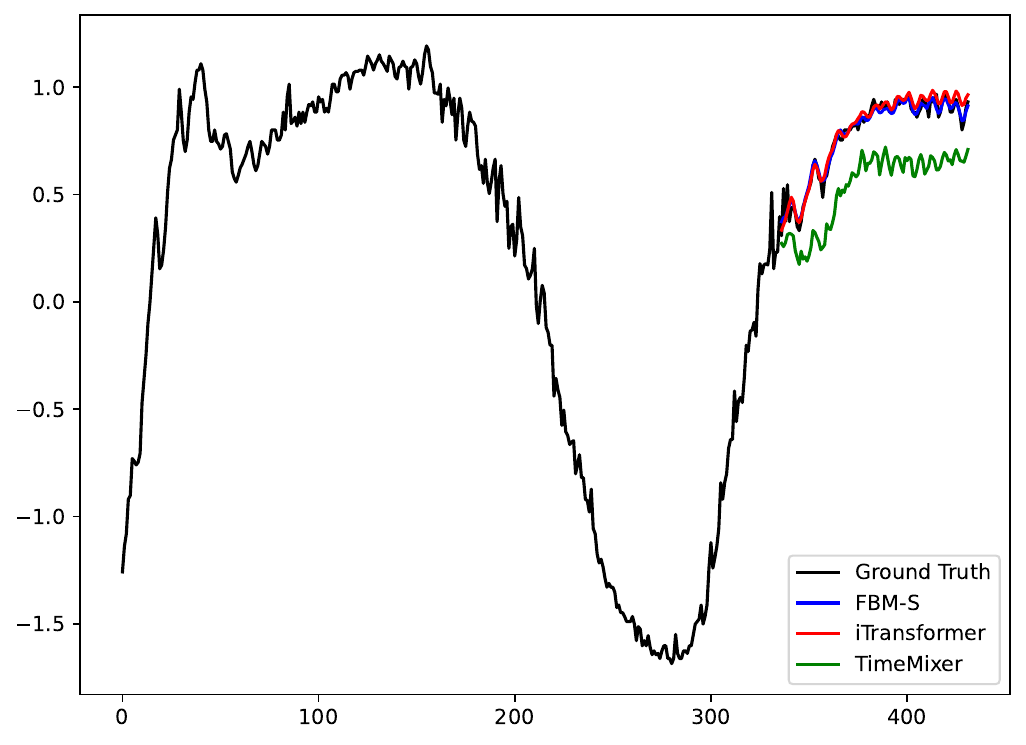}
\caption*{(c)}
\end{minipage}
\hfill
\begin{minipage}[t]{0.23\linewidth}
\centering
\includegraphics[width=\textwidth,height=0.7\textwidth]{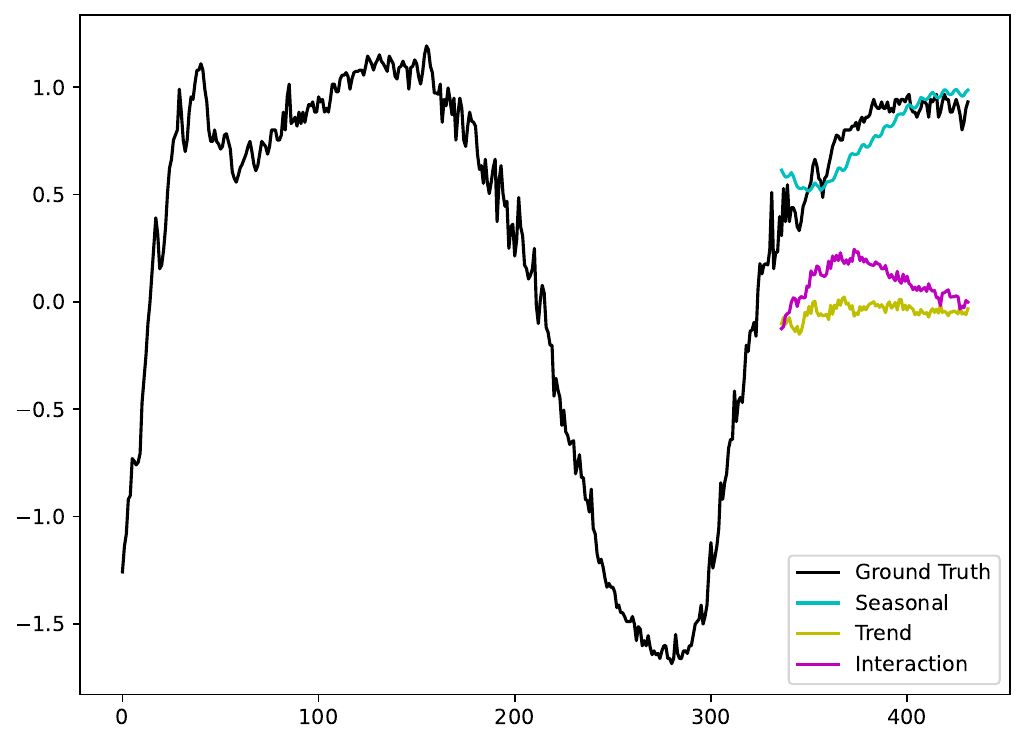}
\caption*{(d)}
\end{minipage}

\vspace{0.5em}
\caption{Forecasting Performance Visualization: Two Case Studies on PEMS08. (a) and (b) form the first case, and (c) and (d) form the second case. (a) and (c) compare FBM-S with TimeMixer and iTransformer, while (b) and (d) show each forecasting component: trend, seasonal, and interaction blocks.}
\label{case}
\end{figure*}

\begin{figure*}[t]
\centering
\begin{minipage}[t]{0.235\linewidth}
\centering
\includegraphics[width=\textwidth,height=0.8\textwidth]{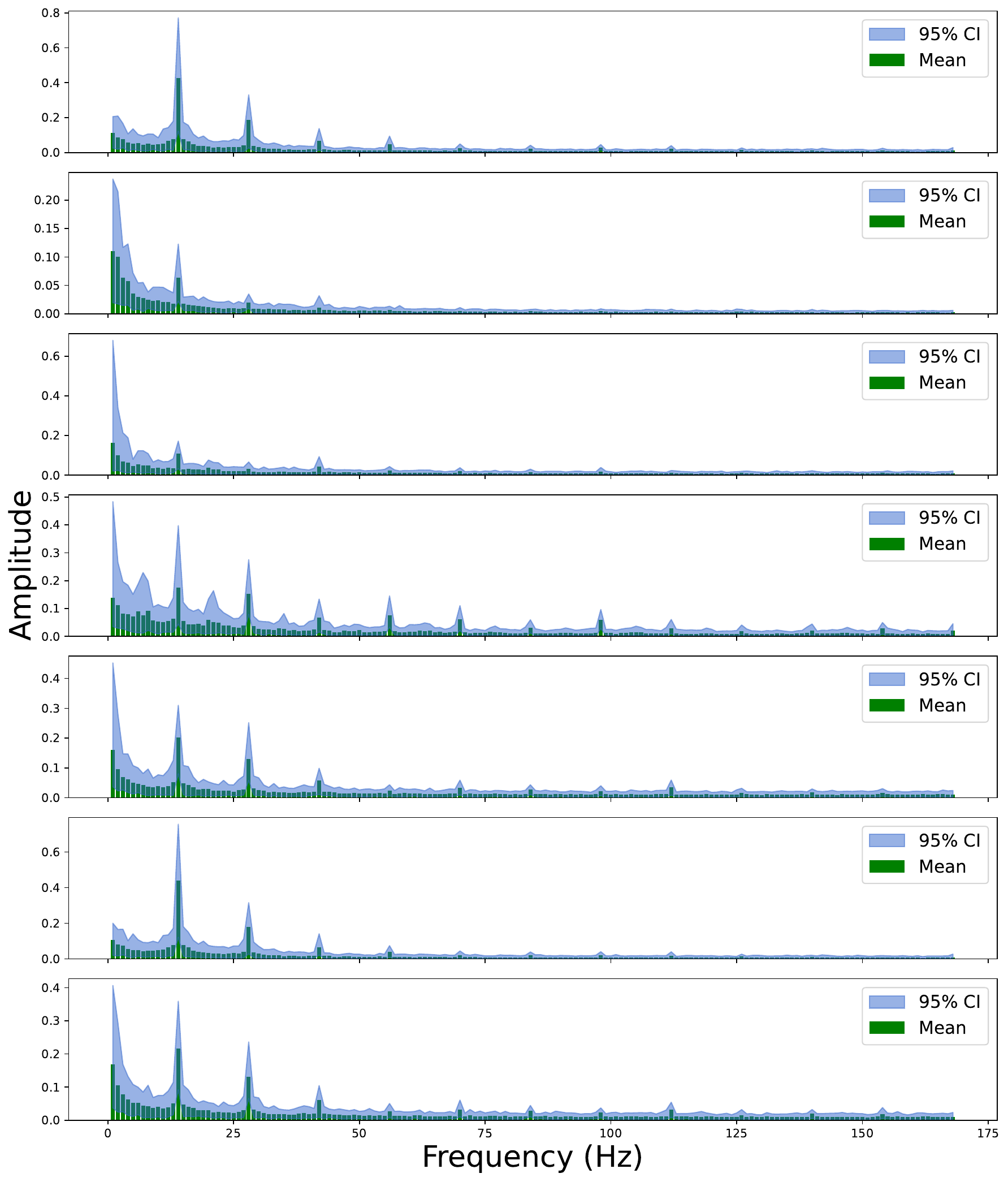}
\caption*{ETTh1}
\end{minipage}%
\begin{minipage}[t]{0.235\linewidth}
\centering
\includegraphics[width=\textwidth,height=0.8\textwidth]{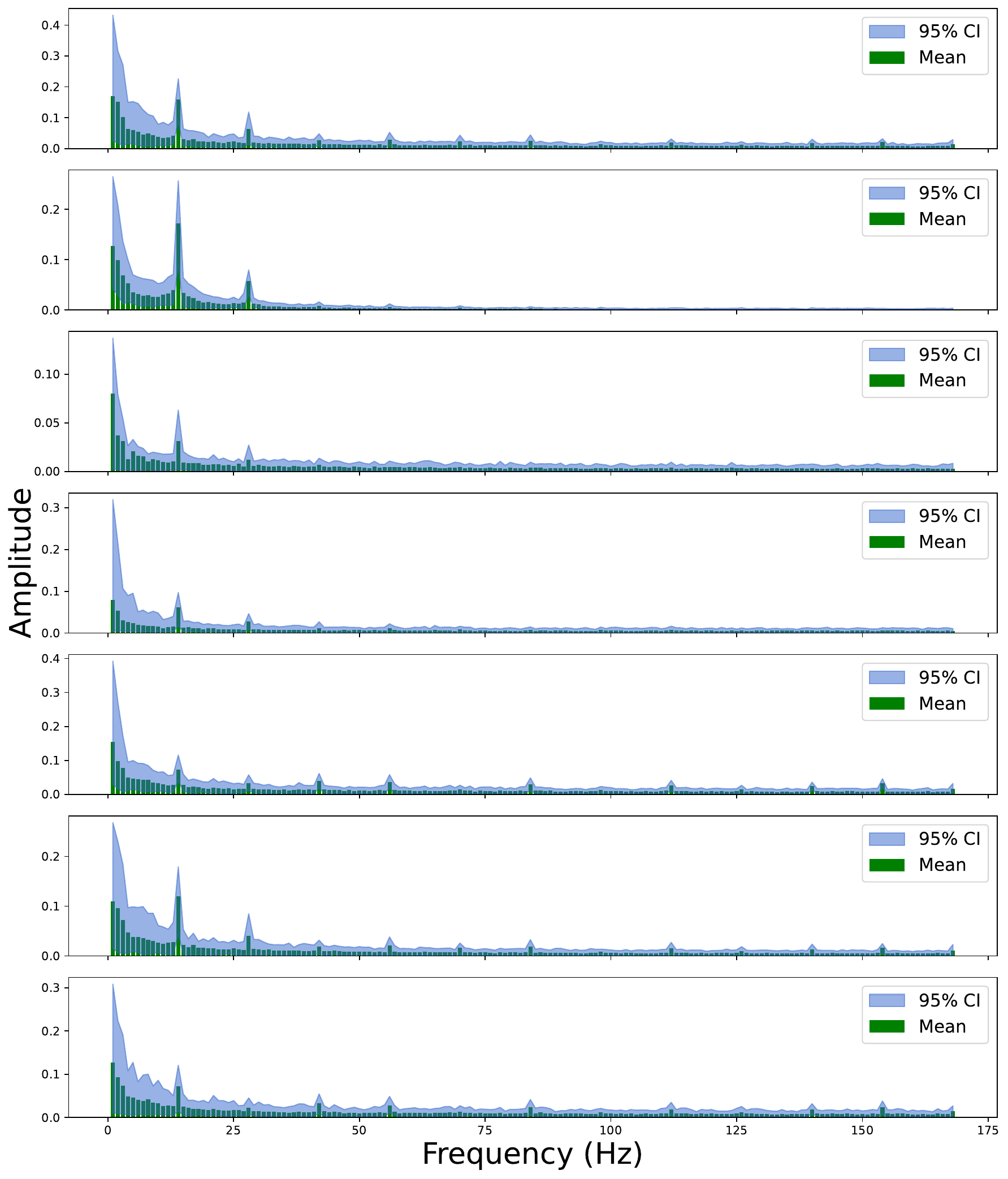}
\caption*{ETTh2}
\end{minipage}%
\begin{minipage}[t]{0.235\linewidth}
\centering
\includegraphics[width=\textwidth,height=0.8\textwidth]{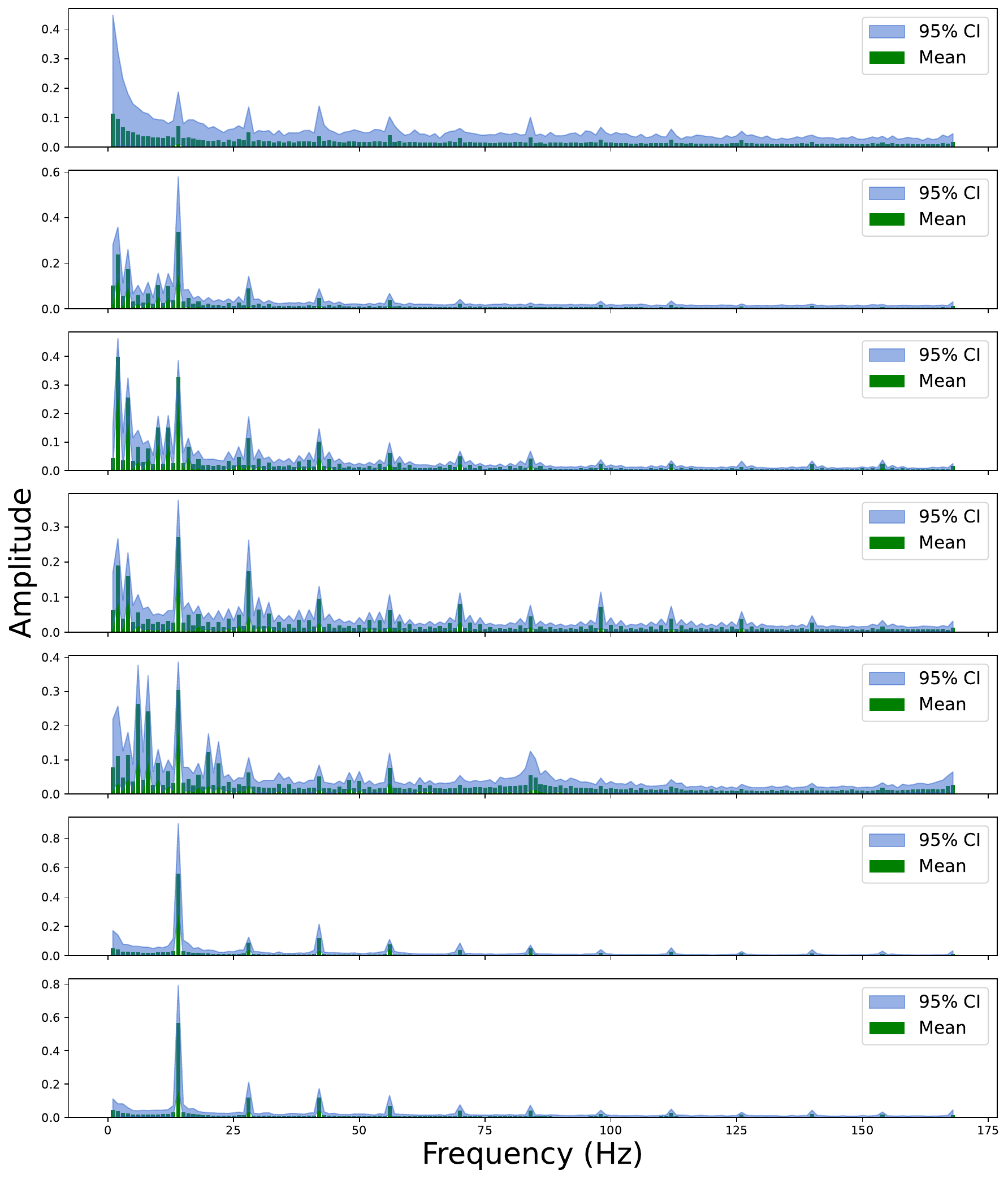}
\caption*{Electricity}
\end{minipage}
\begin{minipage}[t]{0.235\linewidth}
\centering
\includegraphics[width=\textwidth,height=0.8\textwidth]{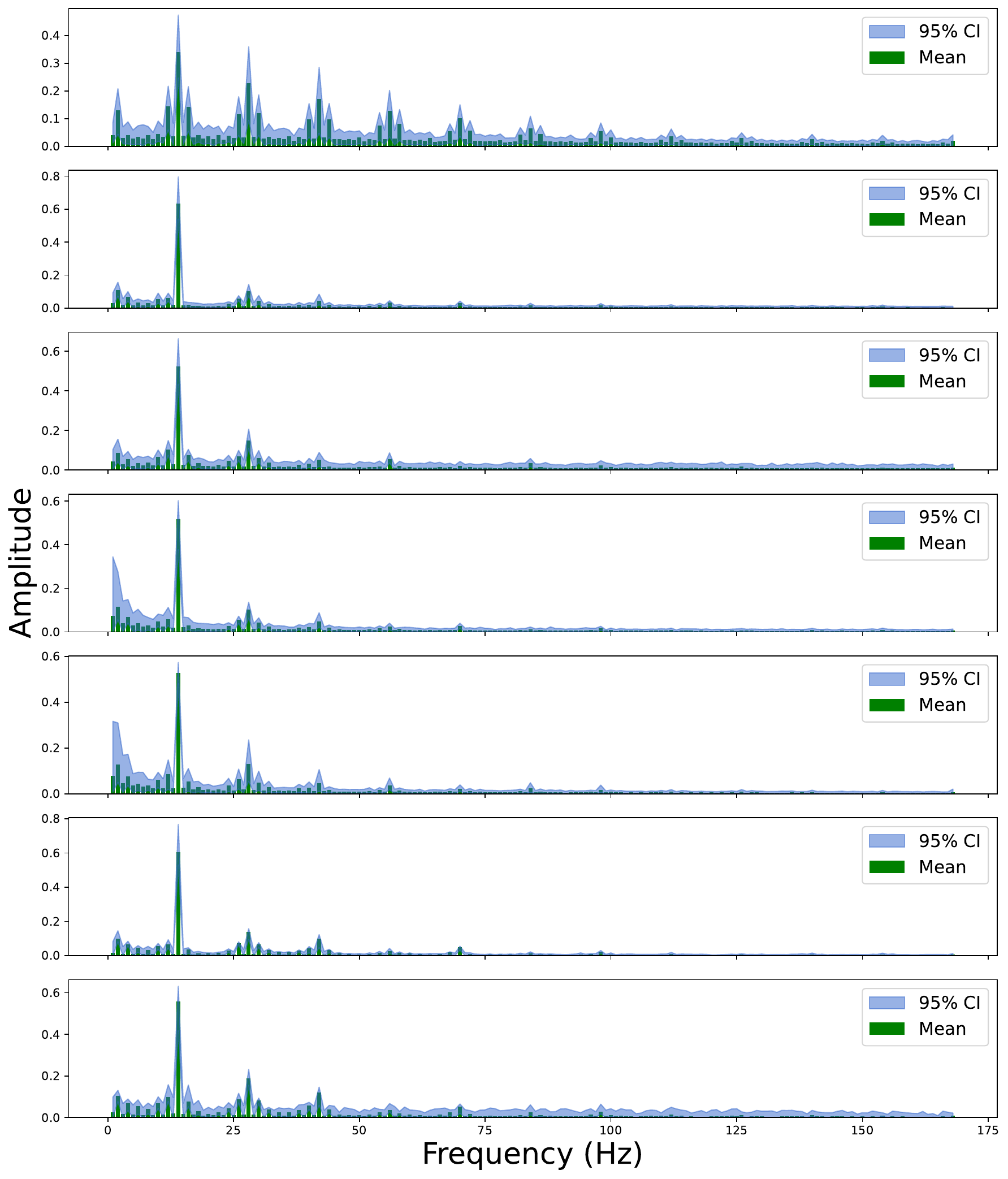}
\caption*{Traffic}
\end{minipage}

\begin{minipage}[t]{0.235\linewidth}
\centering
\includegraphics[width=\textwidth,height=0.8\textwidth]{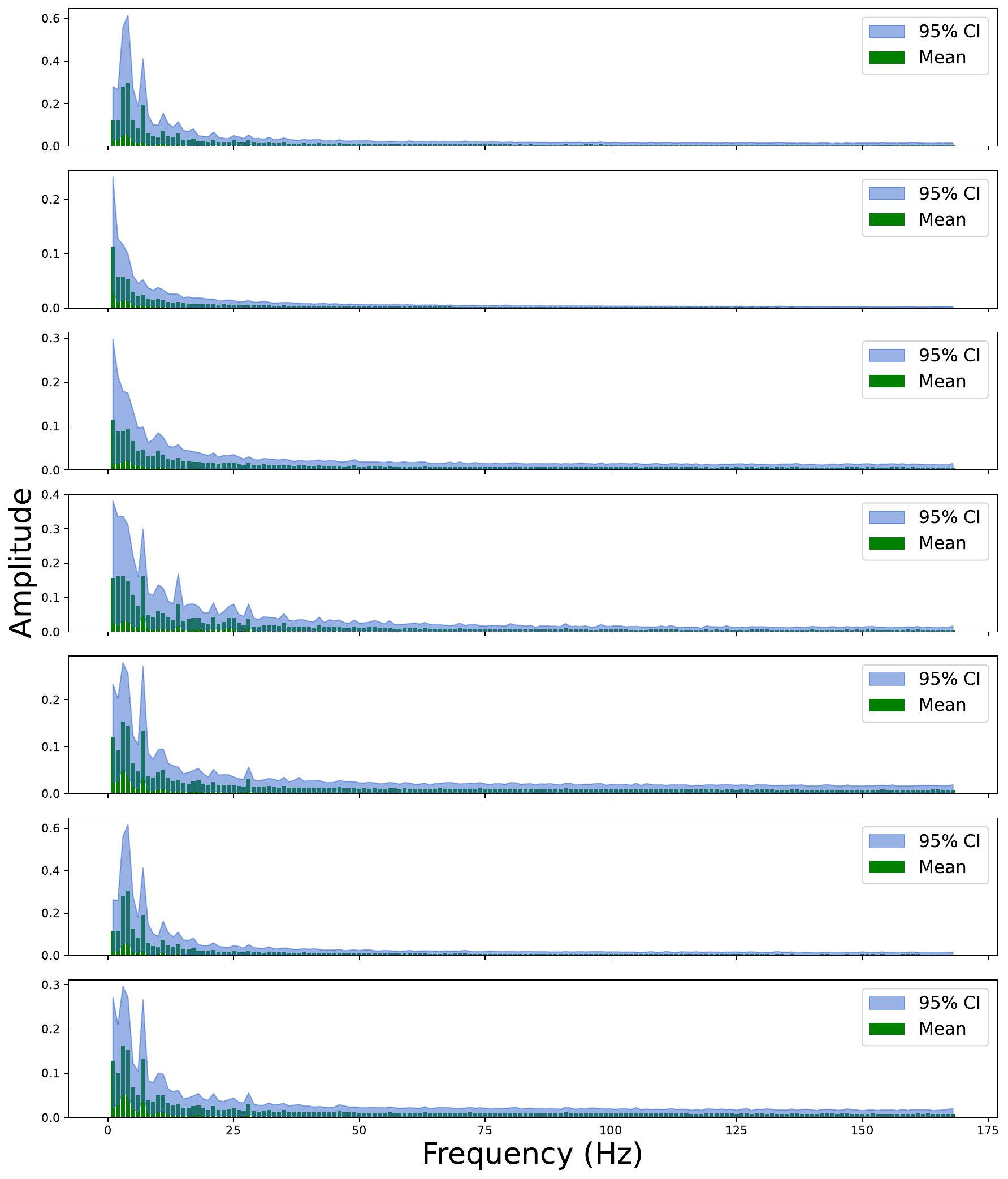}
\caption*{ETTm1}
\end{minipage}%
\begin{minipage}[t]{0.235\linewidth}
\centering
\includegraphics[width=\textwidth,height=0.8\textwidth]{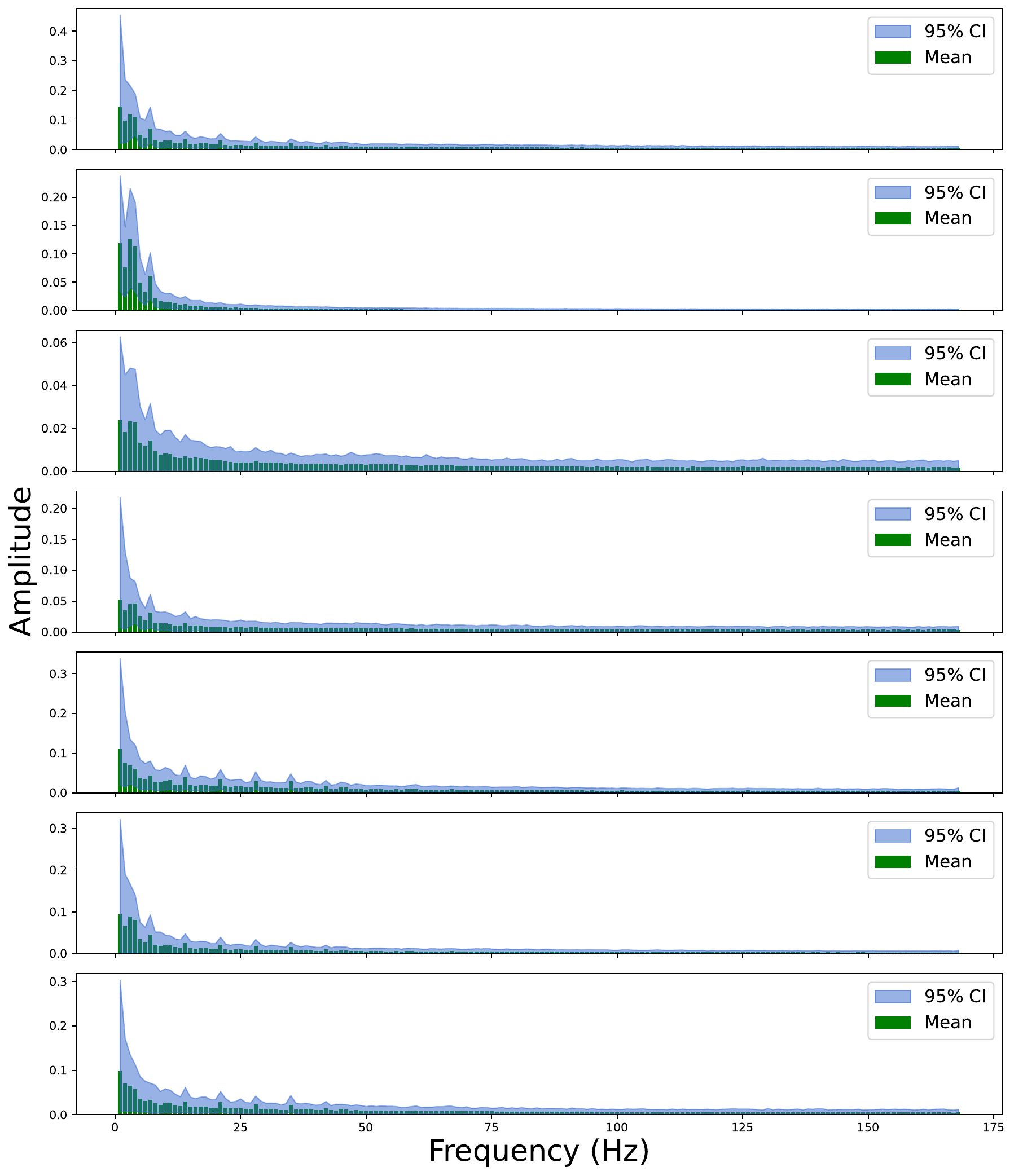}
\caption*{ETTm2}
\end{minipage}%
\begin{minipage}[t]{0.235\linewidth}
\centering
\includegraphics[width=\textwidth,height=0.8\textwidth]{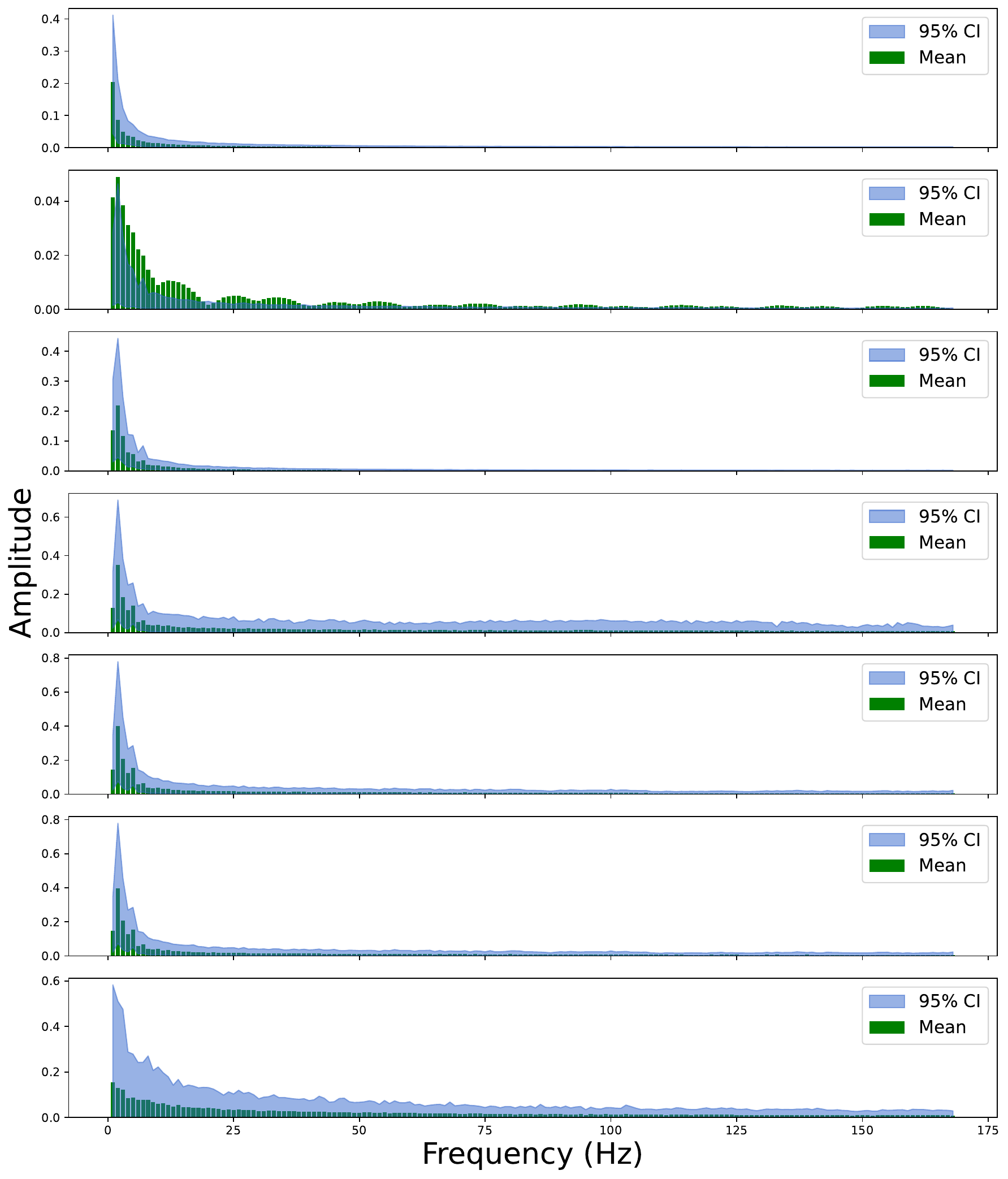}
\caption*{Weather}
\end{minipage}
\begin{minipage}[t]{0.235\linewidth}
\centering
\includegraphics[width=\textwidth,height=0.8\textwidth]{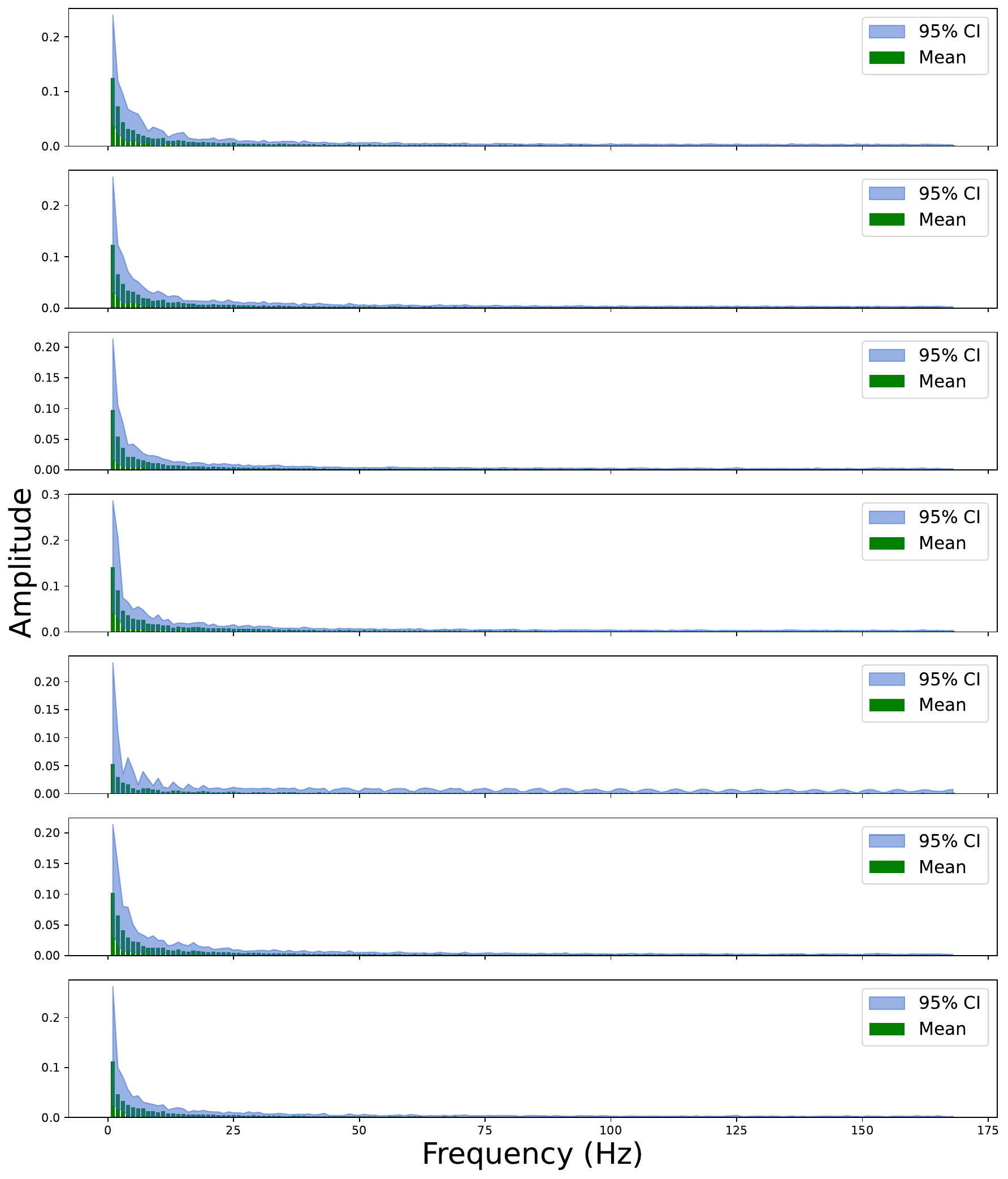}
\caption*{Exchange}
\end{minipage}

\begin{minipage}[t]{0.235\linewidth}
\centering
\includegraphics[width=\textwidth,height=0.8\textwidth]{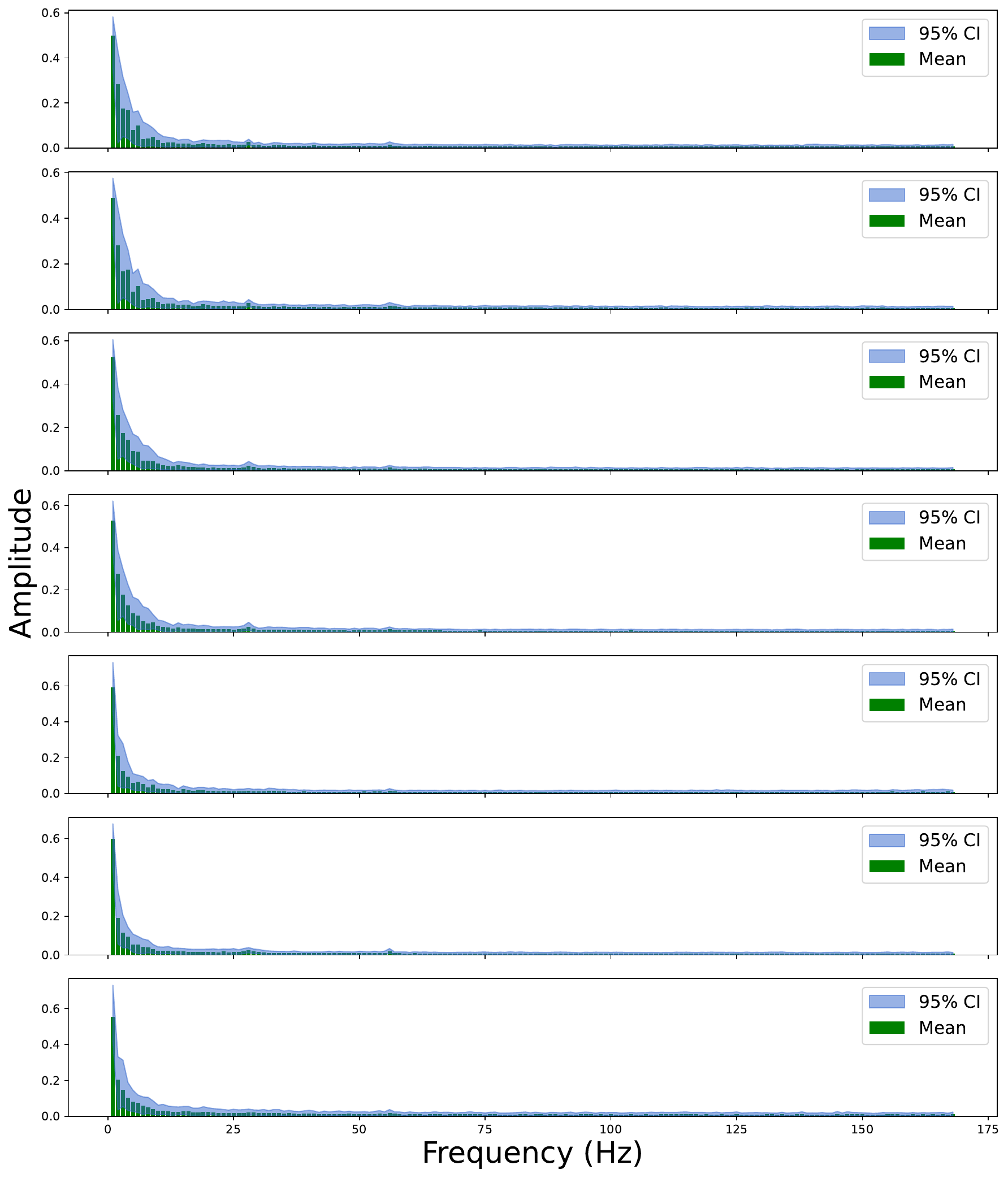}
\caption*{PEMS03}
\end{minipage}%
\begin{minipage}[t]{0.235\linewidth}
\centering
\includegraphics[width=\textwidth,height=0.8\textwidth]{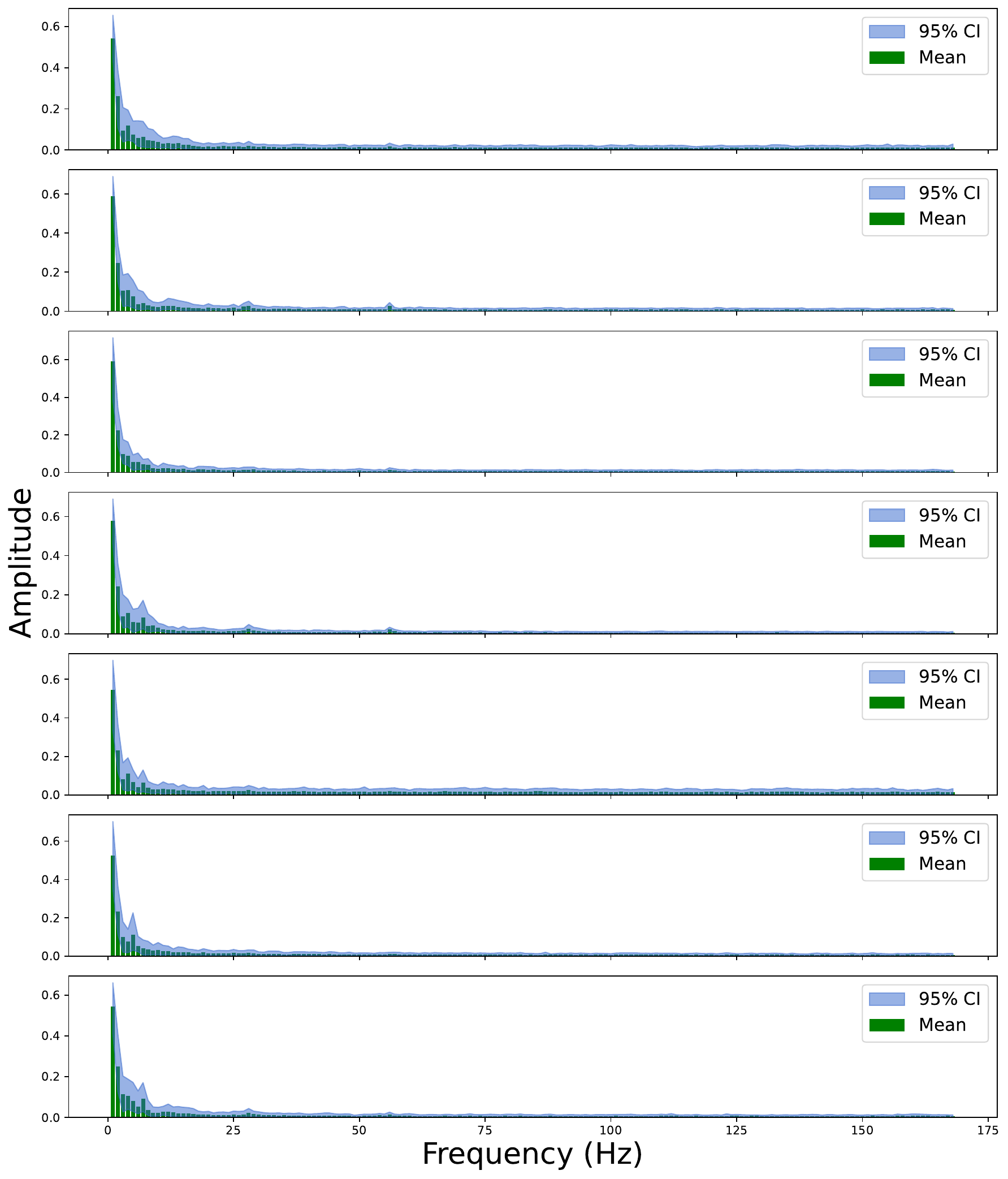}
\caption*{PEMS04}
\end{minipage}%
\begin{minipage}[t]{0.235\linewidth}
\centering
\includegraphics[width=\textwidth,height=0.8\textwidth]{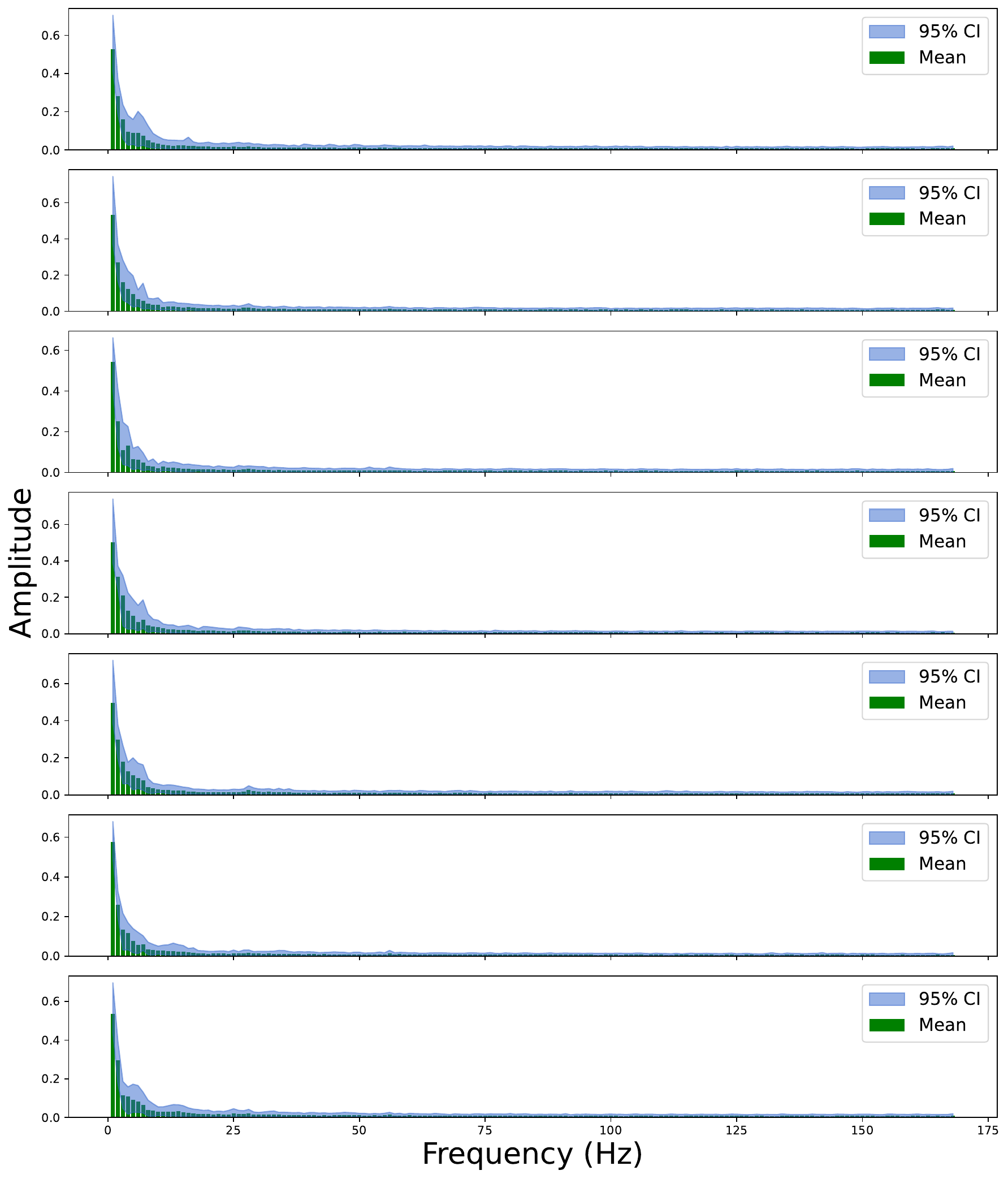}
\caption*{PEMS07}
\end{minipage}
\begin{minipage}[t]{0.235\linewidth}
\centering
\includegraphics[width=\textwidth,height=0.8\textwidth]{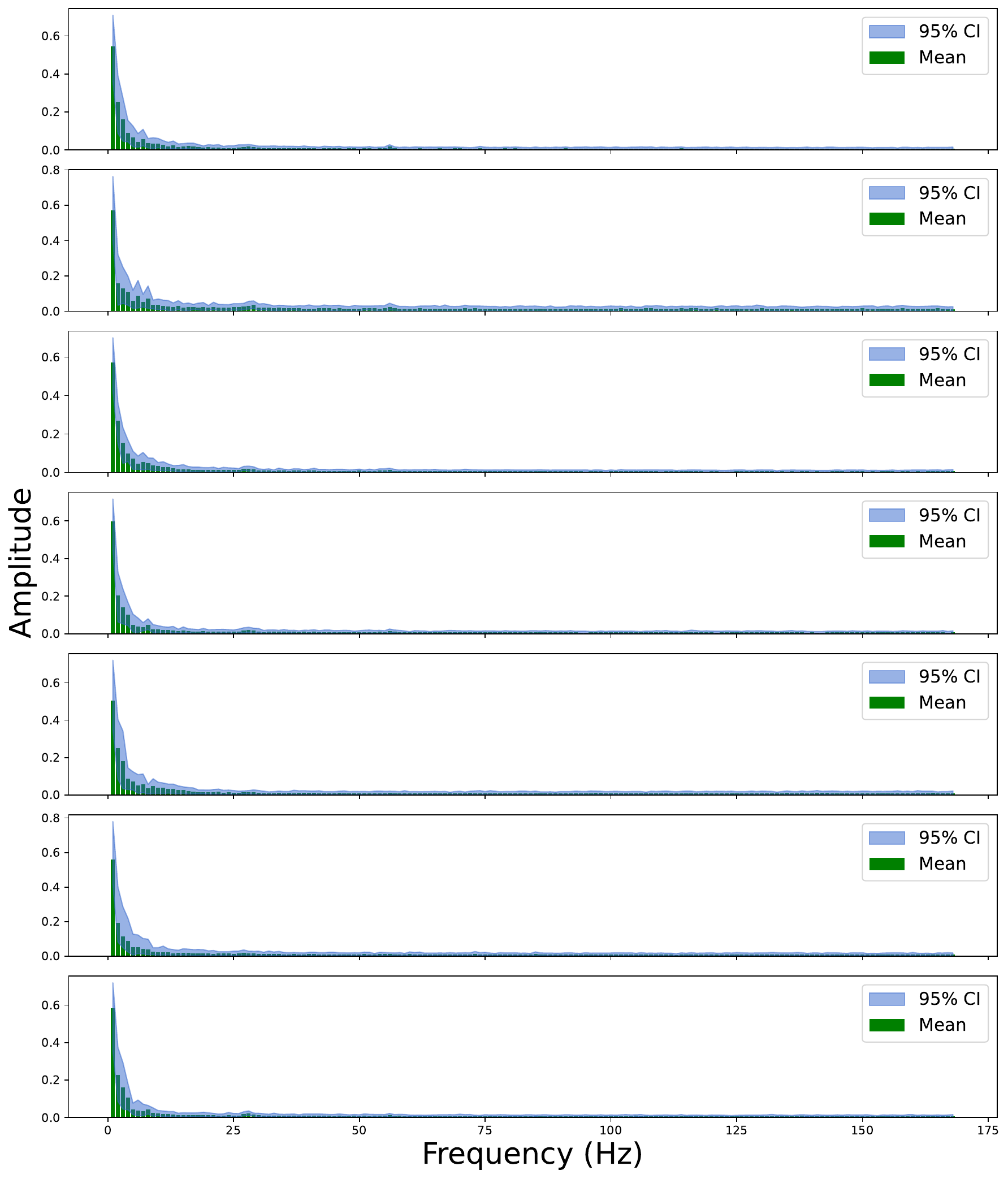}
\caption*{PEMS08}
\end{minipage}

\caption{Frequency Spectrum Distribution of the Last Seven Dimensions on the Twelve Datasets. Here, the green bars represent the mean values, while the light blue shaded area indicates the 95 percentage confidence interval. }
\label{specturm}
\vspace{-10pt}
\end{figure*}

\subsection{Rolling Window and Patching for Enhanced Seasonal and Trend Mapping}
\label{rolling}

Here, we aim to understand why using a sliding window in the seasonal block and the patched projection in the trend block can improve both efficiency and performance. This is because the original Fourier basis follows sinusoidal patterns, and when a window filter slides over the time domain, it can automatically capture seasonal features. To support this claim, we visualize the heatmaps of the learned weights of FBM-L on the Electricity dataset in Fig. \ref{weight2}, to better understand how FBM-L captures the relationships between input time-frequency features and the output time series. Specifically, we decompose the weight matrix $\mathbf{W}$ into $L$ time-specific components, where each $\mathbf{W}_i$ represents the influence of the time-frequency features on the $i$-th time step of the predicted output time series $\mathbf{\hat{Y}}$. The heatmaps resemble Fourier basis patterns but differ in finer details, indicating that it allows the model to remove noise across both the time and frequency domains. We also observe that when heatmaps $\mathbf{W}_i$ gradually shift along $i$, they produce similar patterns over time. This indicates that the similarities captures coarse seasonal features, while the finer differences correspond to residual trend effects. This observation suggests that the weights can be shared via convolution, allowing a single weight matrix $\mathbf{W}$ to capture the overall coarse seasonal structure. Thus, the patched projection combined with an activation function can focus on extracting finer patterns related to trend effects within a specific period.



\subsection{Case Study on Synergistic Effects through Three Blocks}
\label{case_study}
In Fig. \ref{case}, we present two case studies illustrating the forecast performance of our model compared to TimeMixer and iTransformer on the PEMS08 dataset, along with visualizations of the trend, seasonal, and interaction components. These two models represent the SOTA baselines for channel independent and channel dependent modeling, respectively. First, Figs \ref{case}a and \ref{case}c show that our model achieves better forecasting results than both baselines. Second, Figs \ref{case}b and \ref{case}d explain why our synergistic blocks work better. We observe that the seasonal, trend, and interaction components capture meaningful corresponding patterns, respectively. This suggests that the model effectively leverages different modalities: the trend and seasonal signals from endogenous factors, as well as interaction effects from exogenous variables, combined with both time-frequency space.

In Fig. \ref{case}b and Fig. \ref{case}c, we find that the model generates more prominent features from the seasonal and interaction backbones, indicating that these components play a more important role through backpropagation in STSF. On the other hand, We observe that the trend component tends to stay close to the zero axis, implying that it is refining fine-grained patterns not captured by the other blocks. For example, the other baselines shown in Figs \ref{case}a and \ref{case}b fail to capture the overall effects accurately. In contrast, our model produces more precise and reliable forecasts across different modalities. Specifically, the trend block generates outputs greater than zero in Fig. \ref{case}b, while the interaction block produces adjusted values in Fig. \ref{case}d. The synergistic effects help correct the mis-predictions has not been considered by the other baselines in Figs \ref{case}a and \ref{case}b, respectively. Thus, it explains why FBM-S achieves significantly better forecasting results.

\subsection{Data Characteristics for Interpretable Results}
\label{Distribution}
To better understand the data characteristics, we visualize the distribution of the input frequency spectrum in Fig. \ref{specturm}, which aids in interpreting the experimental results and informs our hyperparameter selections. The visualization presents the mean input frequency spectrum with its 95 percent confidence interval across different datasets. In Tables \ref{table1} and \ref{table_short}, we observe that FBM-NL and FBM-NP perform similarly across most datasets, as both are nonlinear models, though FBM-NL generally outperforms FBM-NP except for the Traffic dataset. This is due to the fact that the frequency spectrum distribution of the Traffic dataset shows the least diversity and variation, making it more suitable for a transformer-based architecture. Thus, the Transformer-based architecture is more prone to overfitting in datasets where the frequency spectrum is highly diverse and variable. Secondly, FBM-L and FBM-NL show significant performance differences on the ETTh dataset compared to the others, likely due to a common underlying issue. The ETTh dataset exhibits the most diverse frequency spectrum and the greatest variation, indicating a higher proportion of noise signals on ETTh1 and ETTh2 than on the other datasets. As a result, a simpler linear model is more suitable for capturing the effects on the ETTh datasets.

Fig. \ref{specturm} shows that trend effects generally appear at low-frequency levels, seasonal effects typically occur at intermediate frequencies, and noise signals usually arise high-frequency levels. This explains why the Fourier transform is beneficial for TSF, as it hierarchically separates various effects, with specific effects (e.g., hourly, daily) aligning with their corresponding frequency levels. For instance, Fig. \ref{specturm} shows consistently high energy at multiples of 14 across all hourly-level datasets: ETTh1, ETTh2, Electricity, and Traffic. This is attributed to the day effect falling into these frequency levels, given the repeating cycle of 24 with a look-back window of 336. 

Variations in the frequency spectrum also indicate that the spectral landscape is highly sensitive to the characteristics of the underlying data. For the hourly-level datasets, the frequency spectrum are more diverse, whereas the minute-level datasets tend to exhibit much more energy gathering in the lower frequency range. For instance, the frequency spectrum of ETTm1 and ETTm2 are shifted to the left compared to those of ETTh1 and ETTh2. This shift effectively shortens the forecast horizon in the time domain, resulting in reduced noise in these data. This helps explain why non-Linear network performs slightly better than linear network on ETTm1 when the granularity changes. This also applies to the other datasets. For example, the PEMS dataset, which has the highest data granularity at a five-minute level, exhibits a frequency spectrum that is even more concentrated at lower frequency levels. These findings support our initial hypothesis: simply providing the real and imaginary parts is not sufficient, as frequency interpretations are ambiguous when changes in data granularity or sequence length alter the meaning of those components. In conclusion, when a dataset contains a large proportion of noise and limited data, simpler models tend to perform better; vice versa.

\section{Conclusion and future work}

We make the first attempt to theoretically and empirically discuss several issues related to existing Fourier-based methods for time series forecasting from the perspective of basis functions. Our insights and findings disclose two issues commonly appearing in existing Fourier-based studies: inconsistent starting cycles and inconsistent series length issues. Thus, we address the aforementioned issue by including the basis functions, retaining fine grain time-domain information. This allows the model to take advantage of both time and frequency modality. Therefore, we propose a new time-frequency learning  framework, namely Fourier Basis Mapping (FBM). Extensive experiments demonstrate the effectiveness of the FBM approach via its three variants: FBM-L, FBM-NL, and FBM-NP, which enhance various mapping networks. Then, we further propose a novel synergistic FBM model, referred to as FBM-S, which further enhances the SOTA performance in both short-term and long-term TSF tasks. This model decomposes the effects into three components: trend, seasonal, and interaction, combining the strengths of independent channel and dependent channel modeling. We also propose several useful techniques for modeling time-frequency features, which are validated by seven ablation studies conducted on both tasks. However, there is one limitation of our methods that the input sequence length should not be too short, as the meaningful Fourier basis functions are bounded by the input sequence length. Finally, we conclude that time-frequency features hold great potential for future time series analysis tasks. The FBM framework offers a new pathway for time-frequency learning for TSF. In the future, we will further extend its applications to other domains, such as anomaly detection and classification.


\bibliographystyle{IEEEtran}
\bibliography{mybibfile}

\begin{IEEEbiography}[{\includegraphics[width=1in,height=1.25in,clip,keepaspectratio]{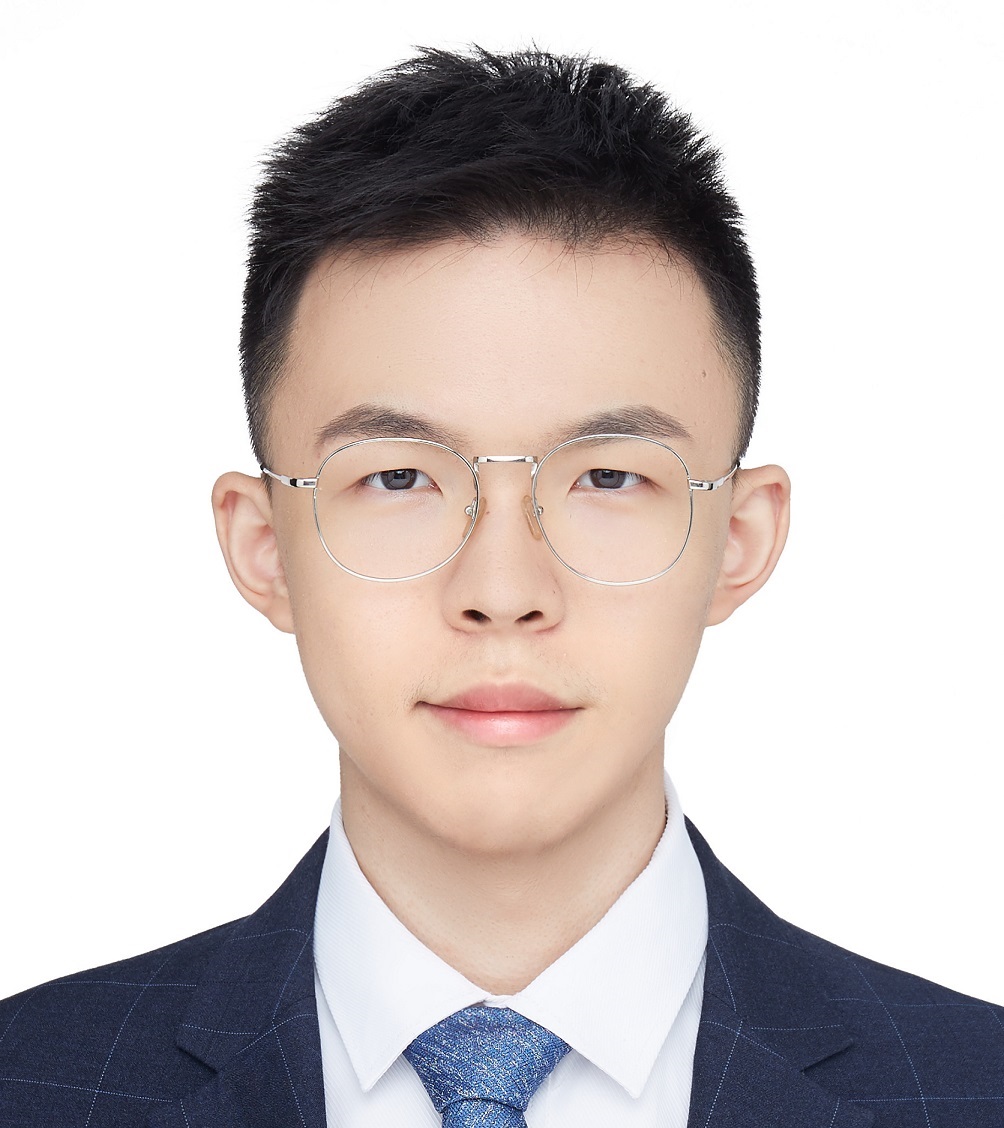}}]{Runze Yang} received the B.S. and M.S. degrees from the Faculty of Mathematical and Physical Sci
ences, University College London, London, U.K.,
 in 2019 and 2020, respectively. He is a Cotutelle PhD student at both Shanghai Jiao Tong University and Macquarie University. His research interests include multivariate time series forecasting, signal processing, pattern recognition, and machine learning.
\end{IEEEbiography}

\begin{IEEEbiography}[{\includegraphics[width=1in,height=1.25in,clip,keepaspectratio]{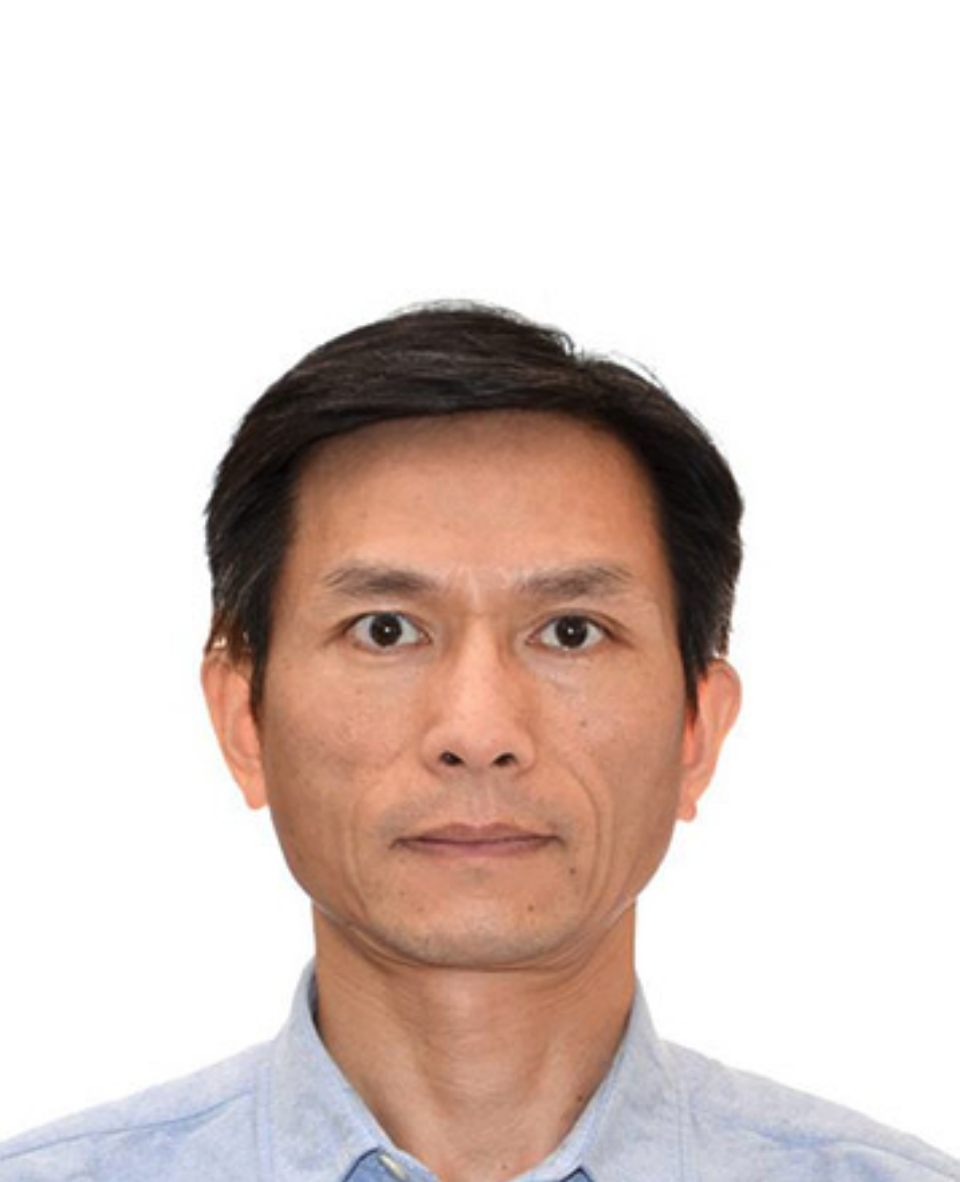}}]{Longbing Cao}(SM'06) received a PhD degree in pattern recognition and intelligent systems at Chinese Academy of Sciences in 2002 and another PhD in computing sciences at University of Technology Sydney in 2005. He is the Distinguished Chair Professor in AI at Macquarie University and an Australian Research Council Future Fellow (professorial level). His research interests include AI and intelligent systems, data science and analytics, machine learning, behavior informatics, and enterprise innovation.
\end{IEEEbiography}

\begin{IEEEbiography}
[{\includegraphics[width=1in,height=1.25in,clip,keepaspectratio]{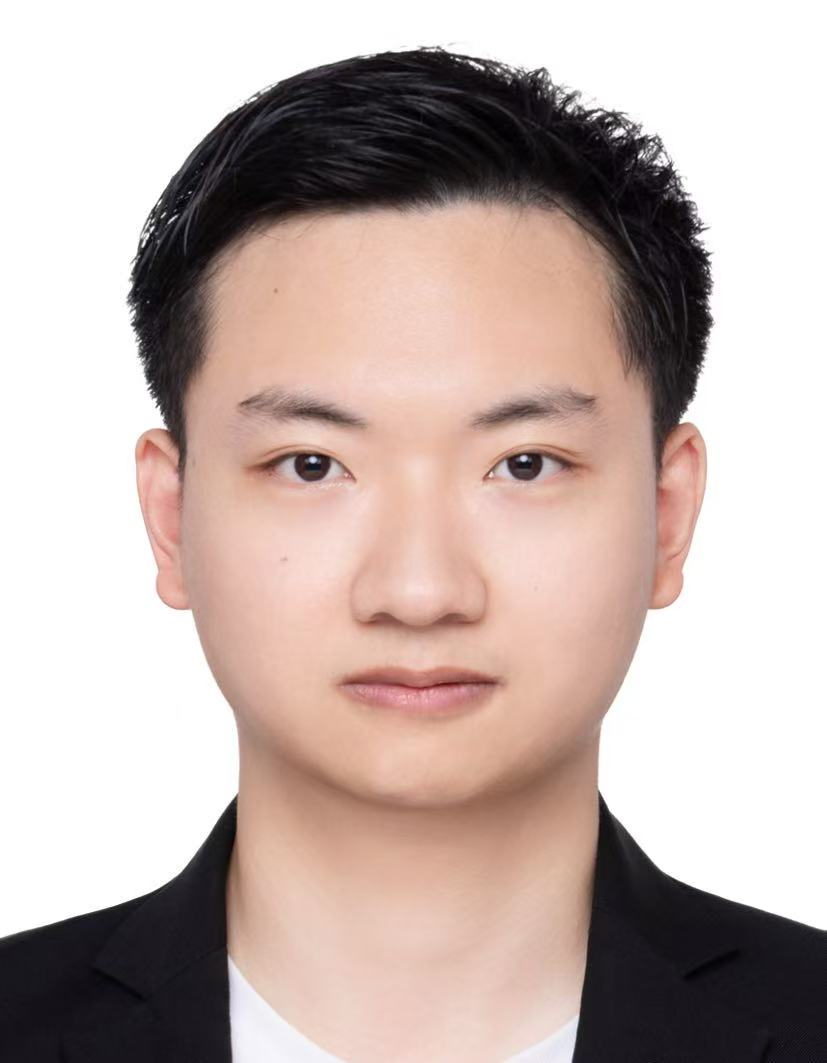}}]{Xin You} received the B.S. degree in Department of Automation from Harbin Institute of Technology, Harbin, China, in 2020. He is currently working towards the Ph.D. degree majoring at the Institute of Image Processing and Pattern Recognition, Department of Automation, Shanghai Jiao Tong University, supervised by Prof. Yun Gu. His research interests include medical image segmentation, video frame interpolation, medical image synthesis.
\end{IEEEbiography}

\begin{IEEEbiography}[{\includegraphics[width=1in,height=1.25in,clip,keepaspectratio]{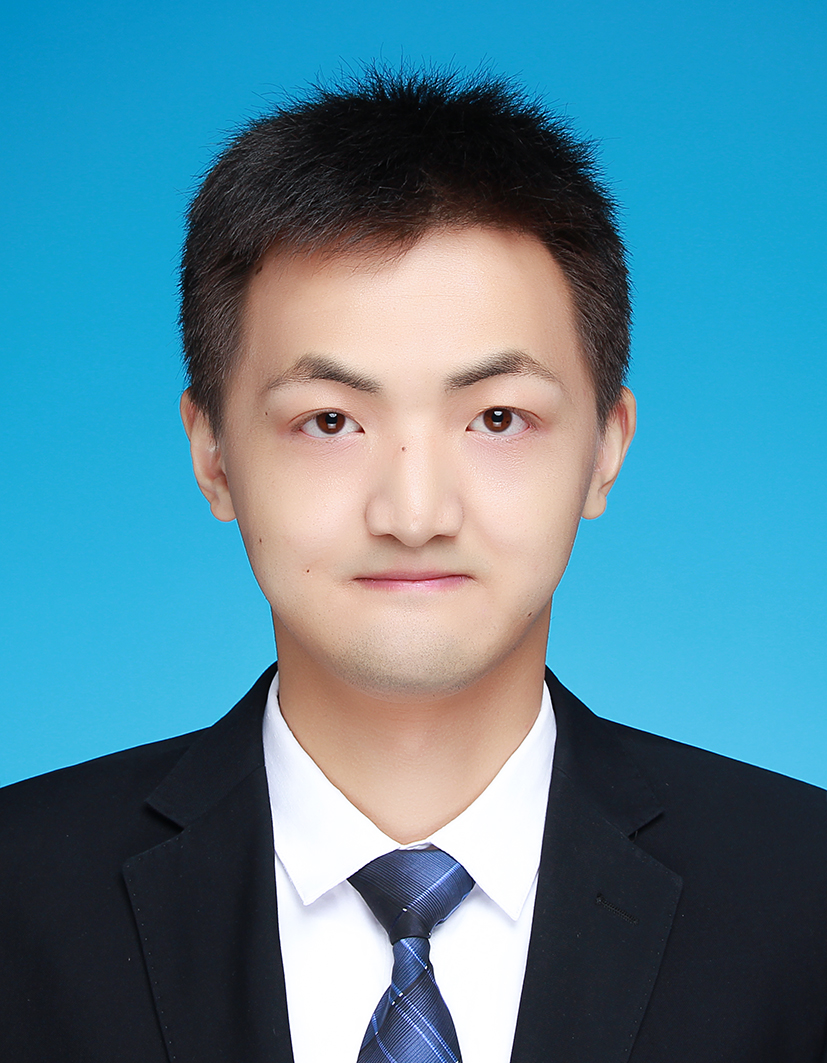}}]{Kun Fang} received the B.S. degree from Tongji University, Shanghai, China, in 2018, and the M.S. and Ph.D. degrees from Shanghai Jiao Tong University, Shanghai, China, in 2021 and 2025, respectively. He is now a postdoctoral fellow at The Hong Kong Polytechnic University. His current research interests include robustness and privacy of deep neural networks.
\end{IEEEbiography}

\begin{IEEEbiography}[{\includegraphics[width=1in,height=1.25in,clip,keepaspectratio]{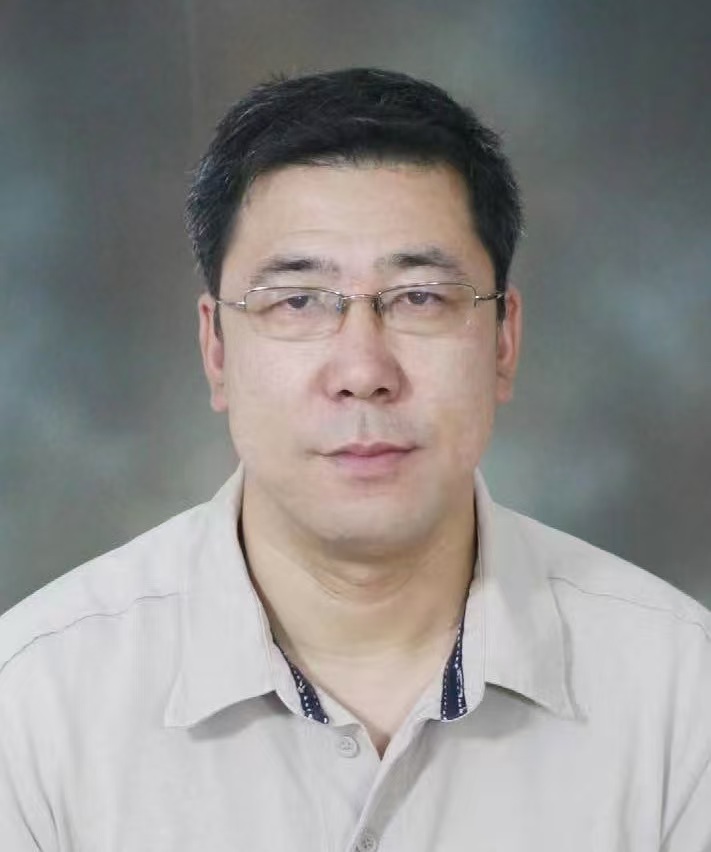}}]{Jianxun Li}
received the Dr. Eng. degree in control theory and engineering with highest honors from Northwestern Polytechnical University, Xi'an, China, in 1996. He is currently a Professor with the Department of Automation, Shanghai Jiao Tong University, Shanghai, China. His research interests include information fusion, infrared image processing, and parameter estimation.
\end{IEEEbiography}

\begin{IEEEbiography}[{\includegraphics[width=1in,height=1.25in,clip,keepaspectratio]{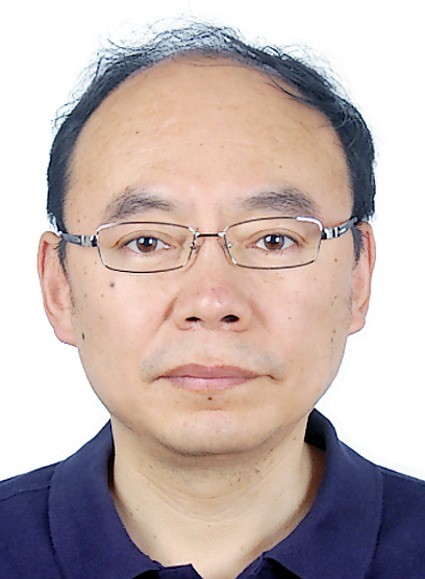}}]{Jie Yang} received the bachelor’s and master’s degrees from Shanghai Jiao Tong University, Shanghai, China, in 1985 and 1988, respectively, and the Ph.D. degree from the University of Hamburg, Hamburg, Germany, in 1994.  He is currently a Professor and the Director of the Institute of Image Processing and Pattern Recognition, Shanghai Jiao Tong University. His research interests include image processing, pattern recognition, data mining, and artificial intelligence.
\end{IEEEbiography}
\appendix

\section{Proof of real-valued discrete Fourier transform via basis functions}
\label{proof1}
We provide the proof that the real-valued inverse discrete Fourier transform can be represented by sine and cosine functions as follows:

\begin{equation}
\begin{aligned}
& \mathbf{X}[n]=\frac{1}{T}\sum_{k=0}^{T-1} \mathbf{H}[k] \exp \left(i \frac{2 \pi k n}{T} \right) \\
& =\sum_{k=0}^{\frac{T}{2}} \mathbf{H}[k] (\cos (\frac{2 \pi k n }{T})+i \sin (\frac{2 \pi k n }{T}))\\
 &+ \sum_{k=1}^{\frac{T}{2}-1} \mathbf{H}[T-k] (\cos (\frac{-2 \pi k n }{T})+i \sin (\frac{-2 \pi k n }{T}))), \\
 & =\frac{1}{T}(\mathbf{H_R}[0]+ \mathbf{H_R}[\frac{T}{2}]\cos ( \pi n)\\
 &+ \sum_{k=1}^{\frac{T}{2}-1} (\mathbf{H_R}[k]+i \mathbf{H_I}[k]) (\cos (\frac{2 \pi k n }{T})+i \sin (\frac{2 \pi k n }{T}))\\
&+ \sum_{k=1}^{\frac{T}{2}-1} (\mathbf{H_R}[k]-i \mathbf{H_I}[k] ) (\cos (\frac{2 \pi k n }{T})-i \sin (\frac{2 \pi k n }{T}))), \\
&=\frac{2}{T}\sum_{k=1}^{\frac{T}{2}-1}\left(\mathbf{H_R}[k] \cos \left(\frac{2 \pi k n}{T}\right)-\mathbf{H_I}[k] \sin \left(\frac{2 \pi k n}{T}\right)\right)\\
&+\frac{1}{T}(\mathbf{H_R}[0]+\mathbf{H_R}[\frac{T}{2}]\cos (\pi T)),\\
&=\frac{1}{T}\sum_{k=0}^{\frac{T}{2}}\left(a_k \cos \left(\frac{2 \pi k n}{T}\right)-b_k \sin \left(\frac{2 \pi k n}{T}\right)\right)\\ 
 n&=0,\ldots,T-1,\\
a_k&= \begin{cases} \mathbf{H_R}[k],  \\ 2 \cdot \mathbf{H_R}[k], &\end{cases}  
b_k= \begin{cases} \mathbf{H_I}[k], & k=0,\frac{T}{2} \\ 2 \cdot \mathbf{H_I}[k], & k=1, \ldots, \frac{T}{2}-1.\end{cases}\\
\end{aligned}
\end{equation}

\vfill
\break

\begin{figure}[h]
    \centering
    \includegraphics[width=\linewidth]{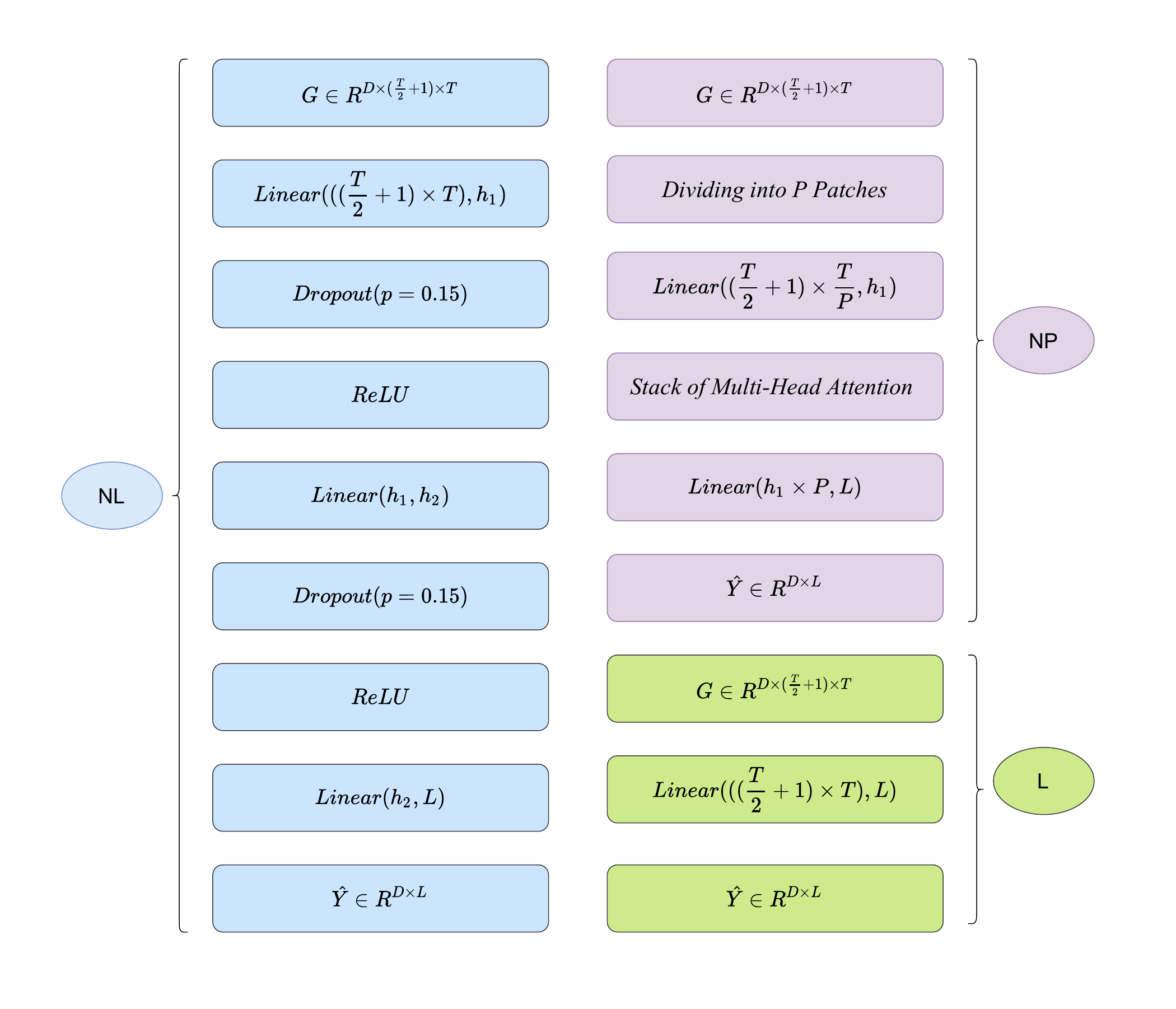}
    \caption{
    The Layers of FBM-L, FBM-NL, and FBM-NP in Detail. FBM-L is a vanilla linear network, FBM-NL is a three-layer MLP, and FBM-NP shares the same structure as PatchTST but performs patching based on time segments of the time-frequency features. Let $G$ denote the time-frequency features obtained after Fourier basis expansion.
     }
    \label{MLP}
\end{figure}

\end{document}